\documentclass{article}
\PassOptionsToPackage{numbers, compress}{natbib}
\usepackage[final]{neurips_2021}
\usepackage{comment}
\usepackage{cuted}
\usepackage{mathtools}
\usepackage{caption}
\usepackage{wrapfig}

\usepackage[utf8]{inputenc} %
\usepackage[T1]{fontenc}    %
\usepackage{hyperref}       %
\usepackage{url}            %
\usepackage{booktabs}       %
\usepackage{amsfonts}       %
\usepackage{nicefrac}       %
\usepackage{microtype}      %
\usepackage{xcolor,colortbl}         %

\usepackage{multirow}
\usepackage{enumitem}

\usepackage{bibentry}

\usepackage{appendix}

\begin{document}
\newcommand\blfootnote[1]{%
  \begingroup
  \renewcommand\thefootnote{}\footnote{#1}%
  \addtocounter{footnote}{-1}%
  \endgroup
}

\title{A-NeRF: Articulated Neural Radiance Fields for Learning Human Shape, Appearance, and Pose}
\author{
Shih-Yang Su$^1$ \qquad Frank Yu$^1$ \qquad Michael Zollh\"ofer$^2$ \qquad Helge Rhodin$^1$\\
\\
$^1$University of British Columbia \qquad $^2$Facebook Reality Labs
}

\newcommand{\red}[1]{{\color{red}#1}}
\newcommand{\tb}[1]{\textbf{#1}}
\newcommand{\tbi}[1]{\textit{\textbf{#1}}}
\newcommand{\norm}[1]{\left\|#1\right\|}

\newcommand{\figref}[1]{Figure~\ref{#1}}
\newcommand{\secref}[1]{Section~\ref{#1}}
\newcommand{\feqref}[1]{Equation~\eqref{#1}}
\newcommand{\tabref}[1]{Table~\ref{#1}}
\newcommand{\sy}[1]{{\color{blue}#1}}
\newcommand{\SY}[1]{{\color{blue}(Shih-Yang:#1)}}
\newcommand{\hr}[1]{{\color{red}#1}}
\newcommand{\HR}[1]{{\color{red}(Helge:#1)}}
\newcommand{\MZ}[1]{{\color{cyan}(Michael:#1)}}
\newcommand{\fy}[1]{{\color{green}#1}}
\newcommand{\FY}[1]{{\color{green}(Frank:#1)}}

\newcommand{\rev}[1]{{\color{red}#1}}

\newcommand{\hmvec}[1]{
\begin{bmatrix}#1 \\ 1
\end{bmatrix}
}

\newcommand{\veclist}[2]{
\left[#1_1,#1_2,\cdots,#1_{#2}\right]
}
\newcommand{\veclistki}[2]{
[#1_{k,1},\cdots,#1_{k,#2}]
}
\newcommand{\veclistkiz}[2]{
[#1_{k,0},\cdots,#1_{k,#2}]
}
\newcommand{\subki}[1]{#1_{k,i}}
\newcommand{\subkj}[1]{#1_{k,j}}
\newcommand{\subkm}[1]{#1_{k,m}}
\newcommand{\subkl}[1]{#1_{k,l}}

\newcommand{\wpe}[2]{\gamma\left(#1_{k,#2},w_{k,#2}\right)}
\newcommand{\pesin}[2]{\sin\left(2^{#1}\pi#2\right)}
\newcommand{\pecos}[2]{\cos\left(2^{#1}\pi#2\right)}

\definecolor{olive}{RGB}{50,150,50}
\definecolor{frank}{RGB}{198, 3, 252}

\newcommand{\TODO}[1]{\textcolor{red}{TODO: #1}}
\newcommand{\todo}[1]{\textcolor{red}{#1}}

\newcommand{\R}{\mathbb{R}}
\newcommand{\argmin}{\operatornamewithlimits{argmin}}

\newcommand{\supp}{supplemental material}

\newcommand{\mA}{\mathbf{A}}
\newcommand{\mB}{\mathbf{B}}
\newcommand{\mC}{\mathbf{C}}
\newcommand{\mD}{\mathbf{D}}
\newcommand{\mE}{\mathbf{E}}
\newcommand{\mF}{\mathbf{F}}
\newcommand{\mG}{\mathbf{G}}
\newcommand{\mGamma}{\mathbf{\Gamma}}
\newcommand{\mH}{\mathbf{H}}
\newcommand{\mI}{\mathbf{I}}
\newcommand{\mJ}{\mathbf{J}}
\newcommand{\mK}{\mathbf{K}}
\newcommand{\mL}{\mathbf{L}}
\newcommand{\mM}{\mathbf{M}}
\newcommand{\mN}{\mathbf{N}}
\newcommand{\mO}{\mathbf{O}}
\newcommand{\mP}{\mathbf{P}}
\newcommand{\mQ}{\mathbf{Q}}
\newcommand{\mR}{\mathbf{R}}
\newcommand{\mRvr}{\mathbf{R}_{\text{virt}\rightarrow\text{real}}}
\newcommand{\mRrv}{\mRvr^{-1}}%
\newcommand{\mS}{\mathbf{S}}
\newcommand{\mT}{\mathbf{T}}
\newcommand{\mU}{\mathbf{U}}
\newcommand{\mV}{\mathbf{V}}
\newcommand{\mW}{\mathbf{W}}
\newcommand{\mX}{\mathbf{X}}
\newcommand{\mY}{\mathbf{Y}}
\newcommand{\mZ}{\mathbf{Z}}

\newcommand{\cA}{\mathcal A}
\newcommand{\cB}{\mathcal B}
\newcommand{\cC}{\mathcal C}
\newcommand{\cD}{\mathcal D}
\newcommand{\cE}{\mathcal E}
\newcommand{\cF}{\mathcal F}
\newcommand{\cG}{\mathcal G}
\newcommand{\cH}{\mathcal H}
\newcommand{\cI}{\mathcal I}
\newcommand{\cJ}{\mathcal J}
\newcommand{\cK}{\mathcal K}
\newcommand{\cL}{\mathcal L}
\newcommand{\cM}{\mathcal M}
\newcommand{\cN}{\mathcal N}
\newcommand{\cO}{\mathcal O}
\newcommand{\cP}{\mathcal P}
\newcommand{\cQ}{\mathcal Q}
\newcommand{\cR}{\mathcal R}
\newcommand{\cS}{\mathcal S}
\newcommand{\cT}{\mathcal T}
\newcommand{\cU}{\mathcal U}
\newcommand{\cV}{\mathcal V}
\newcommand{\cW}{\mathcal W}
\newcommand{\cX}{\mathcal X}
\newcommand{\cY}{\mathcal Y}
\newcommand{\cZ}{\mathcal Z}

\newcommand{\va}{\mathbf{a}}
\newcommand{\vb}{\mathbf{b}}
\newcommand{\vc}{\mathbf{c}}
\newcommand{\vd}{\mathbf{d}}
\newcommand{\ve}{\mathbf{e}}
\newcommand{\vf}{\mathbf{f}}
\newcommand{\vg}{\mathbf{g}}
\newcommand{\vh}{\mathbf{h}}
\newcommand{\vi}{\mathbf{i}}
\newcommand{\vj}{\mathbf{j}}
\newcommand{\vk}{\mathbf{k}}
\newcommand{\vl}{\mathbf{l}}
\newcommand{\vm}{\mathbf{m}}
\newcommand{\vn}{\mathbf{n}}
\newcommand{\vo}{\mathbf{o}}
\newcommand{\vp}{\mathbf{p}}
\newcommand{\vq}{\mathbf{q}}
\newcommand{\vr}{\mathbf{r}}
\newcommand{\vs}{\mathbf{s}}
\newcommand{\vt}{\mathbf{t}}
\newcommand{\vu}{\mathbf{u}}
\newcommand{\vv}{\mathbf{v}}
\newcommand{\vw}{\mathbf{w}}
\newcommand{\vx}{\mathbf{x}}
\newcommand{\vy}{\mathbf{y}}
\newcommand{\vz}{\mathbf{z}}

\newcommand{\bR}{\mathbb{R}}
\newcommand{\mx}{\mathbf{x}}
\newcommand{\mj}{\mathbf{j}}

\newcommand{\RNum}[1]{\uppercase\expandafter{\romannumeral #1\relax}}

\definecolor{Gray}{gray}{0.85}
\newcolumntype{a}{>{\columncolor{Gray}}c}
\newcolumntype{b}{>{\columncolor{white}}c}

\maketitle
\begin{abstract}
While deep learning reshaped the classical motion capture pipeline with feed-forward networks,
 generative models are required to recover fine alignment via iterative refinement.
Unfortunately, the existing models are usually hand-crafted or learned in controlled conditions, only applicable to limited domains.
We propose a method to learn a generative neural body model from unlabelled monocular videos
by extending Neural Radiance Fields (NeRFs). We equip them with a skeleton to apply to time-varying and articulated motion.
A key insight is that implicit models require the inverse of the forward kinematics used in explicit surface models.
Our reparameterization defines spatial latent variables relative to the pose of body parts and thereby overcomes ill-posed inverse operations with an overparameterization.
This enables learning volumetric body shape and appearance from scratch while jointly refining the articulated pose; all without ground truth labels for appearance, pose, or 3D shape on the input videos.
 When used for novel-view-synthesis and motion capture, our neural model improves accuracy on diverse datasets.  Project website: \href{https://lemonatsu.github.io/anerf/}{\color{magenta}\texttt{https://lemonatsu.github.io/anerf/}}.
\end{abstract}

\newcommand{\nerfparam}{\phi}
\newcommand{\nerffunc}{F_{\nerfparam}}
\newcommand{\render}{C_{\nerfparam}}
\newcommand{\nerfmap}{\Phi}

\newcommand{\jointtrans}{T(\smplpose_k,m)}
\newcommand{\jointtransinv}{T^{-1}(\smplpose_k,m)}

\newcommand{\cameramatrix}{\mP}
\newcommand{\smplpose}{\theta}
\newcommand{\smplshape}{\beta}
\newcommand{\smplspace}{\mathbb{R}^{24\times 3}}
\newcommand{\smplposemv}{\theta^\prime}
\newcommand{\cpe}{\Gamma}

\newcommand{\joint}{J}
\newcommand{\jointspace}{\mathbb{R}^{24\times 3}}

\newcommand{\jointele}{\mathbf{a}}
\newcommand{\jointelespace}{\mathbb{R}^{3}}

\newcommand{\smplbone}{\omega}
\newcommand{\smplbonespace}{\mathbb{R}^{3}}
\newcommand{\smplbonespaceNEW}{\mathbb{R}^{6}}

\newcommand{\local}[1]{\tilde{#1}}
\newcommand{\query}{\mathbf{q}}
\newcommand{\queryspace}{\mathbb{R}^{3}}
\newcommand{\querydist}{\local{\mathbf{v}}}
\newcommand{\queryrot}{\local{\mathbf{r}}}

\newcommand{\viewdir}{\mathbf{d}}
\newcommand{\viewdirspace}{\mathbb{R}^{3}}
\newcommand{\vieworig}{\mathbf{o}}

\newcommand{\density}{\sigma}
\newcommand{\rgb}{\mathbf{c}}

\section{Introduction}

Generative models have evolved from Generative Adversarial Networks (GANs) recreating images \cite{Goodfellow14,karras2019style} to neural scene representations \cite{mildenhall2020nerf,sitzmann2019srn,sitzmann2020implicit_siren} providing control and image understanding for downstream tasks via structured latent variables. %
However, most 3D models require 3D labels that cannot be crowd-sourced on natural images and require dedicated depth sensors. 
It is hence an important research problem to learn 3D representations from 2D observations, which is particularly challenging for
humans with diverse body shapes and appearances and their non-rigid motion.

Modern human motion capture techniques typically combine the advantages of discriminative and generative approaches.
A feed-forward 3D human pose estimation approach provides a rough initial estimate of human pose.
Afterward, a generative approach based on either a high-quality 3D scan of the person \cite{habermann2019TOG}, or a parametric human body model learned from laser scans \cite{alldieck2019learning} refines the estimate iteratively based on the image evidence.
Although achieving unprecedented accuracy, existing models require a low-dimensional, restrictive, shape body model or a personalized 3D scan of the user.

We introduce \emph{Articulated Neural Radiance Fields} (A-NeRF) for learning a user-specific neural 3D body model and underlying skeleton pose from unlabelled videos (see \figref{fig:teaser}).
When applied to motion capture, it alleviates the need for template models while maintaining the advantages and accuracy of current generative approaches.
A-NeRF extends Neural Radiance Fields (NeRF) \cite{mildenhall2020nerf} to work with single videos and articulated motion. NeRF parameterizes the scene implicitly as
\begin{equation}
    \nerffunc(\Gamma(\query), \Gamma(\viewdir)) \mapsto (\density,\rgb), \quad\text{ with } \density \in \R, \rgb \in \R^3, \query \in \R^3, \text{ and } \viewdir \in \R^3,
\label{eq:nerf}
\end{equation}
by chaining $\nerffunc$, a Multi-layer Perceptron (MLP), with $\Gamma$, the Positional Encoding (PE) \cite{vaswani2017attention_transformer}.
First, the PE maps the input scene point $\query$ and view direction $\viewdir$ to a higher dimensional space that enables the MLP to learn a meaningful scene representation function $\nerffunc$ that subsequently outputs the radiance $\rgb$ and opacity $\density$ at every point in space. Second, the implicitly described scene (via conditioning on query locations) is rendered via classical ray-marching from computer graphics.
The advantage of the MLP representation is that it avoids the complexity of volumetric grids \cite{lombardi2019nv}, circumvents the artifacts induced by the implicit bias of screen-space convolution \cite{nguyen2019hologan,shysheya2019texturedneuralavatar}, and, unlike surface meshes, can have flexible topology. However, the original NeRF only works for static scenes captured from dozens of calibrated cameras such that each 3D point is seen from multiple views.
\newlength\teaserscalel
\setlength\teaserscalel{0.3500\textwidth}
\newlength\teaserscaler
\setlength\teaserscaler{0.38\textwidth}
\newlength\teaserscalerR
\setlength\teaserscalerR{0.25\textwidth}
\newlength\teaserscalerRh
\setlength\teaserscalerRh{0.7\teaserscalerR}

\begin{figure}[t]
\centering%
\setlength{\fboxrule}{0pt}%
\parbox[t]{\teaserscalel}{%
\vspace{0mm}\centering%
\fbox{\includegraphics[width=\teaserscalel,trim={0 -24.5 0 -15mm}]{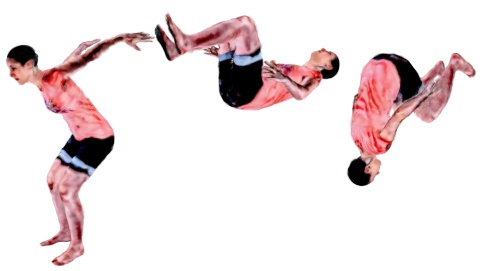}}\\%
{(a) A-NeRF body model for animation and novel-view-synthesis}%
}%
\hfill%
\parbox[t]{\teaserscaler}{%
\vspace{0mm}\centering%
\parbox[t]{\teaserscaler}{
\vspace{0mm}\centering%
\fbox{%
\includegraphics[width=\teaserscaler]%
{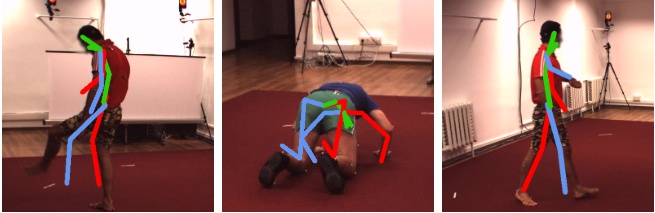}%
}\vspace{-3mm}
$\phantom{\,}\underbracket[1pt][1.0mm]{\hspace{\teaserscaler}}_%
    {\substack{\vspace{-3.0mm}\\\colorbox{white}{~~Initial pose~~}}}$\vspace{-2.5mm}
}\\%
\parbox[t]{\teaserscaler}{
\vspace{0mm}\centering%
\fbox{%
\includegraphics[width=\teaserscaler]%
{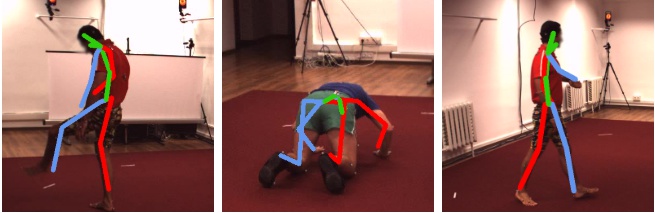}%
}\vspace{-3mm}
$\phantom{\,}\underbracket[1pt][1.0mm]{\hspace{\teaserscaler}}_%
    {\substack{\vspace{-3.0mm}\\\colorbox{white}{~~Refined pose~~}}}$%
\\%
{(b) 3D pose refinement for real images}%
}%
\\%
}
\hfill%
\parbox[t]{\teaserscalerR}{%
\vspace{0mm}\centering%
\parbox[t]{\teaserscalerRh}{
\vspace{0mm}\centering%
\fbox{%
\includegraphics[width=0.28\teaserscalerR,trim={0 0 0 -3mm}]%
{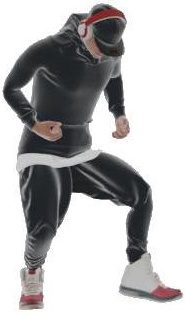}%
}\vspace{-3mm}
$\underbracket[1pt][1.0mm]{\hspace{\teaserscalerRh}}_%
    {\substack{\vspace{-3.0mm}\\\colorbox{white}{~~Input view~~}}}$\vspace{-2.5mm}
}\\%
\parbox[t]{\teaserscalerRh}{
\vspace{0mm}\centering%
\fbox{%
\includegraphics[width=0.215\teaserscalerR]%
{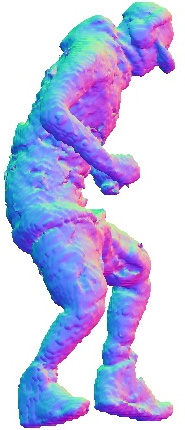}%
}\vspace{-3mm}
$\underbracket[1pt][1.0mm]{\hspace{\teaserscalerRh}}_%
    {\substack{\vspace{-3.0mm}\\\colorbox{white}{~~3D rec.~~}}}$%
\\%
{(c) Body shape}%
}%
\\%
}
\captionof{figure}%
{
Our A-NeRF %
jointly learns a neural body model of the user and works with diverse body poses (left) while also refining the initial 3D articulated skeleton pose estimate from a single or, if available, multiple views without tedious camera calibration (center). Underlying is a template-free neural representation (right) and skeleton-based embedding coupled with volume volumetric rendering. \tbi{Real faces and their reconstructions are blurred in all figures for anonymity.}
\label{fig:teaser}
}
\end{figure}

Our conceptual contribution lies in learning a neural latent representation relative to an articulated skeleton. While explicit models such as the popular SMPL body model~\cite{loper2015smpl} deform a surface via forwards kinematics, the implicit form of A-NeRF makes us re-think how skeletons can be integrated---implicit networks require the inverse transformation from 3D world coordinates to the reference skeleton, a significantly harder task that has not been fully explored.
Our core technical novelty is to come up with and evaluate different parameterizations of $\Gamma(\query), \Gamma(\viewdir)$ in Eq.~\ref{eq:nerf} to create local coordinates relative to the articulated skeleton. Since a point in 3D world coordinates cannot uniquely be associated with a body part, we resolve the mentioned ill-posed inverse problem by overparameterizing with one embedding per bone.
This embeds domain knowledge of how humans move and provides a common frame for the neural network to combine body shape and appearance constraints across the entire captured sequence (see Figure~\ref{fig:overview}). 

We demonstrate that all our contributions together enable learning of a neural body model from monocular video, requiring only rough 3D pose estimates for initialization, that reaches a level of detail previously only attained with parametric surface models or multi-view approaches \cite{Rhodin19}.

\paragraph{Scope.} We apply the model to motion capture, character animation, and appearance and motion transfer and demonstrate that the pose refinement improves on existing monocular skeleton reconstruction. We learn in the transductive setting, for a specific target video that is known at training time but has no ground truth. A-NeRF enables novel view synthesis of dynamic motions, with plausible however non-physical illumination. Additional steps are needed to enable relighting applications. %

\paragraph{General impact.} Building a self-supervised approach for personalized human body modelling promises to be more inclusive to people and activities that are not well represented in supervised datasets. %
However, it bears the risk that 3D models of people are created without consent.
We urge users to only use datasets collected for developing and validating motion capture algorithms.

\section{Related Work}
Our approach builds upon and is related to the following work on human pose and shape estimation, human modeling, and neural scene representations \cite{tewari2020neuralrendering}.

\paragraph{Discriminative Human Pose Estimation.} While feed-forward estimation of the 3D joint positions \cite{Li_2020_CVPR,OriNet2018,martinez2017simple,mpi_3dhp,Park2016,Pavlakos2017,lcrnet2017,sun2017compositional,Tome_2017_CVPR,Xu_2020_CVPR,zhou2019hemlets} or joint angles and bone lengths~\cite{shi2020motionet,zhou_deep_kinematic_arxiv16} of the skeleton is highly accurate, %
such discriminative estimates are prone to misalignment when overlayed onto the input image due to the generalization gap. The skeleton pose can be refined to better match the 2D pose estimates, but this usually leads to larger errors in 3D \cite{VNect_SIGGRAPH2017,XNect_SIGGRAPH2020}.
We use~\cite{kolotouros2019learning_spin} for initializing skeleton pose and combine it with a neural body model.

\paragraph{Surface-based Generative Body Models.}
These are obtained by either constraining template meshes via deformation energies \cite{habermann2019TOG,xu2018monoperfcap} or learning parametric human body models from a large collection of laser scans \cite{balan2007detailed,choutas2020smplx,loper2015smpl}. Their low-dimensional parameters constrain the space of plausible human shapes and motions. This enables real-time reconstructions from single images \cite{bogo2016keep_smplify,Guan2009}, detailed texturing and displacement mapping~\cite{alldieck2018detailed,Alldieck_2019_ICCV}, and alleviates manual rigging \cite{Alldieck18}. It also enables optimization within the bounds of the learned prior ~\cite{dong2020motion,guler2019holopose,Lassner:UP:2017} and weak-supervision when integrated in a differentiable form \cite{liu2019soft} into neural training processes \cite{alldieck2019learning,kanazawa2018hmr,kolotouros2019learning_spin,omran2018neural,pavlakos2018humanshape,tung2017self}. 
Closest to our approach in this category are the model fitting methods by \cite{Zuffi:ICCV:2019} that textures and geometrically refines an untextured parametric quadruped model to zebra images and to \cite{xiang2019monocular} that uses optical flow to refine human pose. Although in a similar setting, our surface-free neural body model and volumetric rendering is fundamentally different to their textured triangle mesh that is rigged with forward kinematics and needs to be obtained a-priori.

\paragraph{Implicit Body Models.}
A-NeRF bears close similarities with body models defined implicitly in terms of level-sets \cite{stoll_fast_iccv2011} and density of a sum of Gaussians \cite{huang2020arch,rhodin2015versatile,rhodin2016general} that are used for refining human pose, shape, and appearance via differentiable ray-tracing. However, sum of Gaussians and other primitives only provide rough approximations.

\paragraph{Neural Scene Representations.}
Recent neural scene representations learn low-dimensional non-linear representations of meshes \cite{lombardi2018DAM,thies2019neuraltex}, point clouds \cite{aliev2019neuralpoint,meshry2019neuralrerendering,wiles2020synsin}, sphere sets \cite{lassner2020pulsar}, and dense volumetric grids \cite{lombardi2019nv,sitzmann2019deepvoxels}.
Their respective geometric output representations enable rendering with classical rendering techniques but have limited expressiveness, e.g., due to the fixed connectivity of a surface mesh and large memory footprint of discretized volumes.
This limitation is overcome by using unconstrained MLPs \cite{mildenhall2020nerf} paired with positional encoding \cite{tancik2020fourfeat}
to characterize an arbitrary point in 3D space. Common are surface definitions via level-sets of the MLP that are rendered with sphere tracing \cite{sitzmann2019srn} and density representations rendered with ray-marching \cite{mildenhall2020nerf}. The rendering step is required for maximum likelihood estimation, to define a likelihood over observable variables---real images---while learning a 3D model. The rendering can be learned too \cite{nguyen2019hologan,Rhodin18b,Rhodin19} but usually leads to inconsistencies, particularly when training data is scarce. Some concurrent works also use neural scene representations for refining camera motion~\cite{yen2020inerf}, video reenactment~\cite{pumarola2020dnerf}, and facial models~\cite{gafni2021dynamic,Gao-freeviewvideo}. Orthogonal to these works, our A-NeRF learns an articulated body model from estimated poses and uncalibrated cameras.

Closely related to ours is the NASA surface body model~\cite{deng2019nasa} that also defines an implicit function as the minimum of individual implicit functions that are rigidly attached to the bones of a skeleton, each conditioned on the entire human pose to model dependencies and learned from 3D scans. By contrast, we learn a volumetric model instead of a surface model and include appearance and rendering. Even more similar is the recent NeuralBody~\cite{peng2020arxiv_neuralbody} representation, which combines a NeRF with a surface body model and underlying skeleton. By contrast to both approaches, we do not require surface supervision or initialization, condition pose differently, and refine pose, which 
 enables us to learn from single videos in unconstrained environments.

\section{Formulation}
\begin{figure*}[t]
\centering
\includegraphics[width=\linewidth,trim=80 0 0 0,clip]{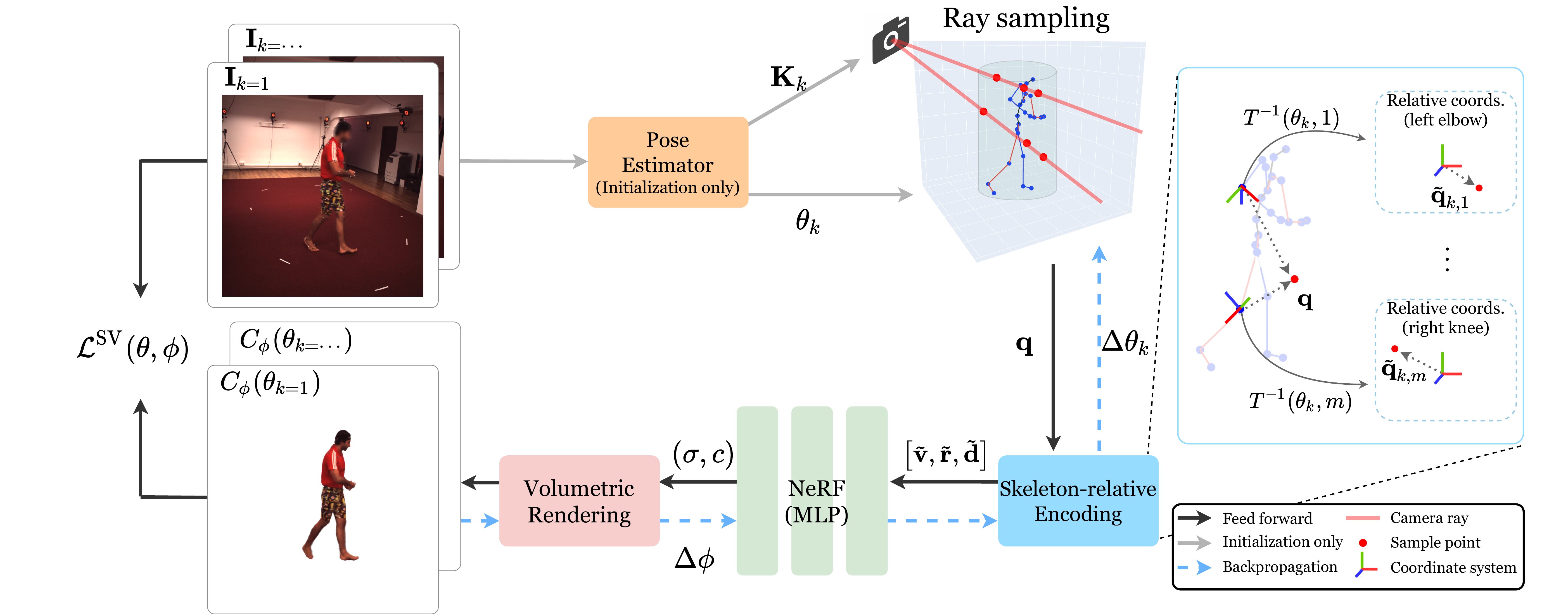}
\caption{\tb{Overview.} A-NeRF is a generative model that can be rendered and optimized on a photometric loss $\cL^\text{SV}$ (white).
First, the skeleton pose is initialized with an off-the-shelf estimator (orange). 
Second, this pose is refined via a skeleton-relative embedding (blue) that, when fed to NeRF (green), drives the implicit body model that is rendered by ray-marching (red).
A key property of the skeleton-relative embedding is that a single 3D query location maps to an overcomplete reparametrization, with the same point represented relative to each skeleton bone (right).
}
\vspace{-3mm}
\label{fig:overview}
\end{figure*}

\paragraph{Objective.} Given a sequence $[\mI_k]^{N}_{k=1}$ of N images $\mI_k\in\bR^{H\times W\times3}$ stemming from one or several videos of the same person, our goal is to simultaneously estimate the time-varying skeleton poses $[\smplpose_k]^{N}_{k=1}$ and learn a detailed body model. Our A-NeRF body model $\render$ is parametrized by neural network parameters $\nerfparam$ that define volumetric shape and color while the skeleton captures motion over time. Figure~\ref{fig:overview} gives an overview of this generative model. It enables a rendering of the virtual body model in unseen poses and optimizes its parameters $\smplpose$ and $\nerfparam$ on the image reconstruction objective
\begin{equation}
\cL^\text{SV}(\smplpose, \nerfparam) = \sum_k 
\underbrace{\norm{\render(\smplpose_k)- \mI_k}_1             \vphantom{\norm{\frac{\partial \smplpose_k}{\partial t}}_2^2}}_\text{data term} + 
\underbrace{\lambda_\smplpose d(\smplpose_k -\hat{\smplpose}_k)   \vphantom{\norm{\frac{\partial \smplpose_k}{\partial t}}_2^2}  }_\text{pose regularizer} +
\underbrace{\lambda_t \norm{\frac{\partial^2 \smplpose_k}{\partial t^2}}_2^2    }_\text{smoothness prior} .
\label{eq:objective}    
\end{equation}
with the influence of all three terms balanced by hyperparameters $\lambda_t$ and $\lambda_\smplpose$.
The data term measures the distance between the images generated by $\render$ and the input image with the L1 distance.
The pose regularizer encourages the solution to stay close to an initial pose estimate $\hat{\smplpose}$ obtained from an off-the-shelf predictor~\cite{kolotouros2019learning_spin}, tolerating small shifts up to $\epsilon=0.01$ with $d(x) = \min(\norm{x}_2^2 - \epsilon, 0)$.
Lastly, the smoothness prior penalizes acceleration $\frac{\partial^2 \smplpose_k}{\partial t^2}$ between poses of consecutive frames.
Minimizing Eq.~\ref{eq:objective} can be seen as maximizing a corresponding probabilistic model, with the quadratic energy terms being the log-likelihoods of Gaussian distributions. Our focus is on formalizing the neural body model. For simplicity, we continue to write equations in terms of the objective functions used during inference with stochastic gradient descent. 

\newlength\relativefscale
\setlength\relativefscale{0.158\textwidth}
\newlength\inputrepscale
\setlength\inputrepscale{0.68\relativefscale}
\newlength\repleftscale
\setlength\repleftscale{0.64\linewidth}
\newlength\reprightscale
\setlength\reprightscale{0.35\linewidth}

\begin{figure}
    \setlength{\fboxrule}{0pt}%
    \begin{minipage}{\repleftscale}
    \centering
    \parbox[t]{\relativefscale}{%
    \vspace{0mm}\centering%
    \fbox{\includegraphics[width=\relativefscale, trim=140 130 140 140, clip=true]{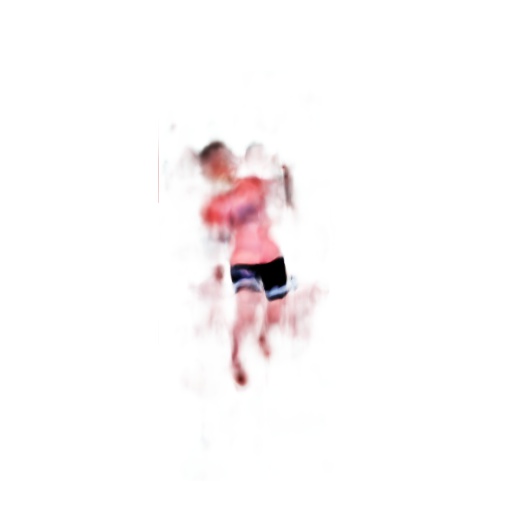}}\\%
    {\footnotesize (a) NeRF}
    }\hfill%
    \parbox[t]{\relativefscale}{%
    \vspace{0mm}\centering%
    \fbox{\includegraphics[width=\relativefscale, trim=140 130 140 140, clip=true]{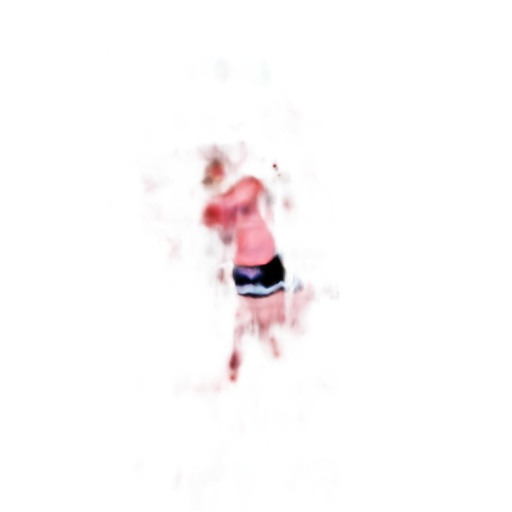}}\\%
    {\footnotesize (b) Rotate $24^\circ$}
    }\hfill%
    \parbox[t]{\relativefscale}{%
    \vspace{0mm}\centering%
    \fbox{\includegraphics[width=\relativefscale, trim=140 130 140 140, clip=true]{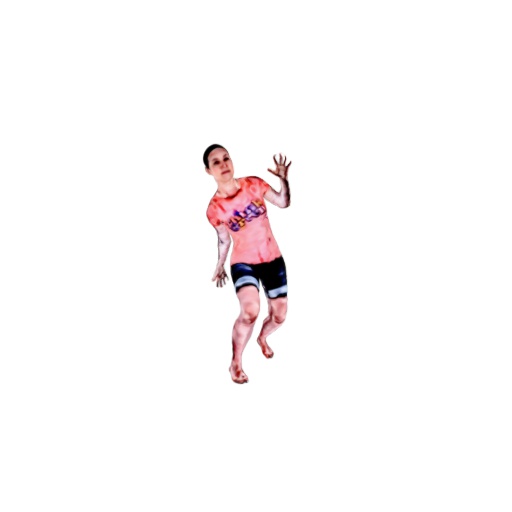}}\\%
    {\footnotesize (c) Ours}
    }\hfill%
    \parbox[t]{\relativefscale}{%
    \vspace{0mm}\centering%
    \fbox{\includegraphics[width=\relativefscale,trim=140 130 140 140, clip=true]{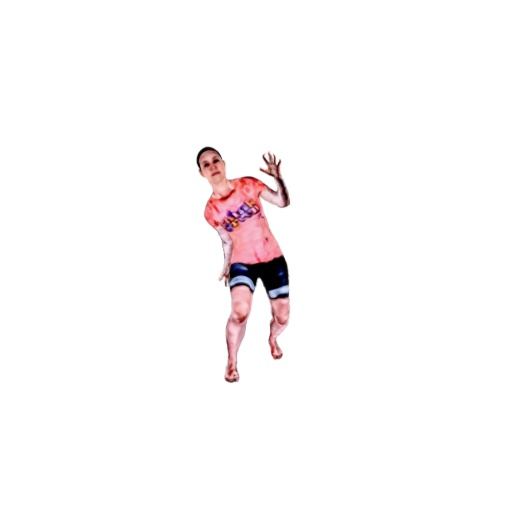}}\\%
    {\footnotesize (d) Rotate $24^\circ$}
    }%
    \end{minipage}\hfill%
    \begin{minipage}{\reprightscale}
    \centering
    \parbox[t]{\inputrepscale}{%
    \vspace{0mm}\centering%
    \fbox{\includegraphics[width=\inputrepscale,trim=70 80 40 80,clip=true]{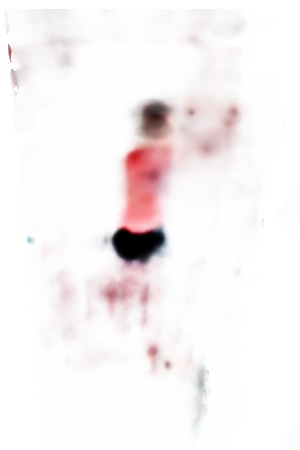}}\\%
    {\footnotesize (e) w/o $\querydist,\queryrot,\local{\viewdir}$}
    }\hfill%
    \parbox[t]{\inputrepscale}{%
    \vspace{0mm}\centering%
    \fbox{\includegraphics[width=\inputrepscale,trim=70 80 40 80,clip=true]{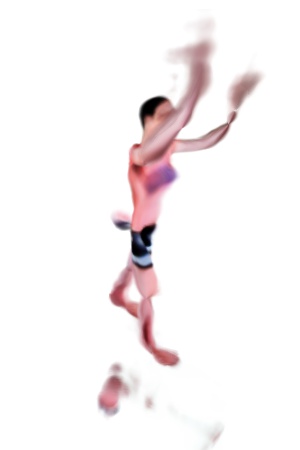}}\\%
    {\footnotesize (f) w/o $\querydist$}%
    }\hfill%
    \parbox[t]{\inputrepscale}{%
    \vspace{0mm}\centering%
    \fbox{\includegraphics[width=\inputrepscale,trim=70 80 40 80,clip=true]{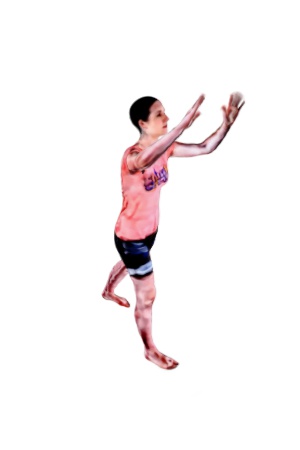}}\\%
    {\footnotesize (g)  Ours}
    }%
\end{minipage}
\caption{\tb{Importance of our skeleton-relative encodings.} %
The original NeRF breaks (a) when training on a diverse set of poses and (b) further degrades when the poses are rotated. Even if (e) conditioned directly on $\smplpose_k$, the NeRF still suffers from artifacts due to the complexity and ambiguity of human articulation. 
With our skeleton-relative encoding (f, g), the geometry for the subject is consistent under rotation, and the quality is greatly improved, with the full model working best.}
\label{fig:input-rep}
\end{figure}

\subsection{NeRF and A-NeRF Image Formation Model}
\label{sec:NeRF}
Instead of modeling the scene as a collection of triangles or other primitives, we define the human implicitly by a neural network as a function (Eq.~\ref{eq:nerf}) defined over all possible 3D points and view directions in space \cite{mildenhall2020nerf}. Similar to NeRF, we render the image of the human subject via ray marching
\begin{align}
\render(u,v;\smplpose_k) &= \sum^Q_{i=1} T_i(1-\exp(-\density_i\delta_i))\rgb_i,\quad T_i=\exp\left(-\sum_{j=1}^{i-1}\density_j\delta_j\right),%
\label{eq:image fromation}
\end{align}
with $(u,v)$ the 2D pixel location on the image, $i$ the index to 3D query positions $\query_i$ sampled along $\viewdir$ and $\delta_i$ the distance to neighboring samples---a constant if samples would be taken at regular intervals. The $T_i$ is the accumulated transmittance for the ray traveling from the near plane to $\query_i$---the fraction of light reaching the sensor from sample point $i$. The $\rgb_i$ is the light color emitted or reflected at $i$. The final pixel color is the sum over all $Q$ samples, with the last sample taking the special role of the background. The background color is easily inferred as the median pixel color over the entire video for static camera setups. The ray direction $\viewdir = \mK_k^{-1} (u,v)$ is computed using the estimated camera intrinsics~\cite{kolotouros2019learning_spin}. In the following, we introduce our skeleton parametrization $\theta_k$ and how to use it to effectively model dynamic articulated human motion.

\subsection{Articulated Skeleton Pose Model}
\label{sec:data-prepreoc}
Our skeleton representation encodes the connectivity and static bone lengths via a rest pose of 3D joint locations. Dynamics are modeled with per-frame skeleton poses $\theta_k$, which define an affine transformation $\jointtrans$ for each bone $m$. Specifically, $\jointtrans$ maps a 3D position $\subkm{\vp} \in\jointelespace$ in the $m$-th local bone coordinates to world coordinates $\vq \in\jointelespace$ using homogeneous coordinates,
\begin{align}
\hmvec{\vq} = \jointtrans \hmvec{\subkm{\vp}},
\label{eq:bone transformation}
\end{align}
where subscript $_{k,m}$ indicates that a variable is related to the $m$-th joint of image $\mI_k$. Conversely, $\jointtrans^{-1}$ maps world to local bone coordinates. Note that our skeleton is equivalent to SMPL~\cite{loper2015smpl} and others, but without their parametric surface model, and can therefore be initialized with any skeleton pose estimator. We include more details of our skeleton representation in the supplementary.

\subsection{A-NeRF Skeleton-Relative Encoding}
\newlength\cutofflinescale
\setlength\cutofflinescale{0.24\linewidth}
\newlength\cutoffscalel
\setlength\cutoffscalel{0.46\cutofflinescale}
\newlength\cutoffscaler
\setlength\cutoffscaler{1\cutofflinescale}
\begin{wrapfigure}[8]{r}{\cutofflinescale}
\centering%
\setlength{\fboxrule}{0pt}%
\parbox[t]{\cutoffscaler}{%
\vspace{0mm}\centering%
\vspace{-0.6cm}
\fbox{\includegraphics[width=\cutoffscaler,trim=0 10 0 0,clip]{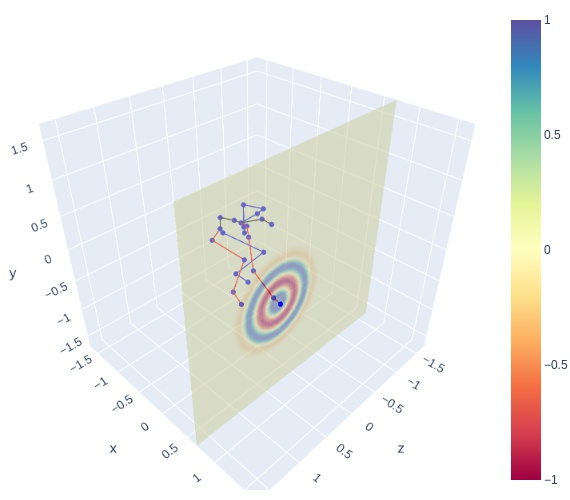}}%
}%
\label{fig:example-cutoff}
\end{wrapfigure}%
Our core contribution is to transform the query locations $\vq$ and view direction $\viewdir$ relative to the skeleton before determining the color and opacity at that transformed point via NeRF. It is a form of reparameterization that explicitly incorporates domain knowledge of how the human body parts are linked and transformed relative to each other. Intuitively, our implicit formulation turns explicit models, such as SMPL~\cite{loper2015smpl}, on its head. Instead of deforming the output surface via skinning, the query location is mapped in the inverse direction to local bone-relative coordinate systems before processing through the NeRF network.
Our final model uses a combined encoding $\ve_k = [h(\querydist_k)\Gamma(\querydist_k),\queryrot_k,h(\querydist_k)\Gamma(\viewdir_k)]$ as input to the NeRF $\nerffunc$. Note that the skeleton embedding introduces the desired time dependency, denoted by subscript $k$.
The inlet shows our most crucial contribution, the relative distance encoding $\querydist_k$ followed by PE with \textbf{Cutoff} to reduce the influence of irrelevant bones.
We derive the components of our encoding $\ve_k$ and the other alternatives below.

\begin{itemize}[leftmargin=0.7cm]%
\item \textbf{Reference Pose Encoding} One could compensate motion by attaching the query $\vq$ in world coordinates at frame $k$ to the closest bone $m$ and transforming it with
\begin{equation}
    \va_k = T(\theta_0, m)\jointtransinv \vq.
\end{equation}
This puts the query relative to the bone $m$ as in frame $k$ but with the skeleton in rest pose $\theta_0$. NeRF could then learn without change in the 3D space of the rest pose as done before for surfaces~\cite{taylor2012vitruvian}. However, this cannot capture non-rigid pose-dependent effects, such as muscle bulging, and has ambiguities when $\vq$ is at equal distance to two bones.

\item \textbf{Bone-relative Position (Rel. Pos.)} To remove these ambiguities and the ill-posed association to a single part, we map $\vq$ relative to each bone $m$ with,
\begin{equation}
\local{\query}_k= \veclistki{\local{\query}}{24} \text{ and } \subkm{\local{\query}} = \jointtransinv \vq.
\end{equation}
The resulting individual bone coordinates are well suited to model the overwhelmingly rigid motion of the corresponding body part. Moreover, the overparameterization of position by concatenating all local encodings enables learning when complex interactions are necessary.
However, such an embedding for all bones increases the dimensionality by an order of magnitude.

\item\textbf{Relative Distance (Rel. Dist.)} 
Much simpler to compute are distances from $\query$ to all bones $m$,
\begin{align}
\querydist_k=\veclistki{\querydist}{24}, \text{ with } \subkm{\querydist} = \norm{\subkm{\local{\query}}}_2\in{\bR}.
\end{align}
This radial encoding is used in our final model in favor of $\local{\query}$ because it naturally captures spherically shaped limbs, is lower-dimensional, and thereby improves reconstruction accuracy.

\item \textbf{Relative Direction (Rel. Dir.)}
Since the distance encoding is invariant to direction, we additionally obtain the direction vector to capture the orientation information~of~$\query$,
\begin{equation}
\queryrot_k=\veclistki{\queryrot}{24},\quad\subkm{\queryrot}=\frac{\subkm{\local{\query}}}{\norm{\subkm{\local{\query}}}_2}\in{\bR^3}.   
\end{equation}
Note that by contrast to all other embeddings, this direction encoding did not profit from subsequent PE. We therefore pass it directly into $\ve_k$. 

\item \textbf{Relative Ray Direction (Rel. Ray.)}
NeRF models the illumination effects in a static 3D scene using the position and view direction. By contrast, our goal is to learn a body model that produces plausible colors with dynamic skeleton poses. Therefore, we transform $\viewdir$ to obtain $\local{\viewdir}$, the outgoing ray direction relative to each bone, similar to query position,
\begin{equation}
\local{\viewdir}_k=\veclistki{\local{\viewdir}}{24},\quad\subkm{\local{\viewdir}} = [\jointtransinv]_{3\times 3} \viewdir\in{\bR^3},
\label{eq:world to bone rot}
\end{equation}
with $[\jointtransinv]_{3\times 3}$ the rotational part of the bone-to-world transformation $\jointtransinv$. Following concurrent works~\cite{martin2020arxiv_nerfiw,peng2020arxiv_neuralbody}, we also optimize an appearance code for each image to handle dynamic
light effects. The combination of $\local{\viewdir}$ and the per-image code enables A-NeRF to approximate the light effects in $\mI_k$ plausibly. See the supplemental material for detailed discussions on modeling view-dependent effects in our setting.
\item \textbf{Cutoff.}
We desire a local embedding where points should not be influenced by all but only nearby bones. To this end, we introduce a windowed version of positional encoding by multiplying the encoding with respect to bone $m$ by $h(\subkm{\querydist})=1-S(\tau(\subkm{\querydist}-t))$, with $S$ the sigmoid step function, $t$ the cutoff point, $\tau$ the sharpness. This leaves queries unaffected by distant bones. 
\end{itemize}
Our embedding choice of $\ve_k = [h(\querydist_k)\Gamma(\querydist_k),\queryrot_k,h(\querydist_k)\Gamma(\viewdir_k)]$ has the advantage of being invariant to the global shift and rotation of the person and preserves the piece-wise rigidity of articulated motion while still allowing for pose-dependent deformation (see~\figref{fig:input-rep}). 
In addition to $\ve_k$, we also consider other embedding variants. See the supplementary and Section~\ref{sec:evaluation} for a detailed discussion.

\section{Evaluation}
\label{sec:evaluation}
We performed experiments to validate that A-NeRF learns accurate body models and poses, with fewer assumptions (single view, uncalibrated, and w/o a parametric surface model) than the related works. This makes it applicable to fine-grained pose refinement that improves the estimates of state-of-the-art methods. 
The supplements provide the implementation details, additional comparisons and ablation studies.
\paragraph{Inference and Implementation Details}

Our A-NeRF model is learned without supervision on a single or multiple videos of the same person. Camera intrinsics, bone lengths for setting $\va_m$, and pose $\theta_k$ are initialized with~\cite{kolotouros2019learning_spin} for every frame $k$. These poses are then optimized on objective Eq.\ref{eq:objective}, alongside the generative A-NeRF model. See supplementary for more details.

\paragraph{Datasets.}
We evaluate on the following benchmarks, and additionally on synthetic data created from \textbf{SURREAL} and \textbf{Mixamo} characters, which are listed in the supplement.%
\begin{itemize}[leftmargin=0.7cm]%
\item \textbf{Human 3.6M~\cite{ionescu_human36_pami14}} The dataset consists of 5 training and 2 testing subjects (S9/S11) with ground truth 3D joint locations. We follow two widely adopted test protocols denoted as Protocol \RNum{1}~\cite{kanazawa2018hmr,kocabas2020vibe,kolotouros2019learning_spin} and Protocol \RNum{2}~\cite{nibali20193d,sun2017compositional}, in which we evaluate 14/17-joint estimation error on every  $5^{th}/64^{th}$ frame of the test videos, respectively. See the supplement for details.
\item \textbf{MPI-INF-3DHP~\cite{mpi_3dhp}} This dataset is a standard benchmark for human pose estimation. It consists of 4 indoor and 2 outdoor subjects with challenging human poses. The number of frames per subject range from 276 to 603. %
\item \textbf{MonoPerfCap~\cite{xu2018monoperfcap}} The dataset consists of human performance video captured with a monocular camera in both indoor and outdoor settings. We use two subjects, Weipeng\_outdoor and Nadia\_outdoor, for our qualitative experiments. The two subjects have 1151 and 1635 frames, respectively, for training. 
\end{itemize}

\paragraph{Pose Metrics.} We report the PA-MPJPE metric, the Euclidean distance between Procrustes-aligned (PA) predictions and ground truth 3D joint position averaged over all frames and joints of the test set. The PA alignment in scale and orientation is essential for comparing approaches that do not assume knowledge of the ground truth calibration and are, hence, ill-posed to the factors that the alignment removes. Following prior work~\cite{kolotouros2019learning_spin,kocabas2020vibe,mpi_3dhp}, we also report percentange of correct keypoints (PCK) for MPI-INF-3DHP; the percentage of joints that lie within a distance of 150mm to the ground truth.

\paragraph{Visual Metrics.} We quantify the visual quality on MonoPerfCap and Human 3.6M datasets by training on a subset and testing on a held-out test set of the same character. Image quality is quantified via the PSNR and SSIM of the rendering compared with the reference image within the character bounding boxes. Because no ground truth pose is available in the required skeleton format on these datasets, we train our model once on the entire dataset to get reliable skeleton pose estimates as pseudo ground truth, and a second time with a part withheld to learn the body model for visual quality evaluation. Since MonoPerfCap has one sequence per actor, we exclude the last 20\% of each video. For Human 3.6M we exclude entire actions, namely \emph{Geeting-1,2}, \emph{Posing-1,2} and \emph{Walking-1,2}.
The body model is then transferred to the held-out portion by using the pseudo ground truth poses as the driving motion. 
Thereby, we can still test the generalization of different models to new poses and viewpoints, irrespective of the underlying skeleton model provided in the dataset. 

\newlength\qfpscale
\setlength\qfpscale{0.0966\linewidth}

\begin{figure*}[h]
\setlength{\fboxrule}{0pt}%
\parbox[t]{\qfpscale}{
\centering
\fbox{\includegraphics[width=\qfpscale,trim=230 300 530 340,clip]{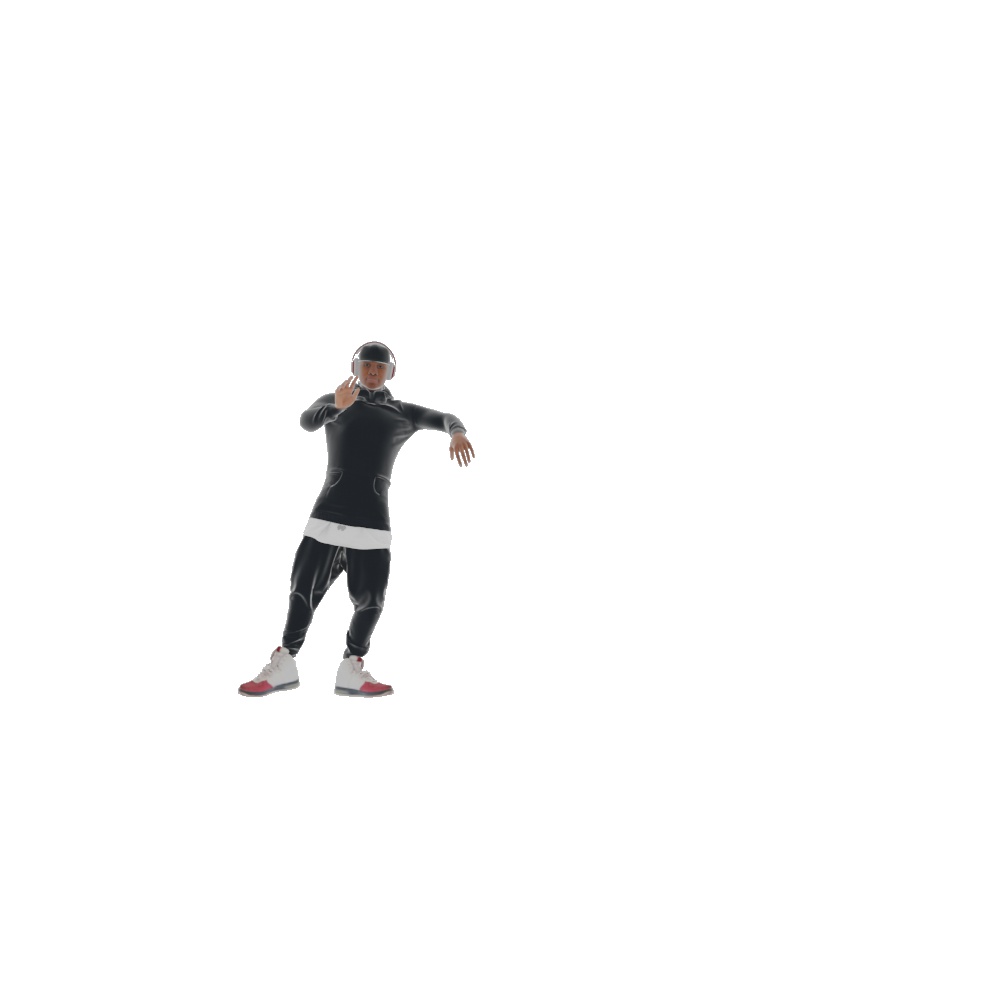}}\\%
{\small Reference}\\%
\vspace{17mm}
{\small Novel view\\$(\rightarrow)$\\Mixamo\\``James''}
}%
\hfill%
\parbox[t]{\qfpscale}{
\centering %
\fbox{\includegraphics[width=\qfpscale,trim=400 330 385 350,clip]{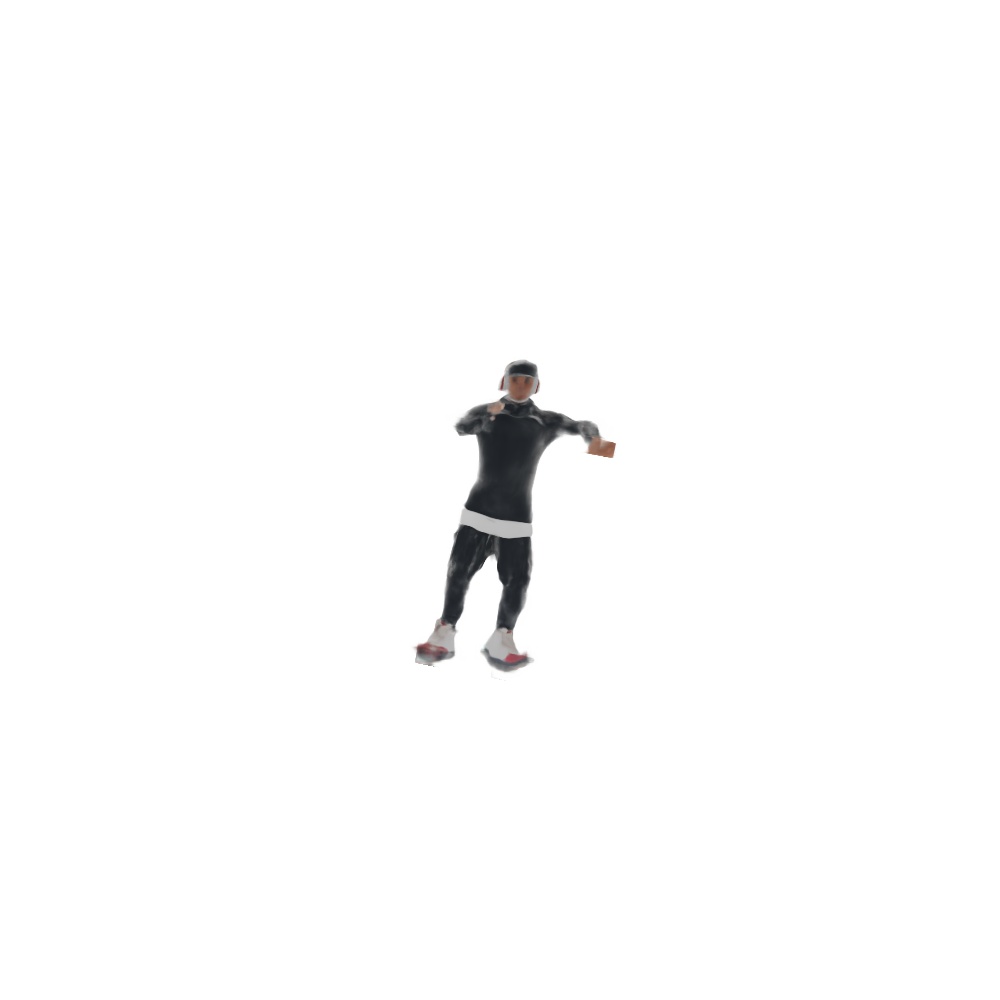}}\\%
\fbox{\includegraphics[width=\qfpscale,trim=403 350 393 381,clip]{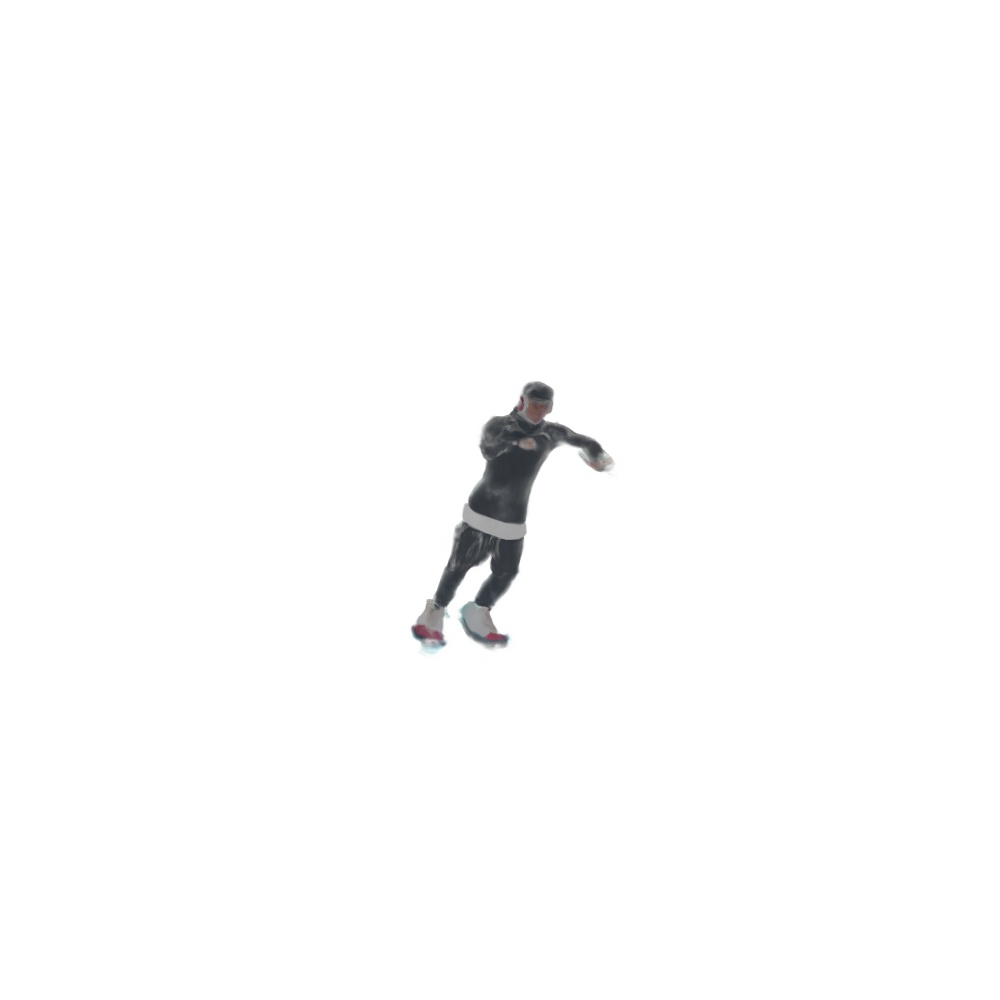}}\\%
\fbox{\includegraphics[width=\qfpscale,trim=400 325 391 355,clip]{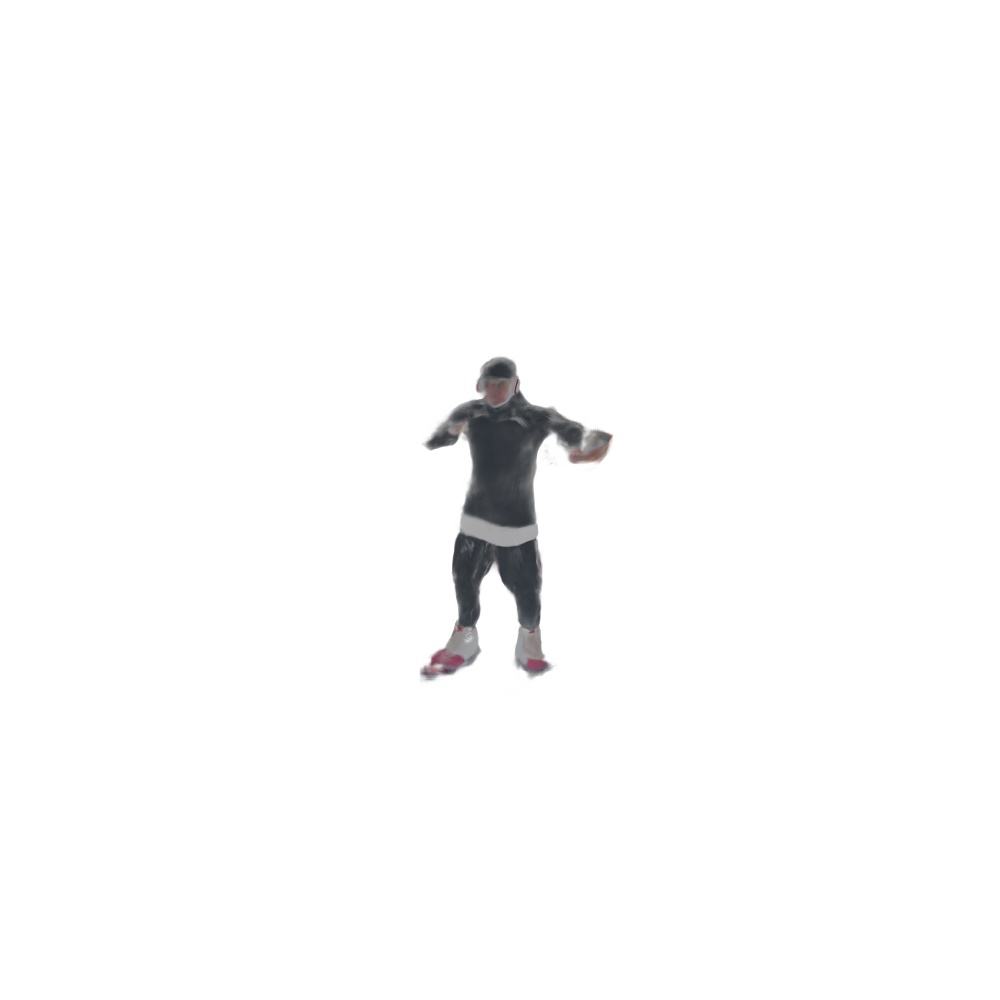}}\\%
{\small NeuralBody}%
}
\hfill%
\parbox[t]{\qfpscale}{
\centering
\fbox{\includegraphics[width=\qfpscale,trim=400 330 385 350,clip]{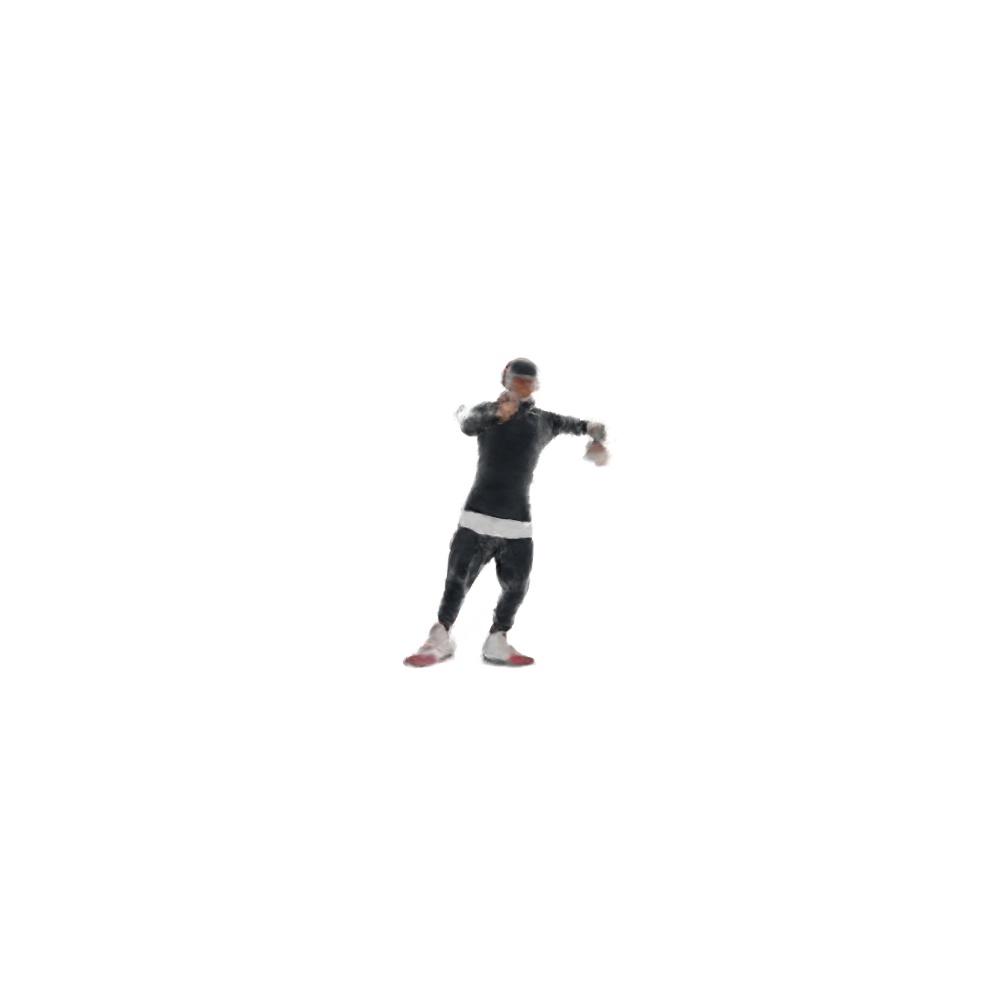}}\\%
\fbox{\includegraphics[width=\qfpscale,trim=398 345 388 376,clip]{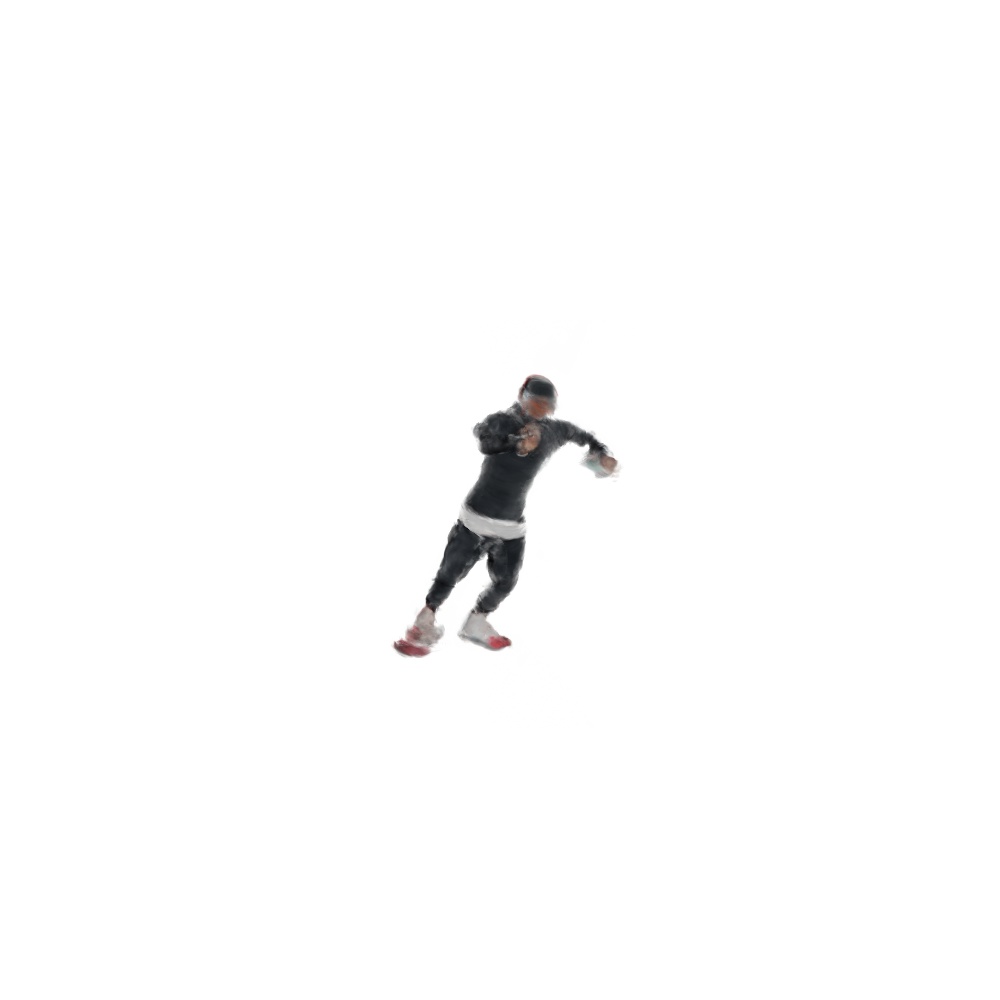}}\\%
\fbox{\includegraphics[width=\qfpscale,trim=396 318 387 347,clip]{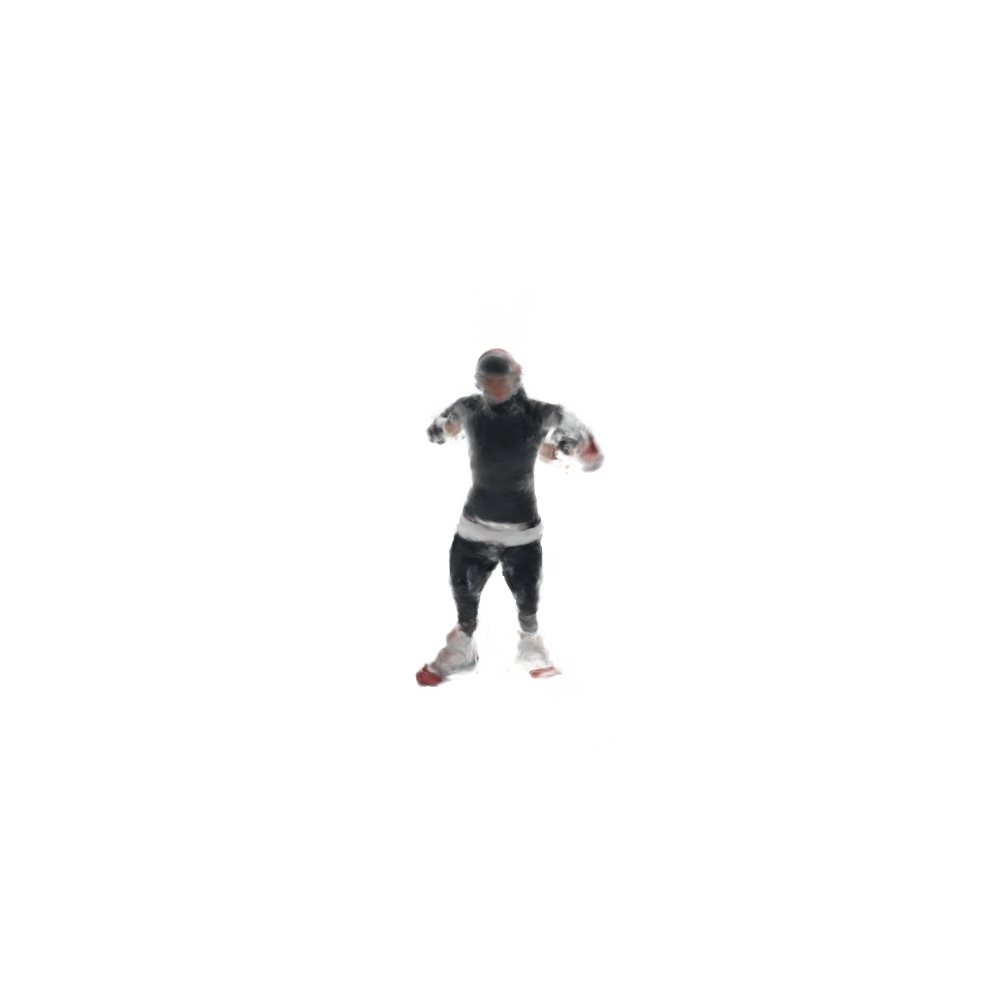}}\\%
{\small Ours w/o rf.}%
}
\hfill%
\parbox[t]{\qfpscale}{
\centering
\fbox{\includegraphics[width=\qfpscale,trim=394 330 384 350,clip]{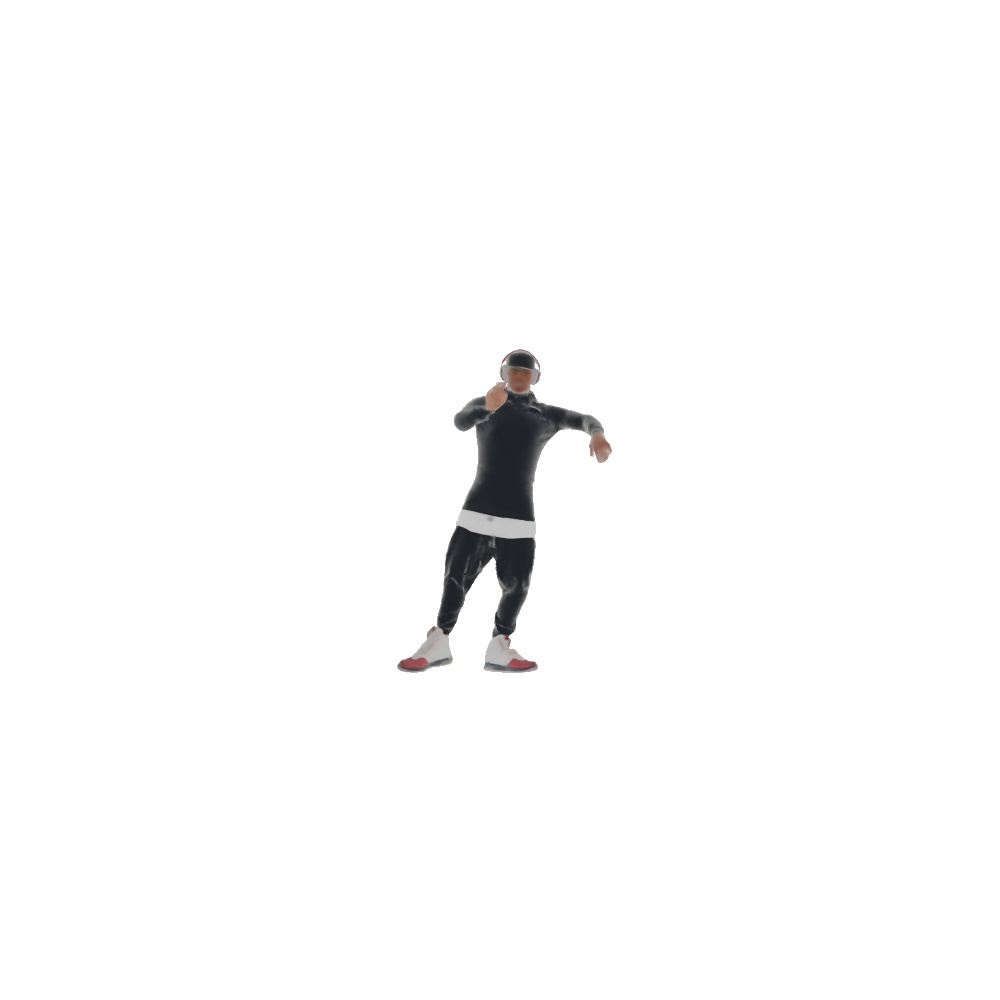}}\\%
\fbox{\includegraphics[width=\qfpscale,trim=398 350 388 371,clip]{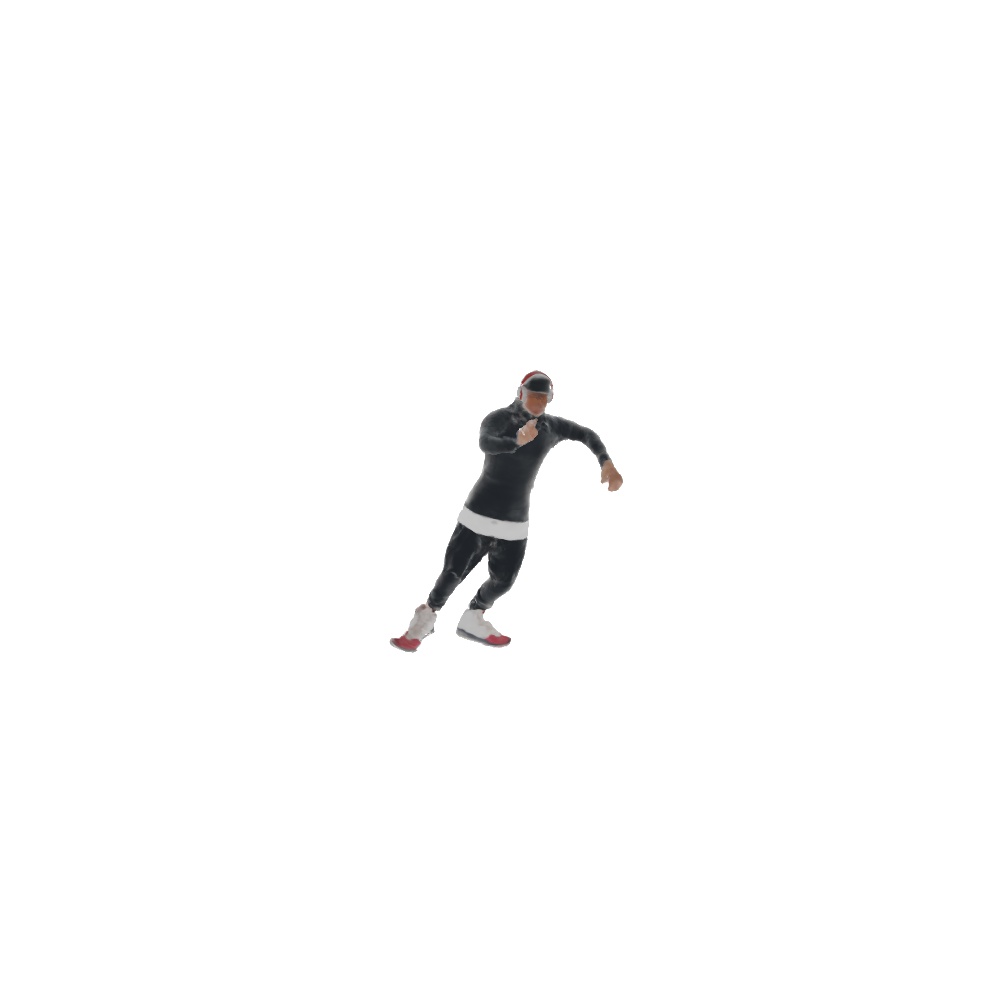}}\\%
\fbox{\includegraphics[width=\qfpscale,trim=396 320 387 345,clip]{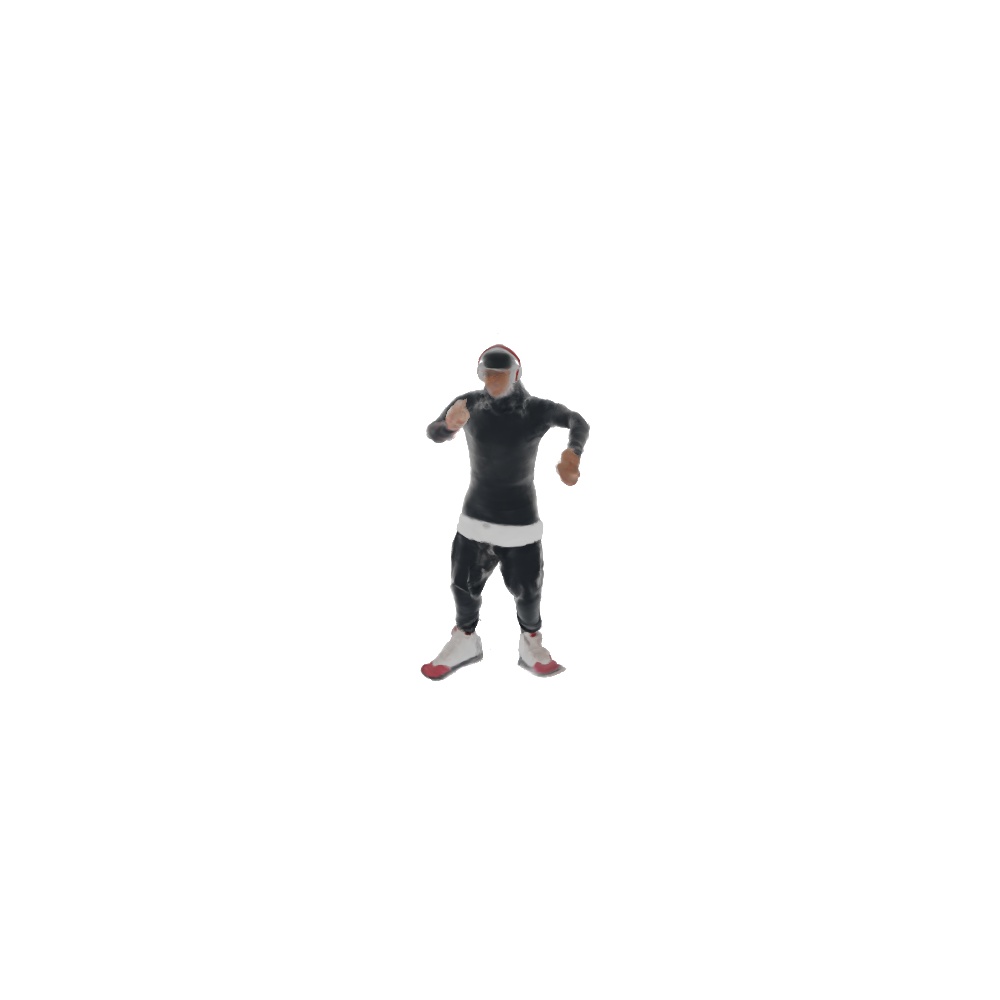}}\\%
{\small Ours}%
}%
\hfill%
\parbox[t]{\qfpscale}{
\centering
\fbox{\includegraphics[width=\qfpscale,trim=868 220 547 105,clip]{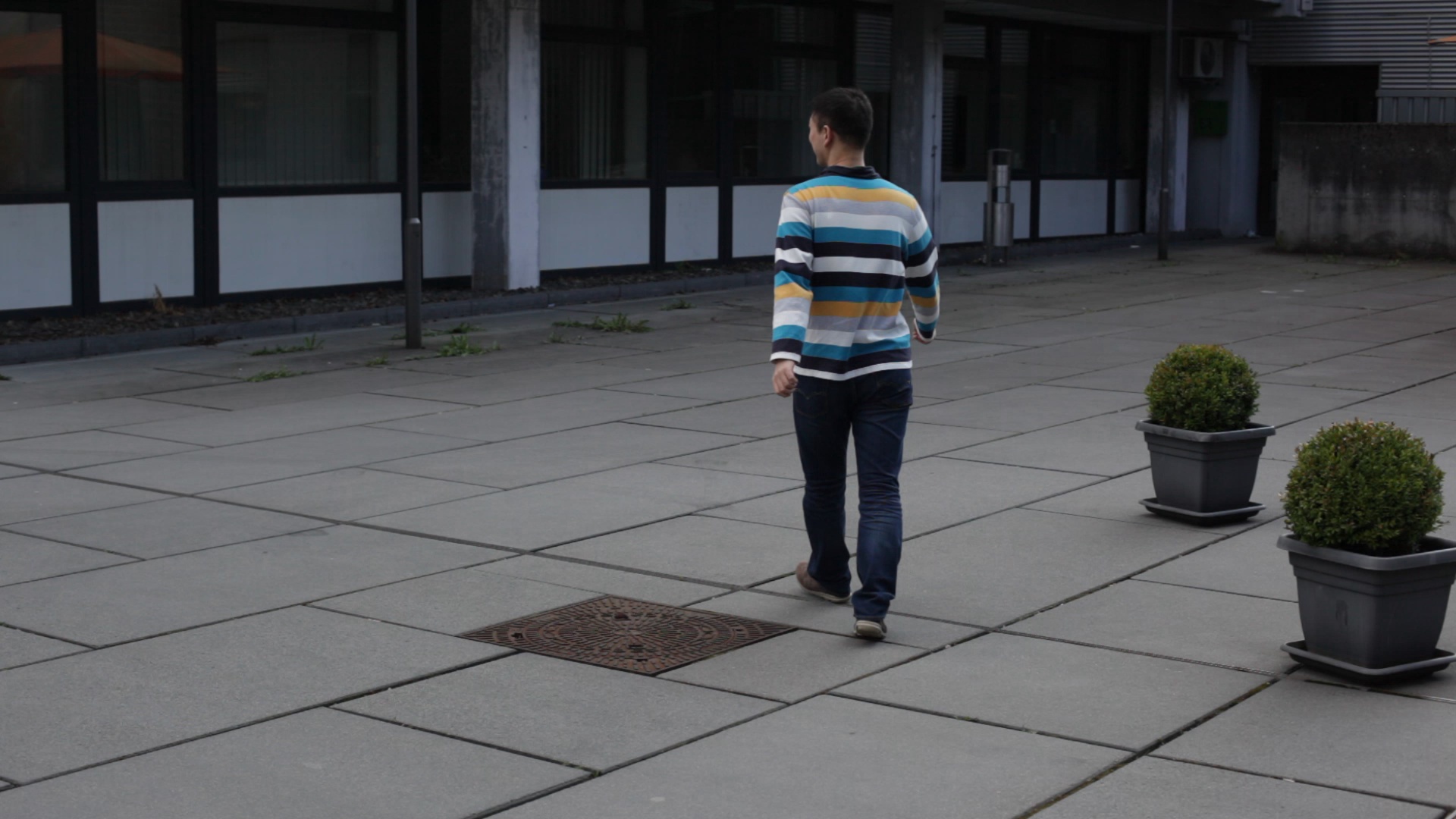}}\\%
{\small Reference}\\%
\vspace{17mm}
{\small ~\\~\\MonoPerfCap\\``Wp\_outdoor''}
}%
\hfill%
\parbox[t]{\qfpscale}{
\centering %
\fbox{\includegraphics[width=\qfpscale,trim=284 80 233 200,clip]{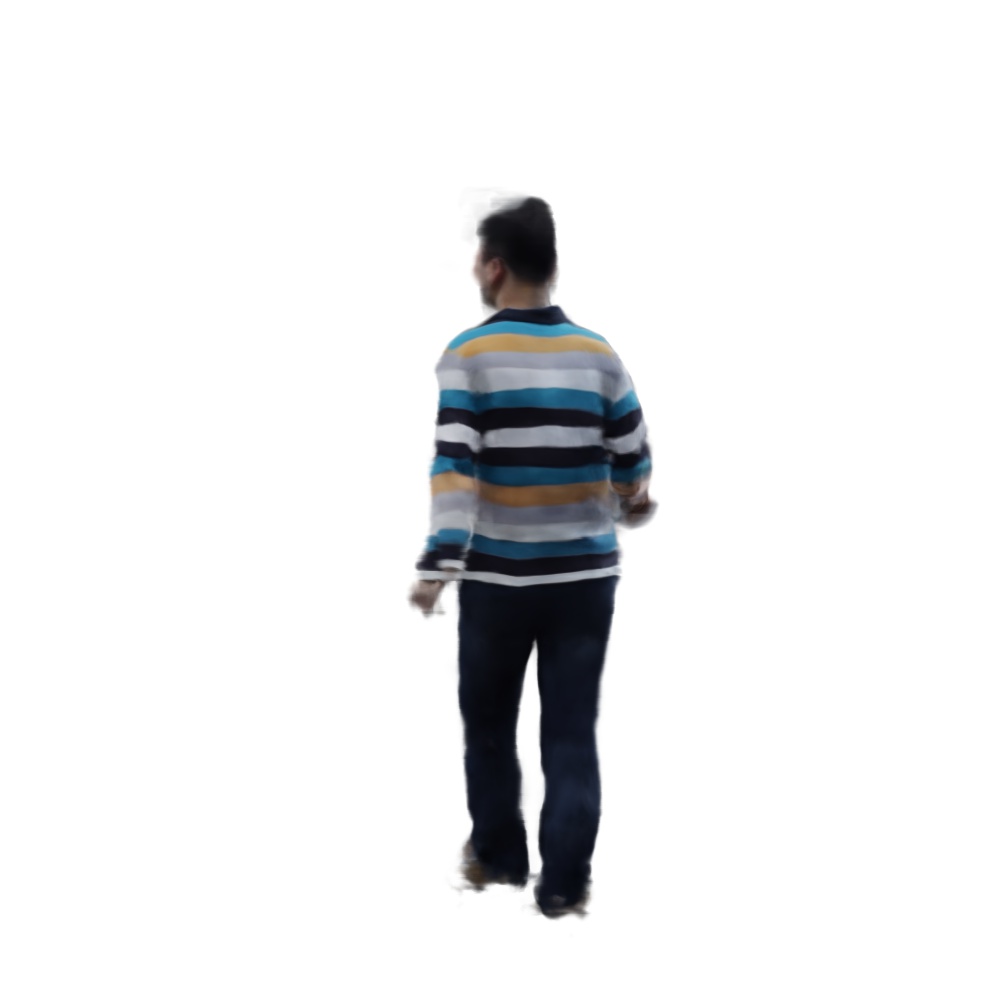}}\\%
\fbox{\includegraphics[width=\qfpscale,trim=248 80 213 205,clip]{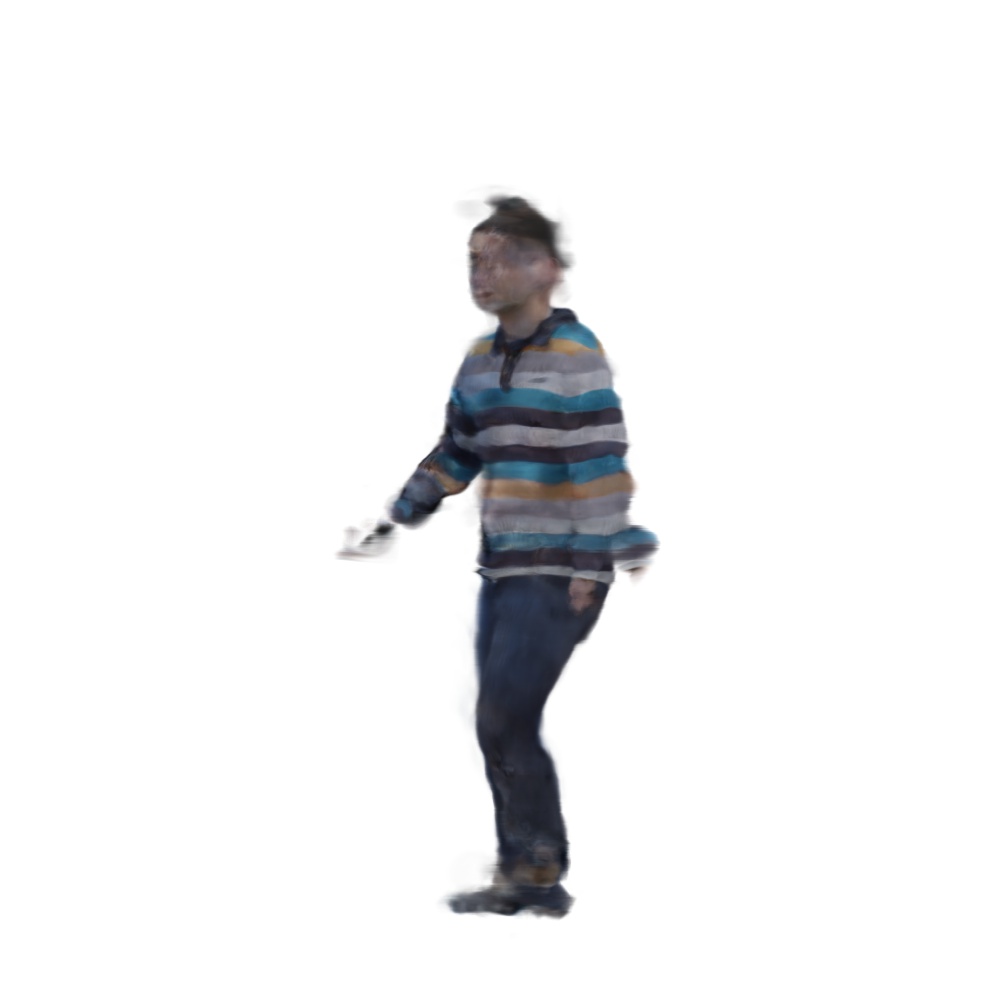}}\\%
\fbox{\includegraphics[width=\qfpscale,trim=287 80 237 205,clip]{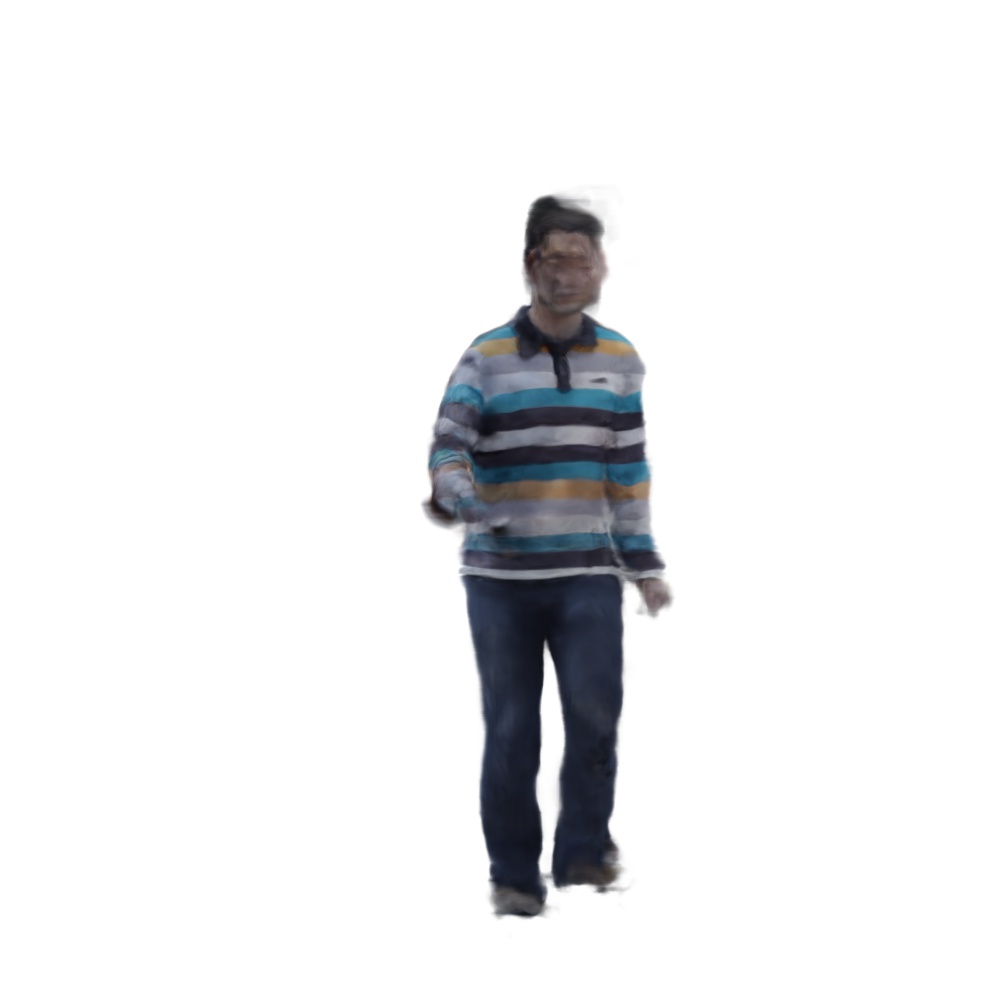}}\\%
{\small NeuralBody}%
}%
\hfill%
\parbox[t]{\qfpscale}{
\centering
\fbox{\includegraphics[width=\qfpscale,trim=300 165 310 215,clip]{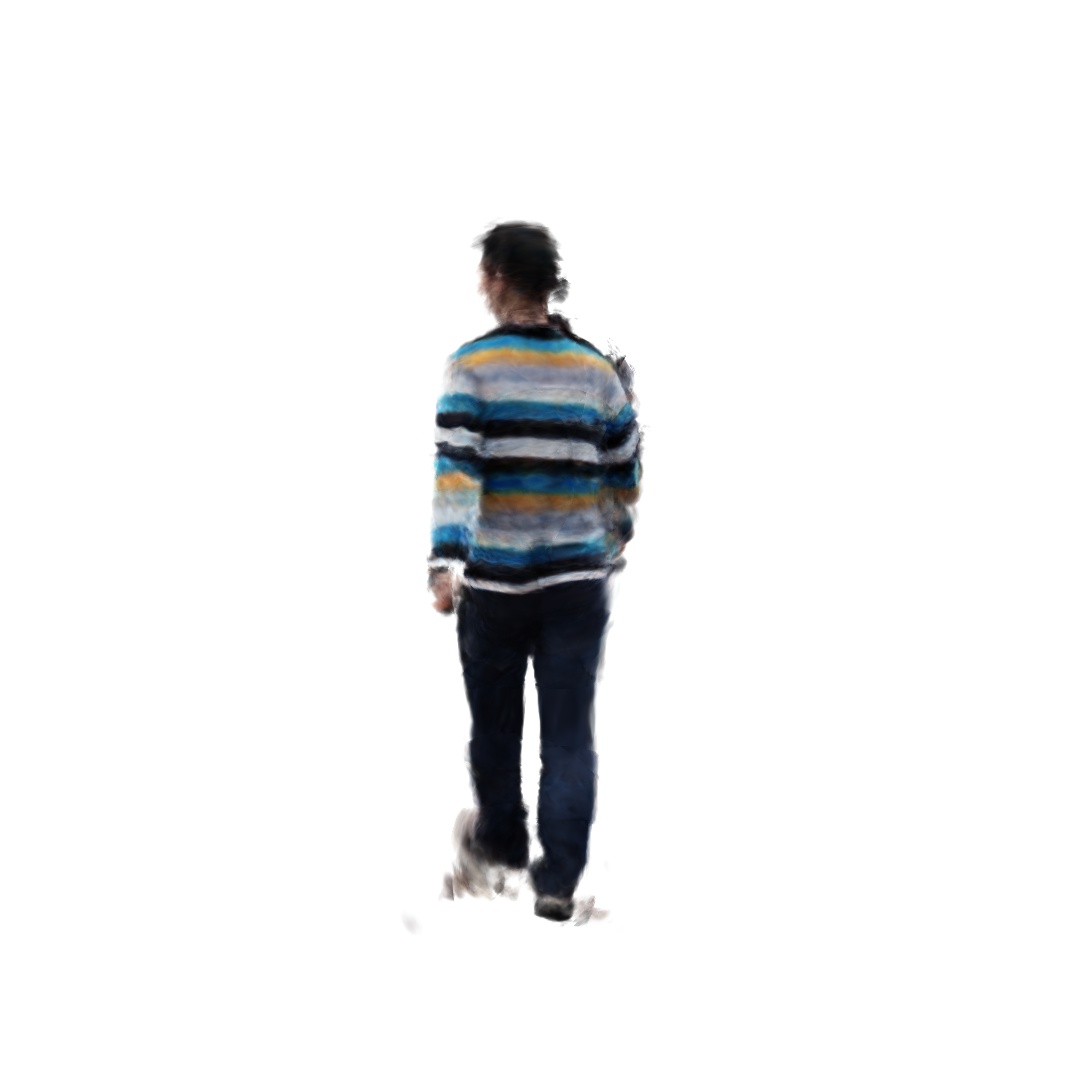}}\\%
\fbox{\includegraphics[width=\qfpscale,trim=268 175 296 220,clip]{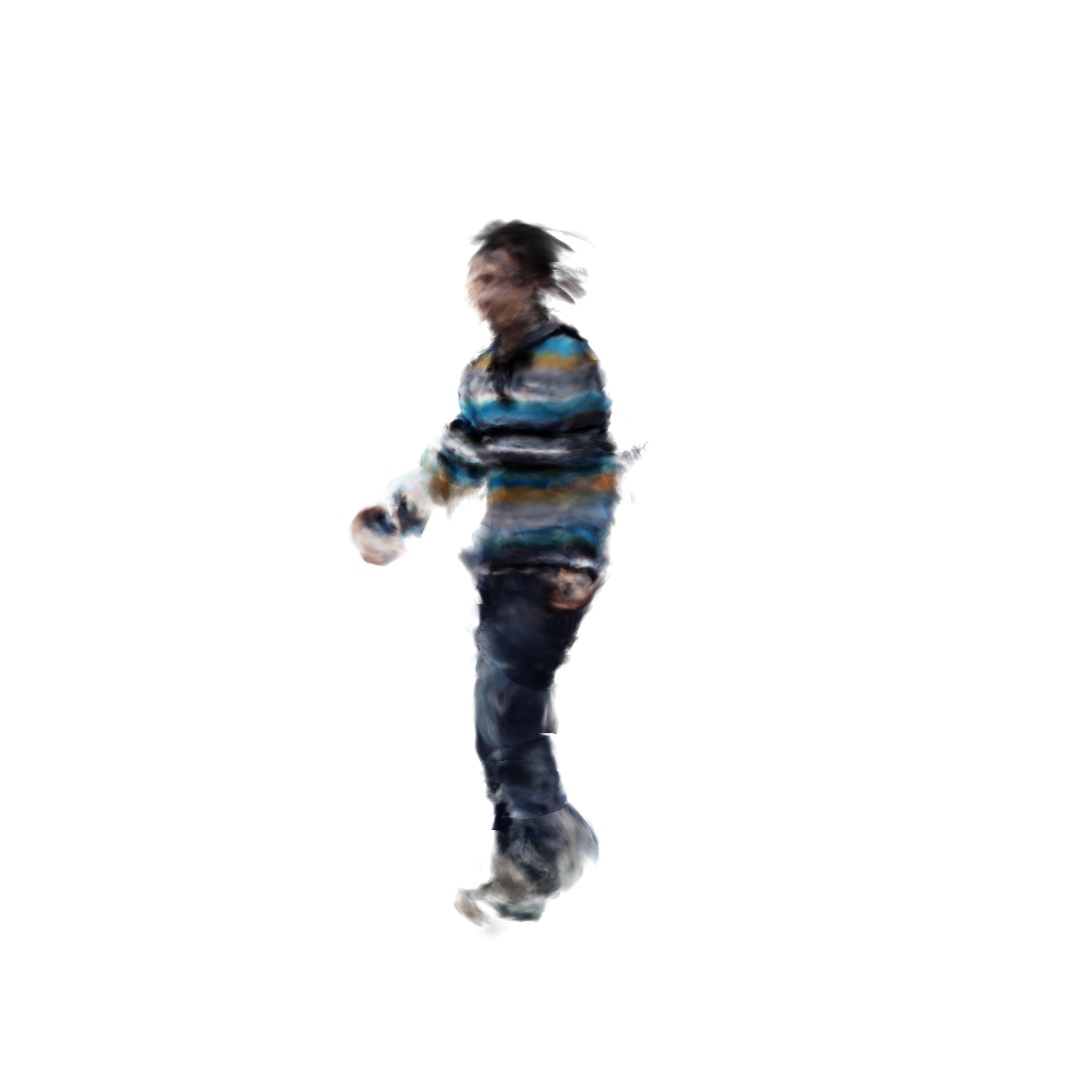}}\\%
\fbox{\includegraphics[width=\qfpscale,trim=303 160 311 220,clip]{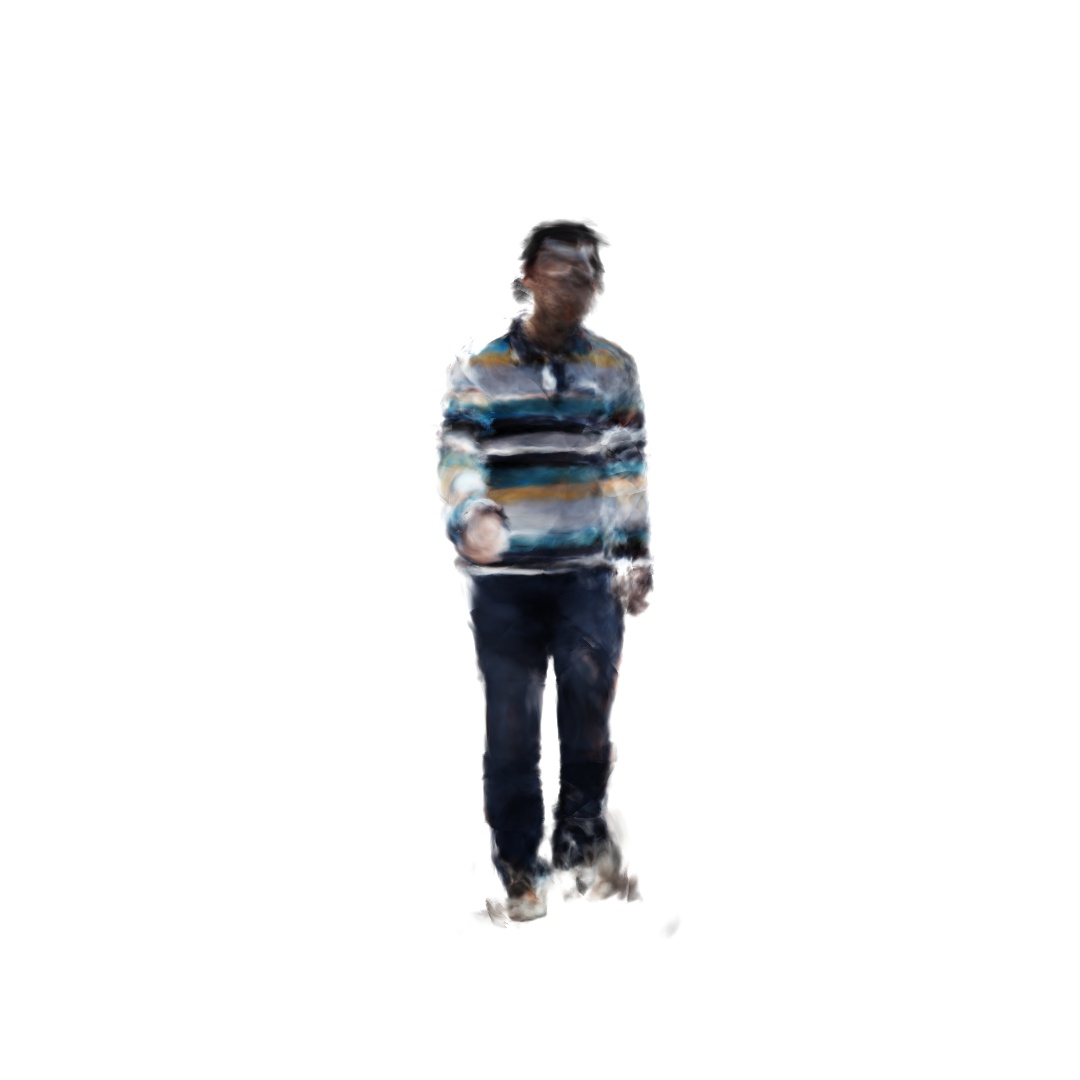}}\\%
{\small Ours w/o rf.}%
}%
\hfill%
\parbox[t]{\qfpscale}{
\centering
\fbox{\includegraphics[width=\qfpscale,trim=300 170 310 210,clip]{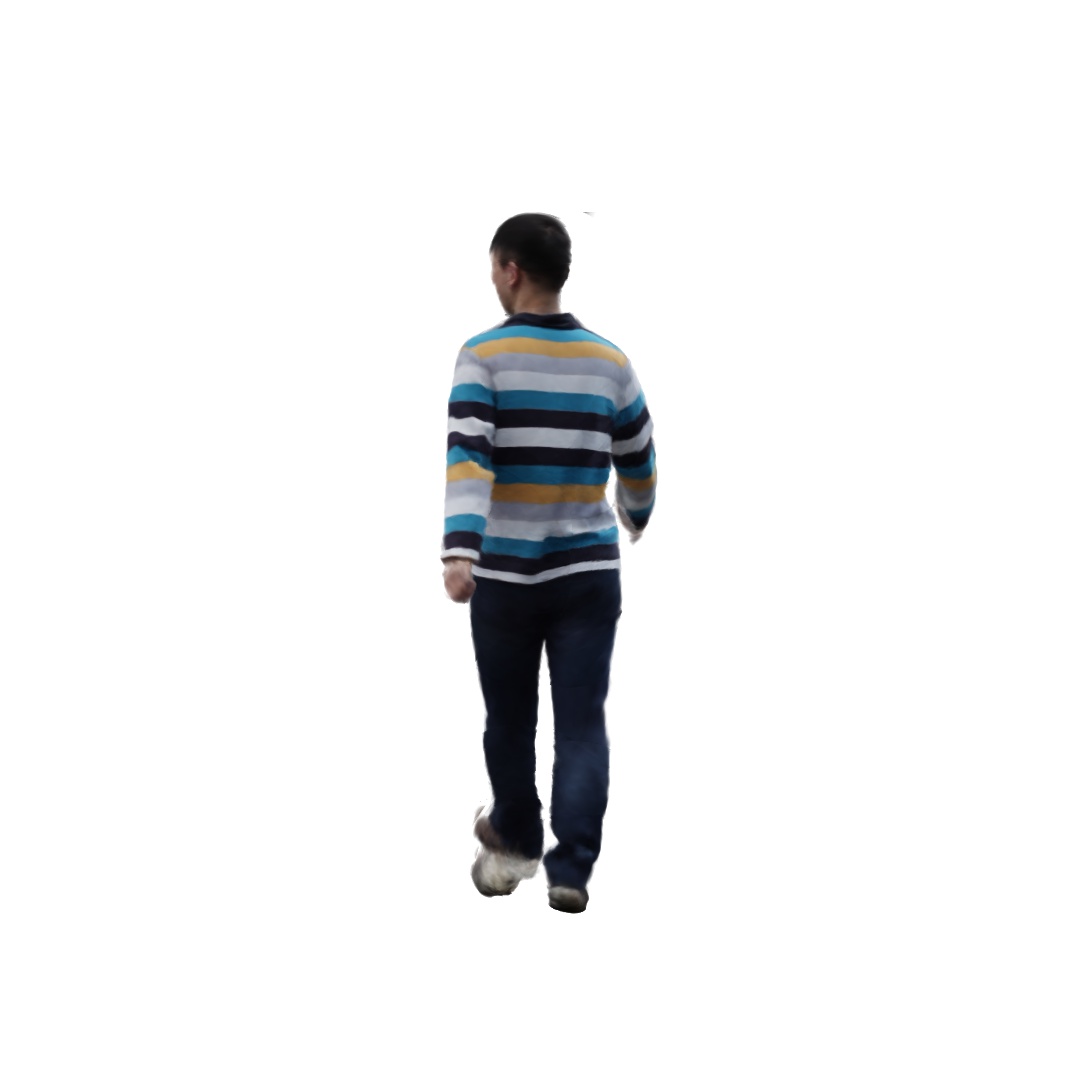}}\\%
\fbox{\includegraphics[width=\qfpscale,trim=268 175 296 220,clip]{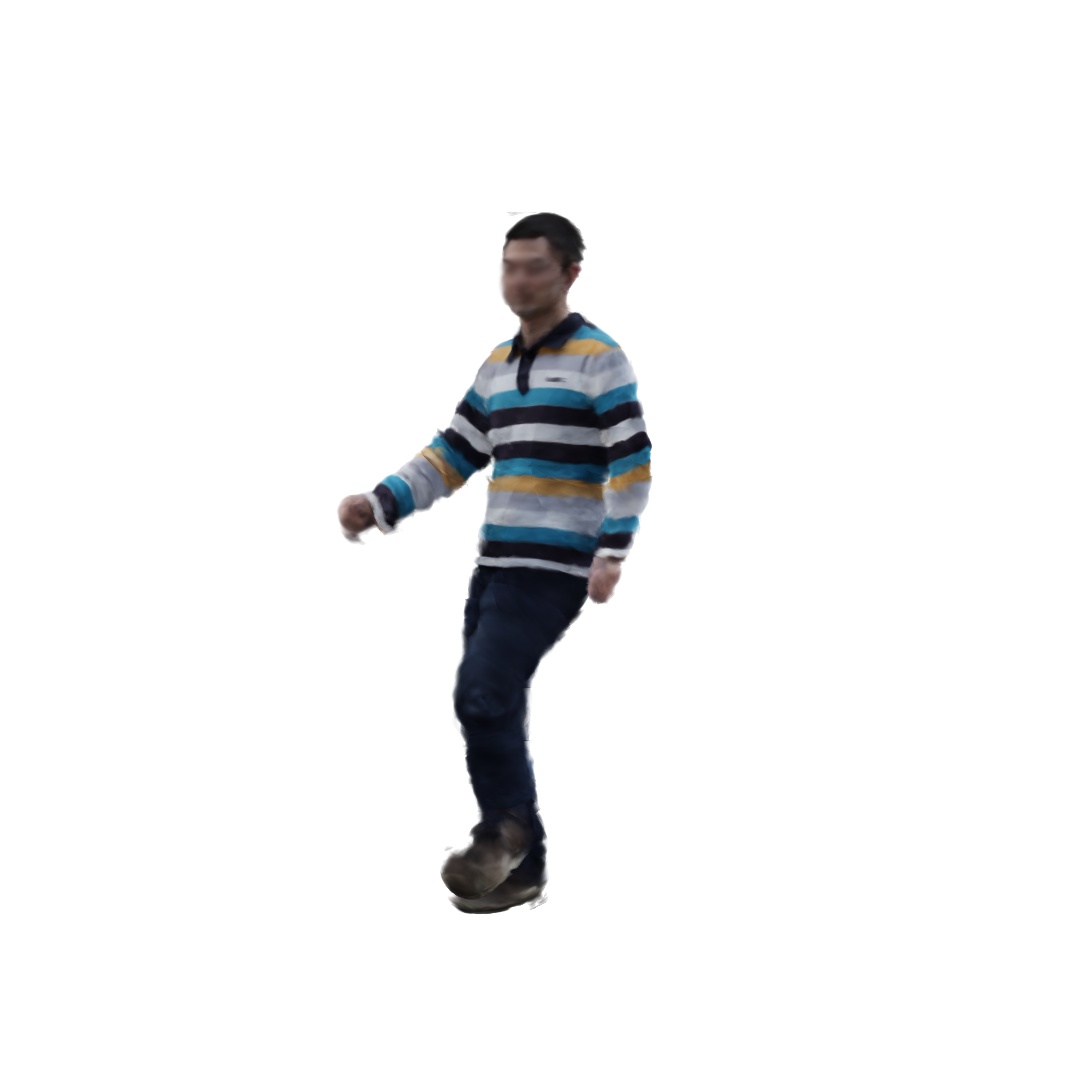}}\\%
\fbox{\includegraphics[width=\qfpscale,trim=303 165 311 215,clip]{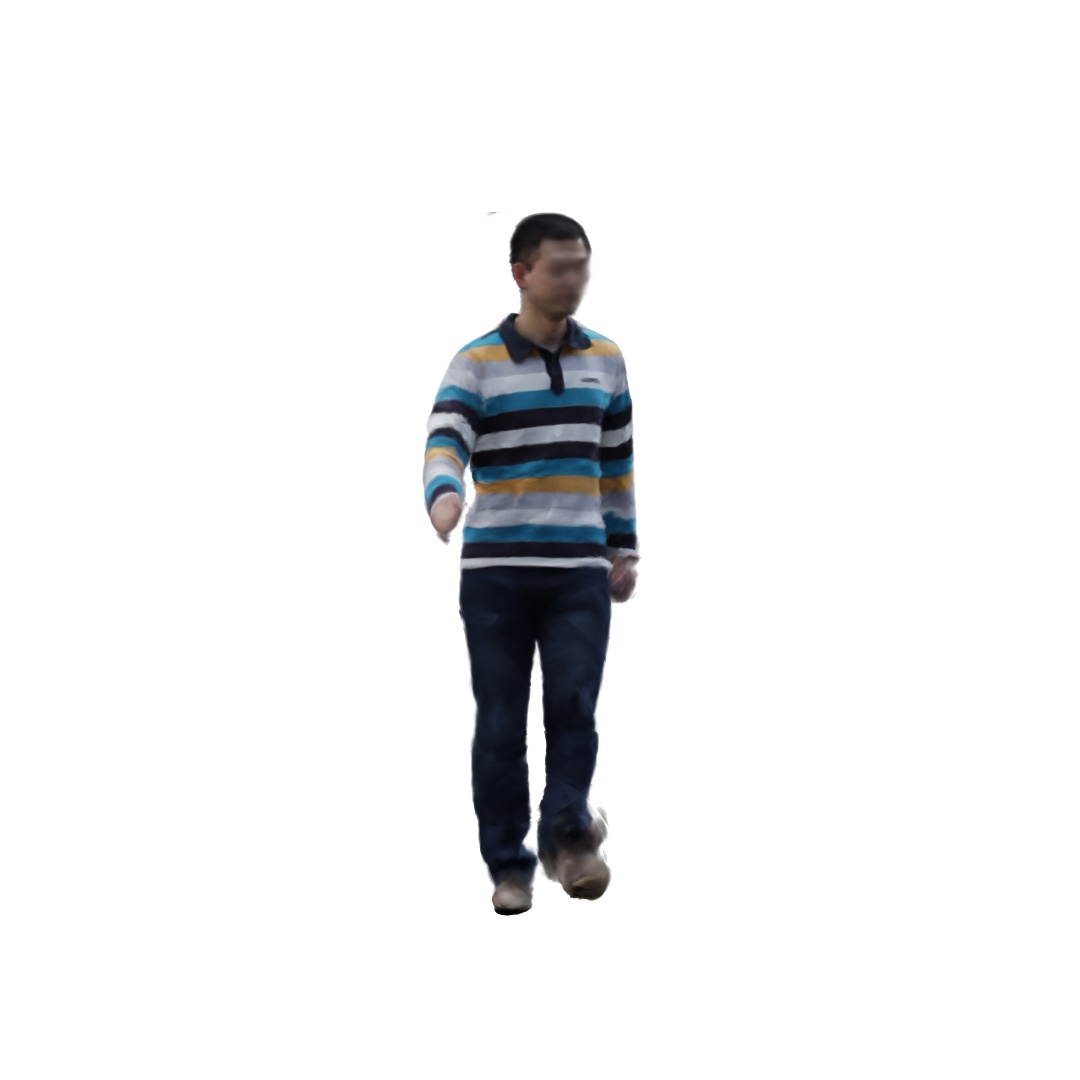}}\\%
{\small Ours}%
}%
\caption{\textbf{Novel view synthesis from all models.} All models are trained with SPIN~\cite{kolotouros2019learning_spin} estimated human pose and camera parameters. A-NeRF renderings (ours) align better with the reference images (top row), and the rendered novel views (2nd and 3rd rows) show better details (limbs, facial features).
}
\label{fig:novel-view-synthesis}
\end{figure*}

\newlength\qrtpscale
\setlength\qrtpscale{0.100\linewidth}
\newlength\qaniscalea
\setlength\qaniscalea{0.55\linewidth}
\newlength\qaniscaleb
\setlength\qaniscaleb{0.27\linewidth}%
\newlength\qaniscalec
\setlength\qaniscalec{0.29\linewidth}%
\newlength\qaniscaled
\setlength\qaniscaled{0.61\linewidth}
\begin{figure*}[h]
\setlength{\fboxrule}{0pt}%
\parbox[t]{\qrtpscale}{
\centering
\fbox{\includegraphics[width=\qrtpscale,trim=795 320 685 130,clip]{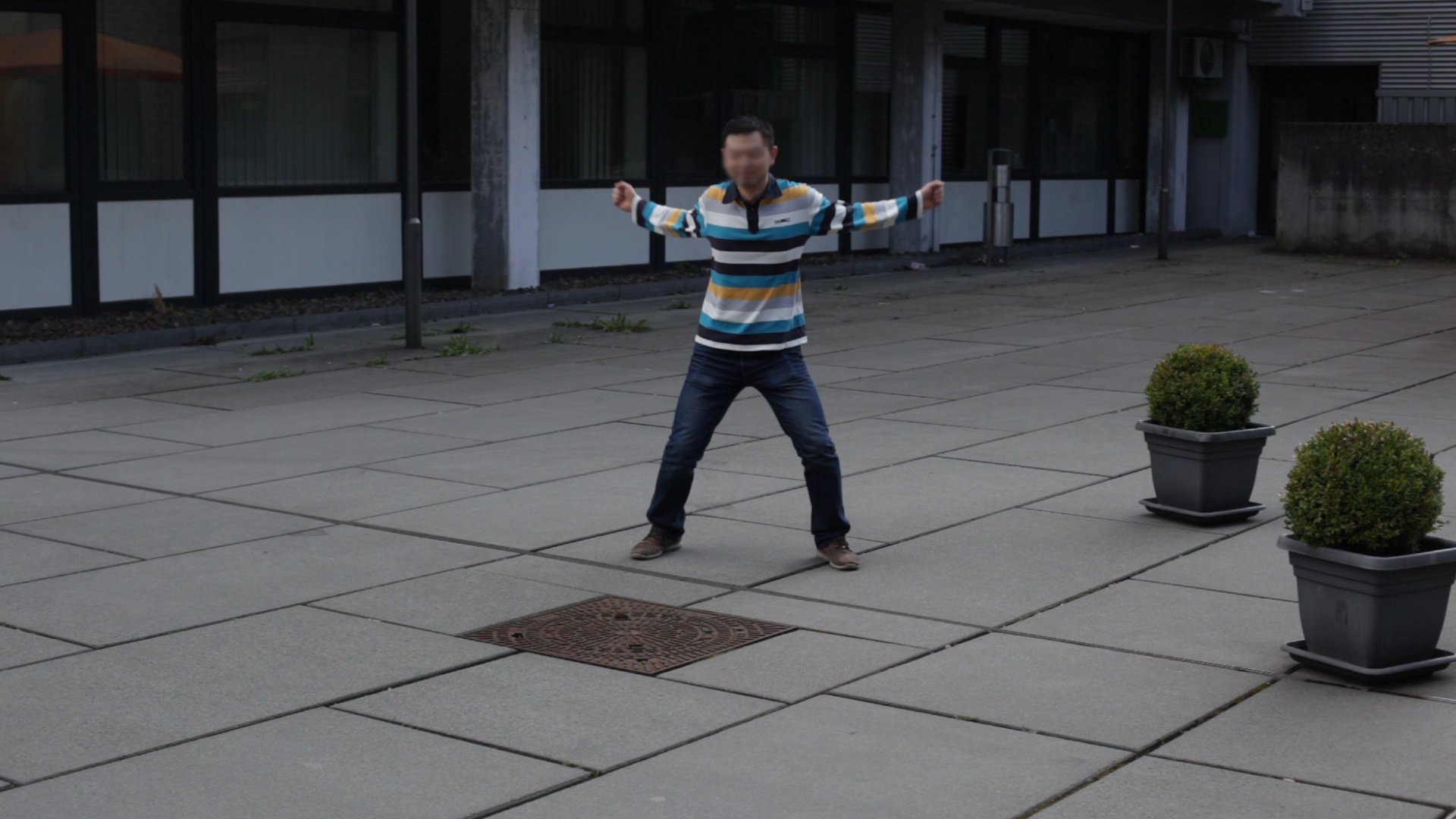}}\\%
\fbox{\includegraphics[width=\qrtpscale,trim=575 0 675 70,clip]{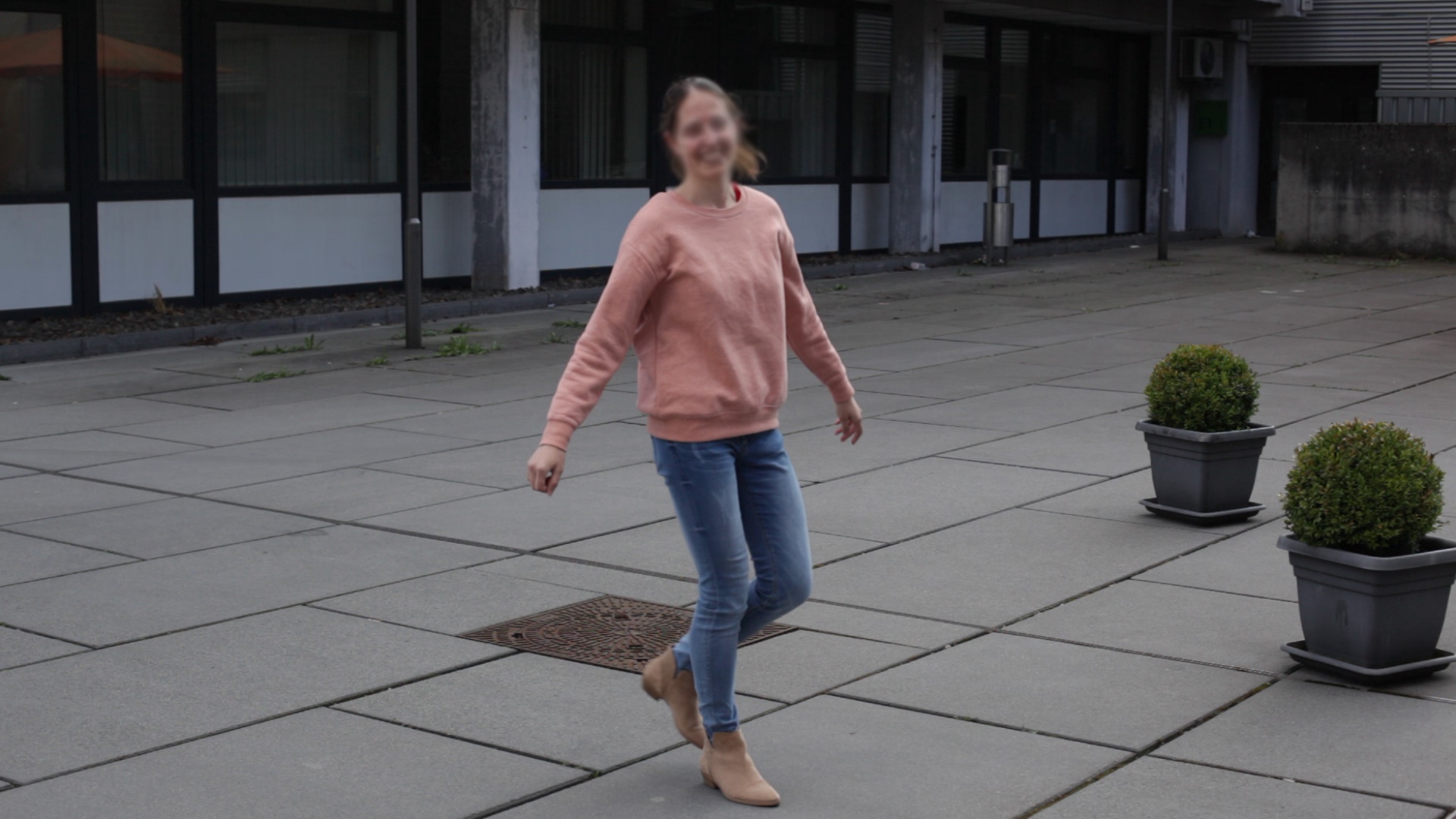}}\\%
{\small Source}
}%
\hfill%
\parbox[t]{\qrtpscale}{
\centering %
\fbox{\includegraphics[width=\qrtpscale,trim=370 300 265 163,clip]{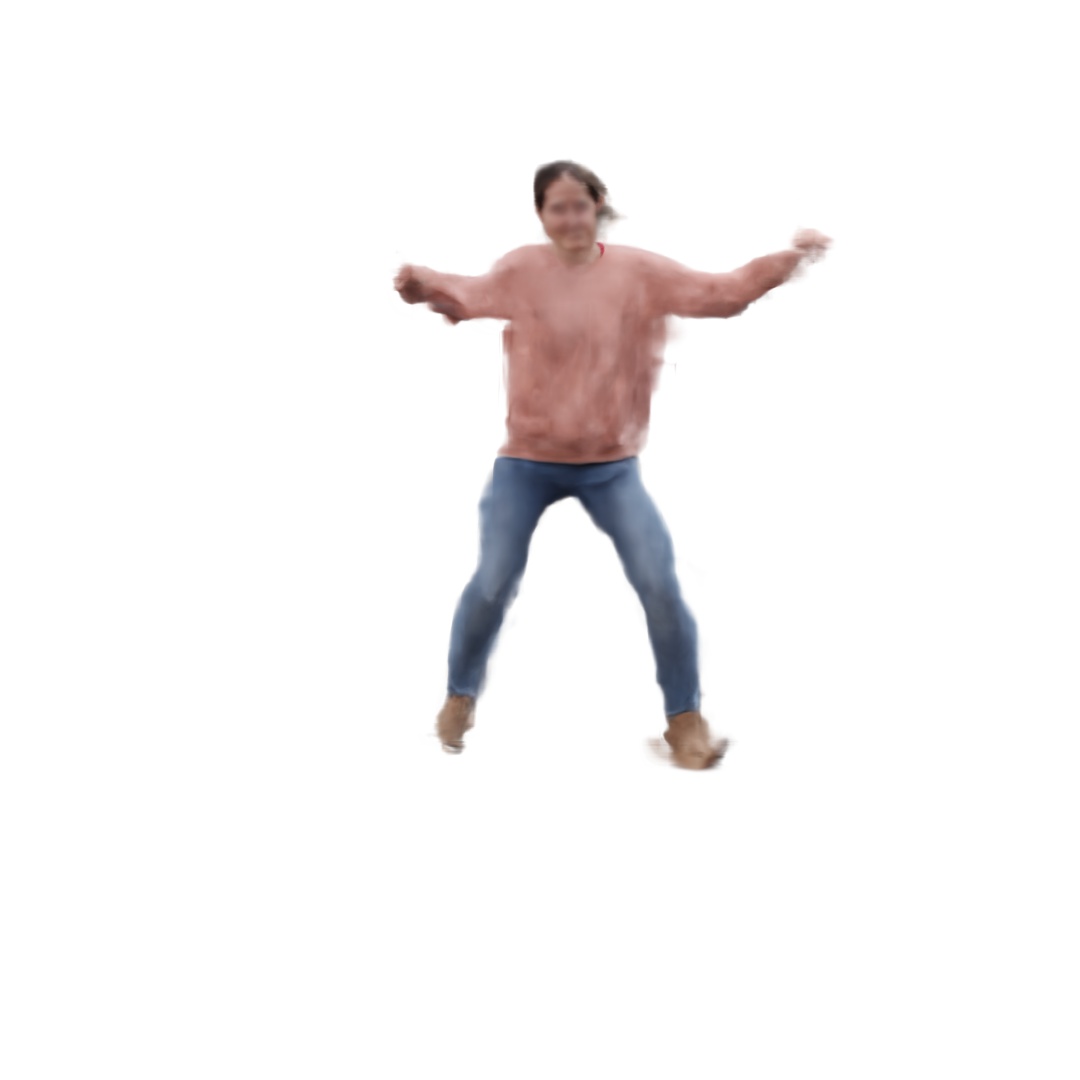}}\\%
\fbox{\includegraphics[width=\qrtpscale,trim=577 0 673 70,clip]{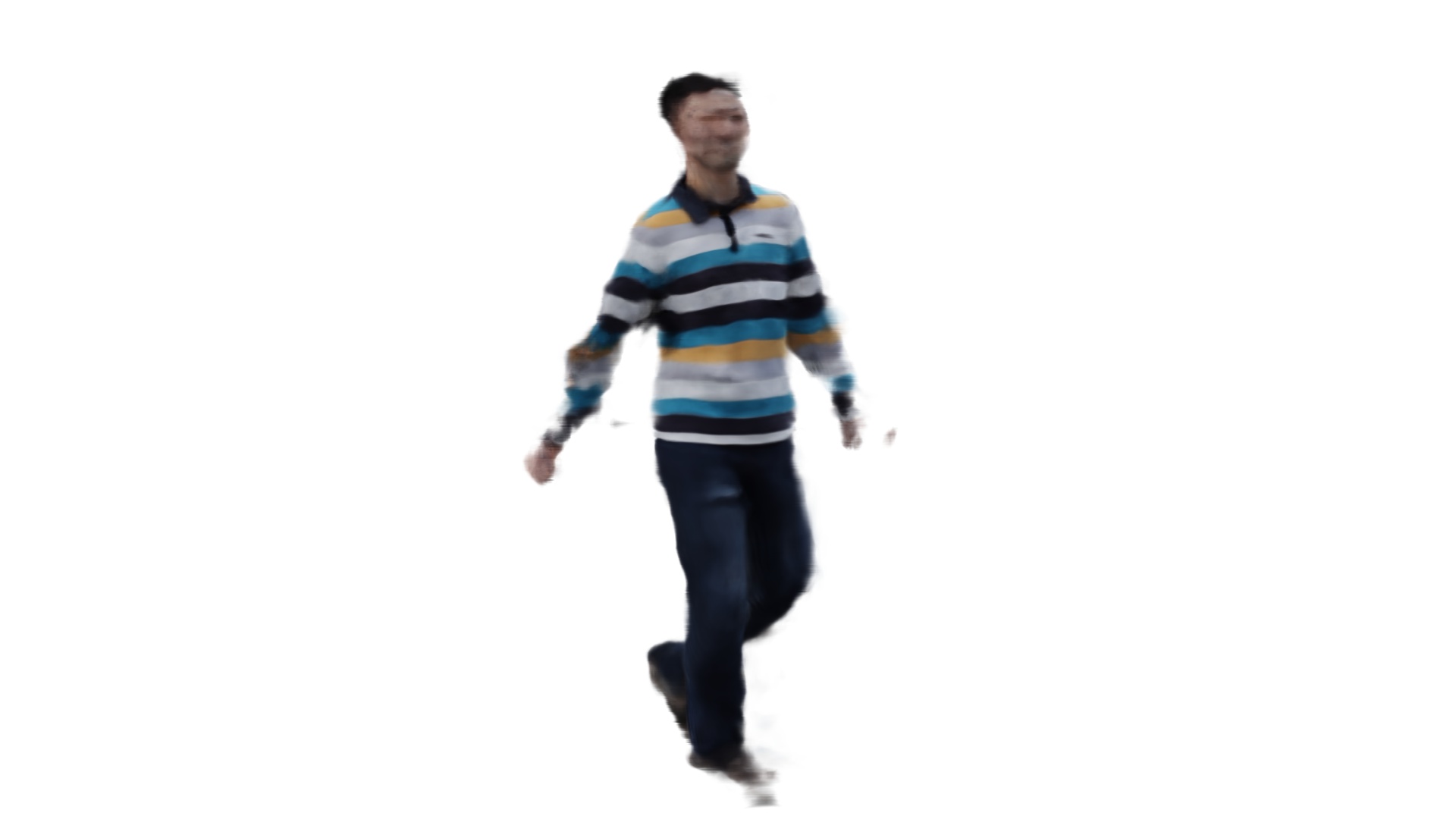}}\\%
{\small NeuralBody}%
}
\hfill%
\parbox[t]{\qrtpscale}{
\centering
\fbox{\includegraphics[width=\qrtpscale,trim=380 333 285 167,clip]{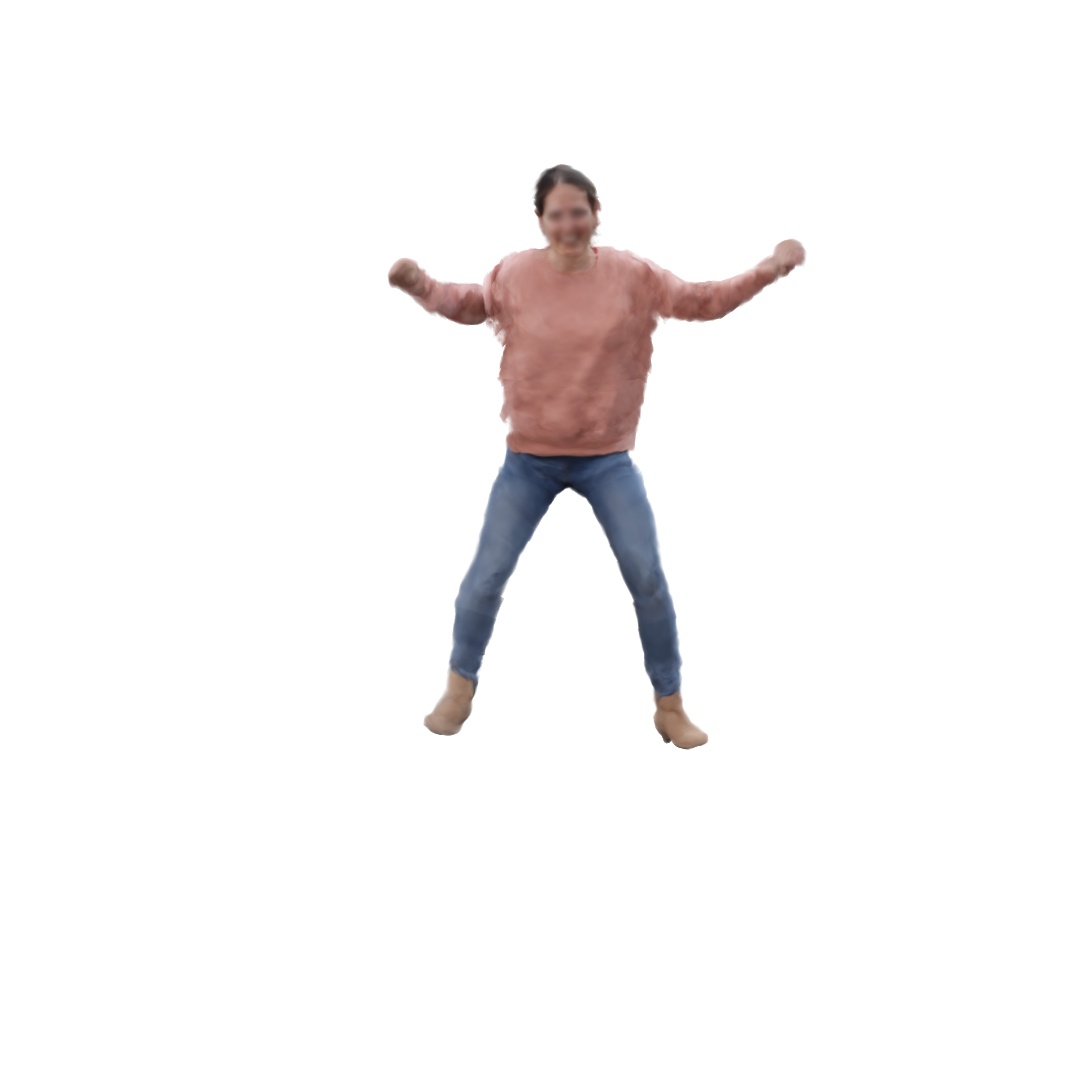}}\\%
\fbox{\includegraphics[width=\qrtpscale,trim=577 0 673 70,clip]{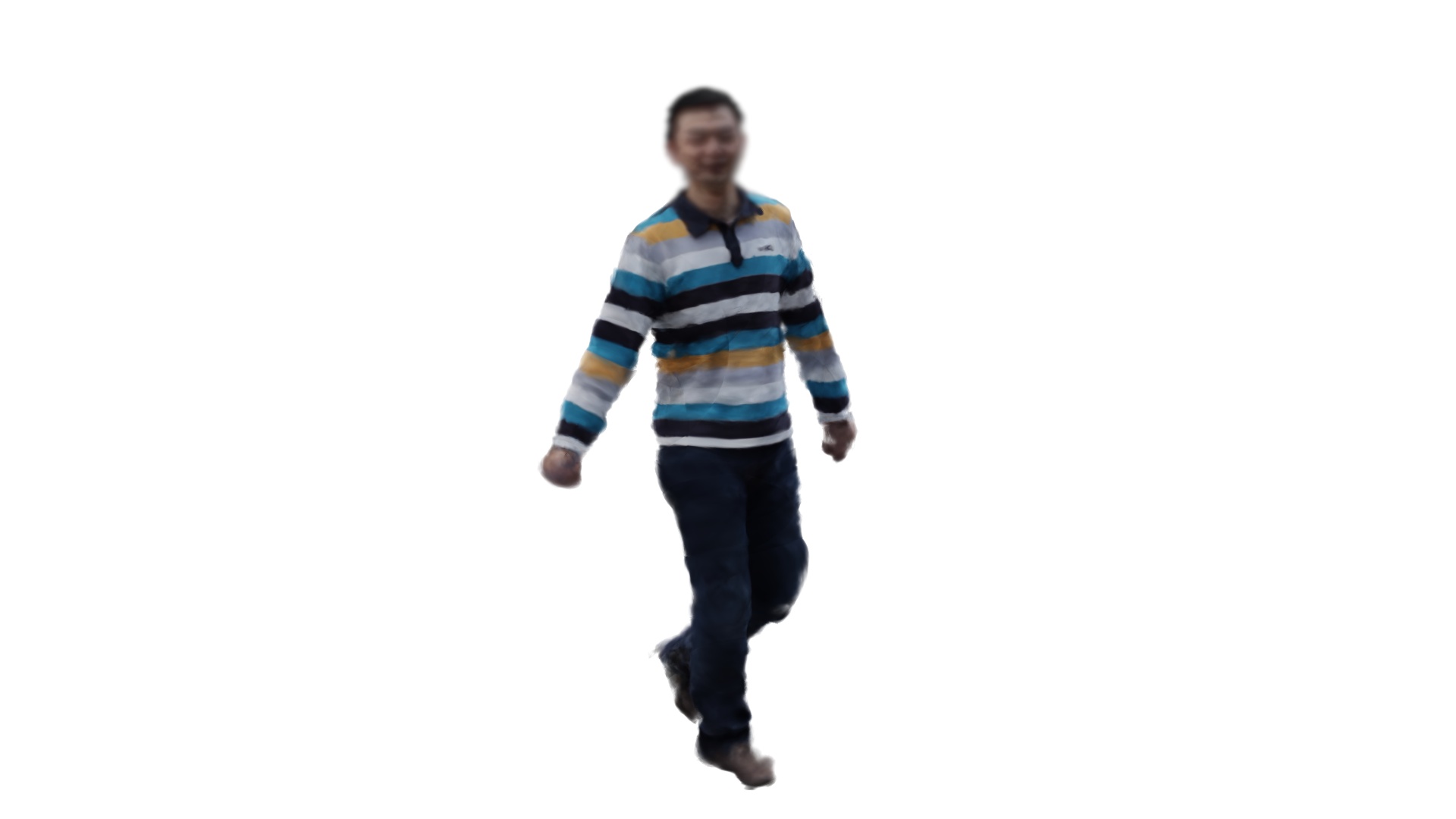}}\\%
{\small Ours }%
}
\hfill%
\parbox[t]{\qrtpscale}{
\centering
\fbox{\includegraphics[width=\qrtpscale,trim=95 300 545 180,clip]{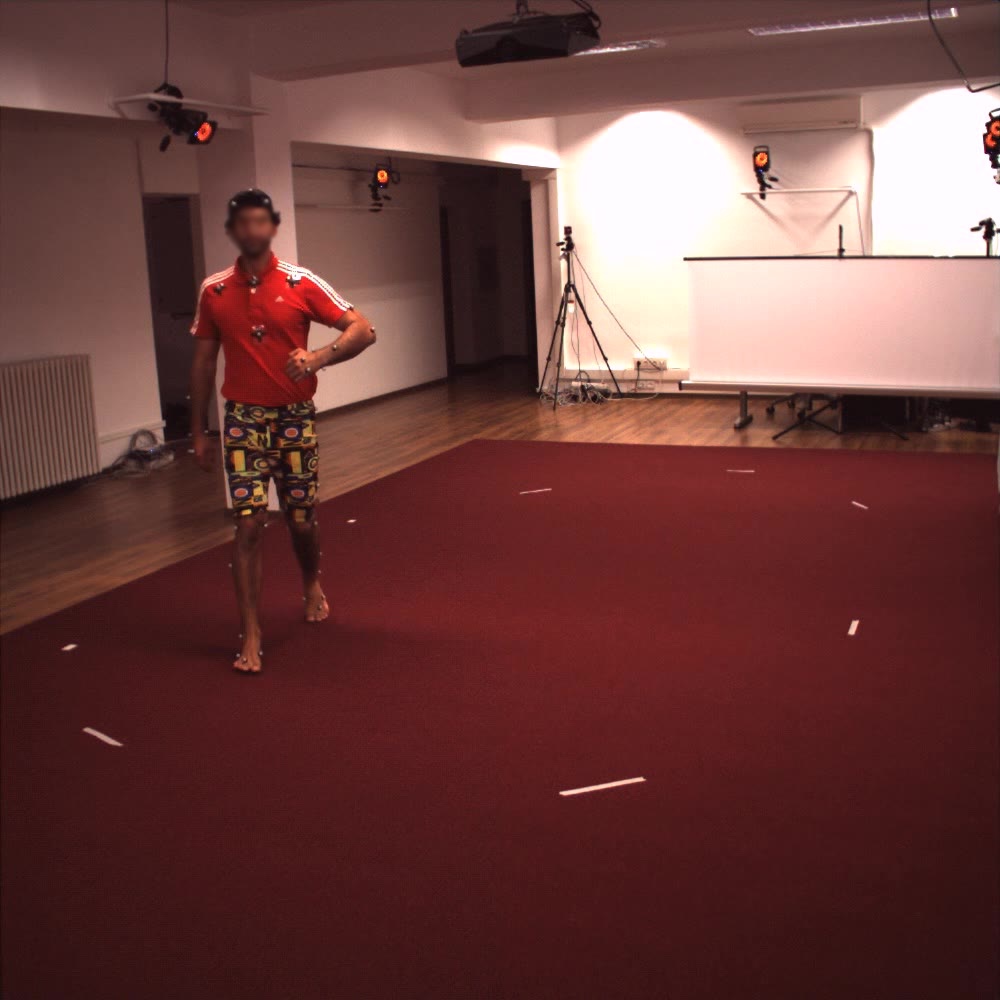}}\\%
\fbox{\includegraphics[width=\qrtpscale,trim=371 280 352 305,clip]{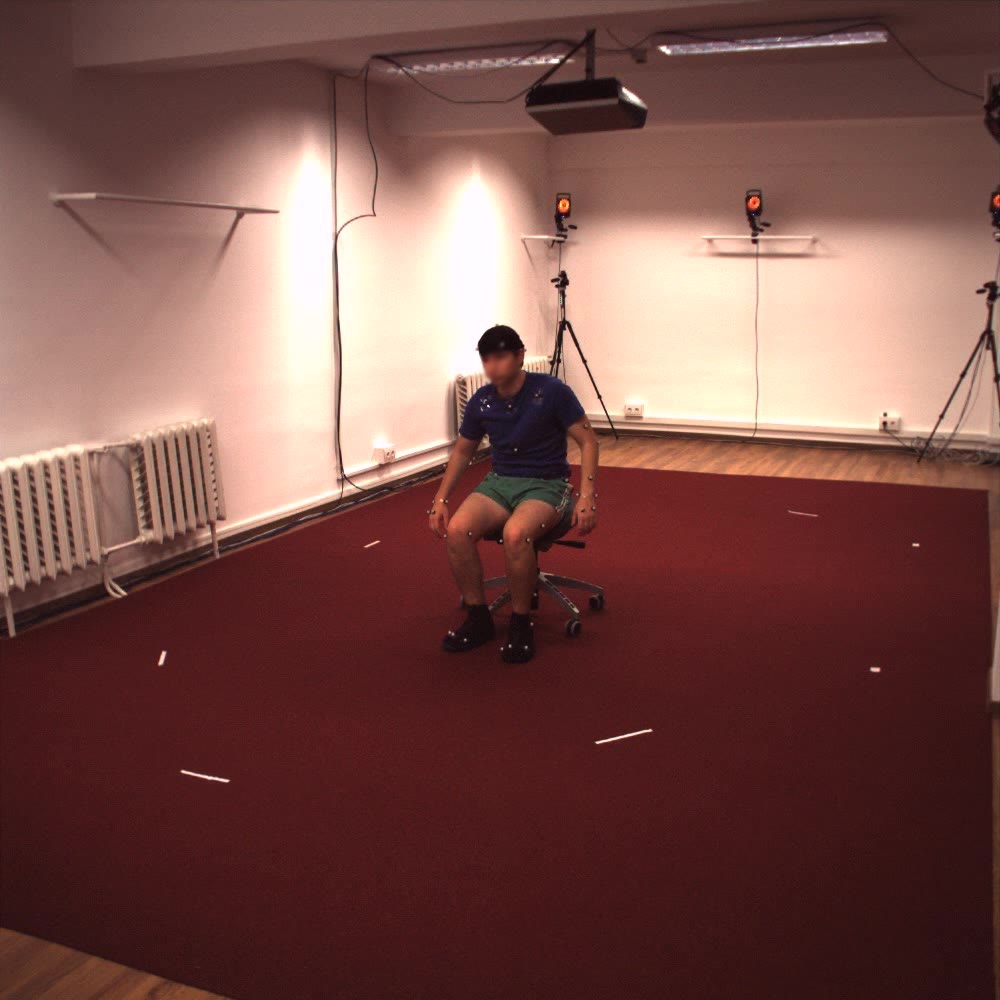}}\\%
{\small Source}
}%
\hfill%
\parbox[t]{\qrtpscale}{
\centering
\fbox{\includegraphics[width=\qrtpscale,trim=95 300 545 180,clip]{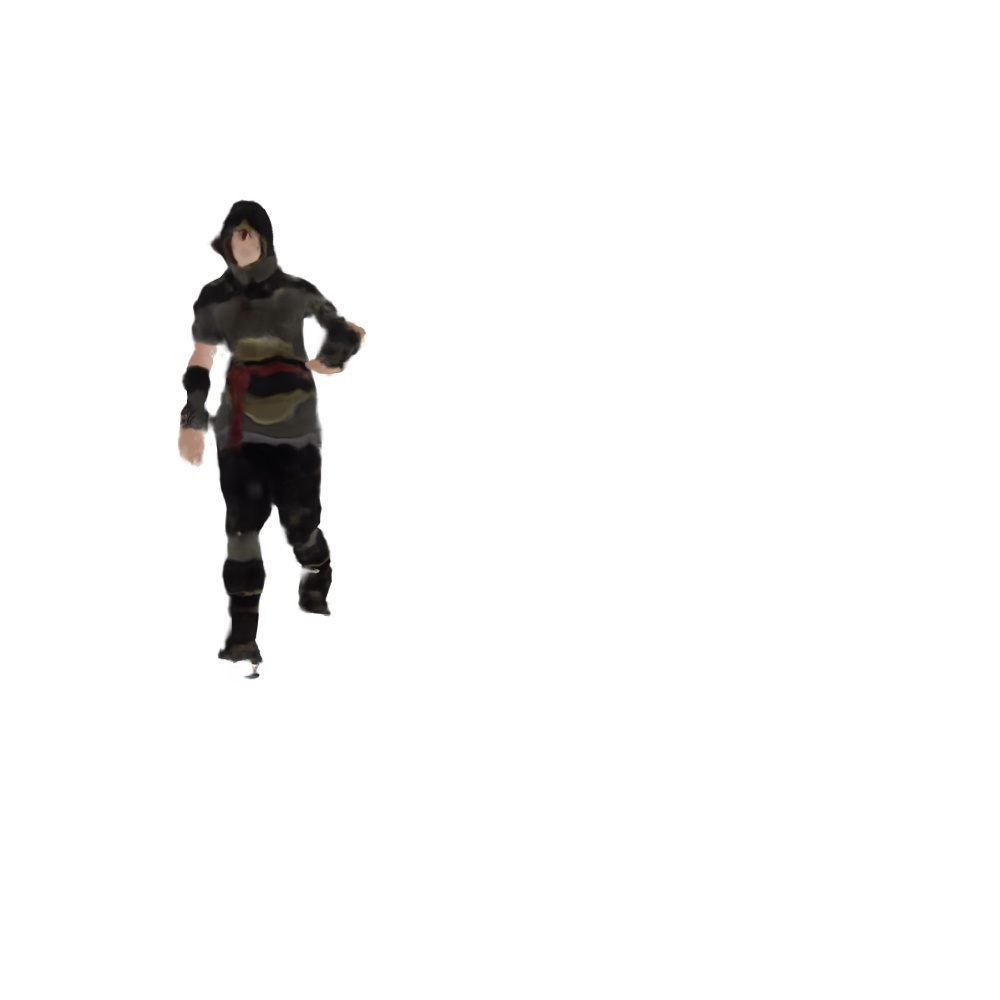}}\\%

\fbox{\includegraphics[width=\qrtpscale,trim=372 280 351 305,clip]{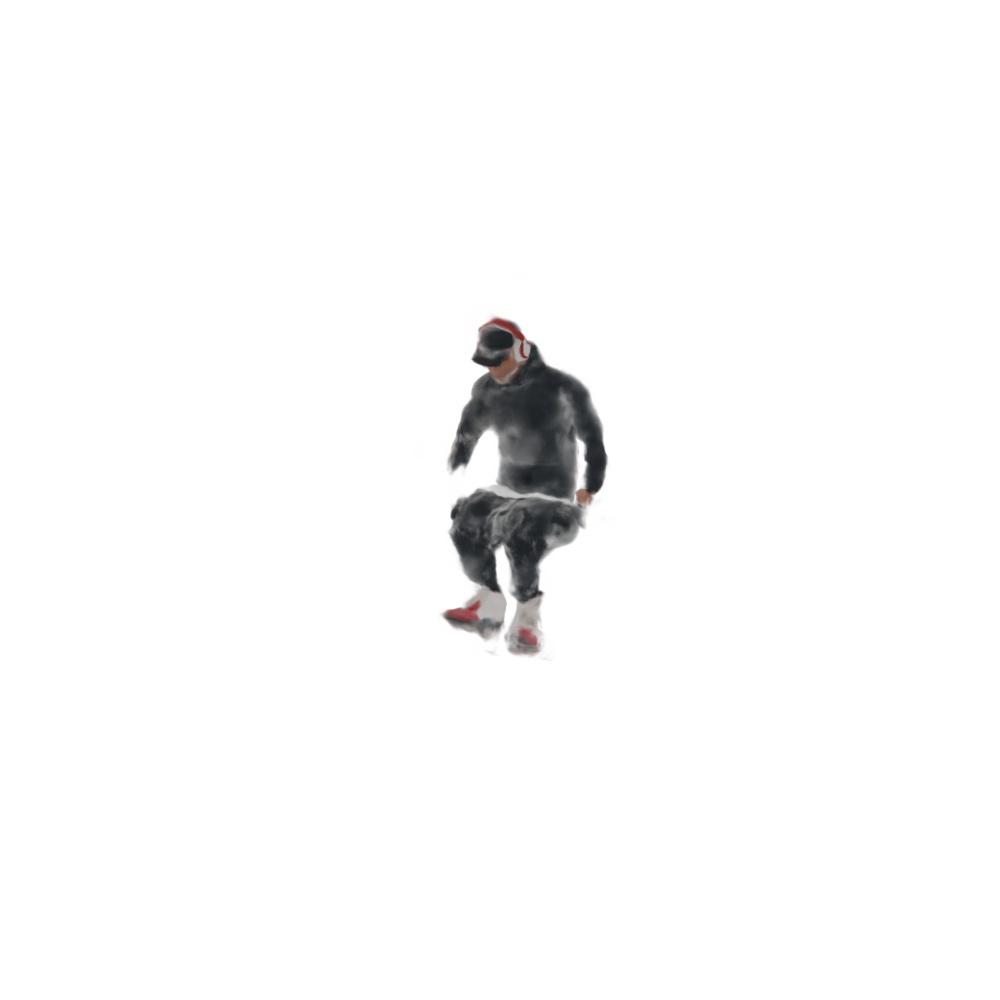}}\\%
{\small NeuralBody}%
}%
\hfill%
\parbox[t]{\qrtpscale}{
\centering
\fbox{\includegraphics[width=\qrtpscale,trim=95 300 545 180,clip]{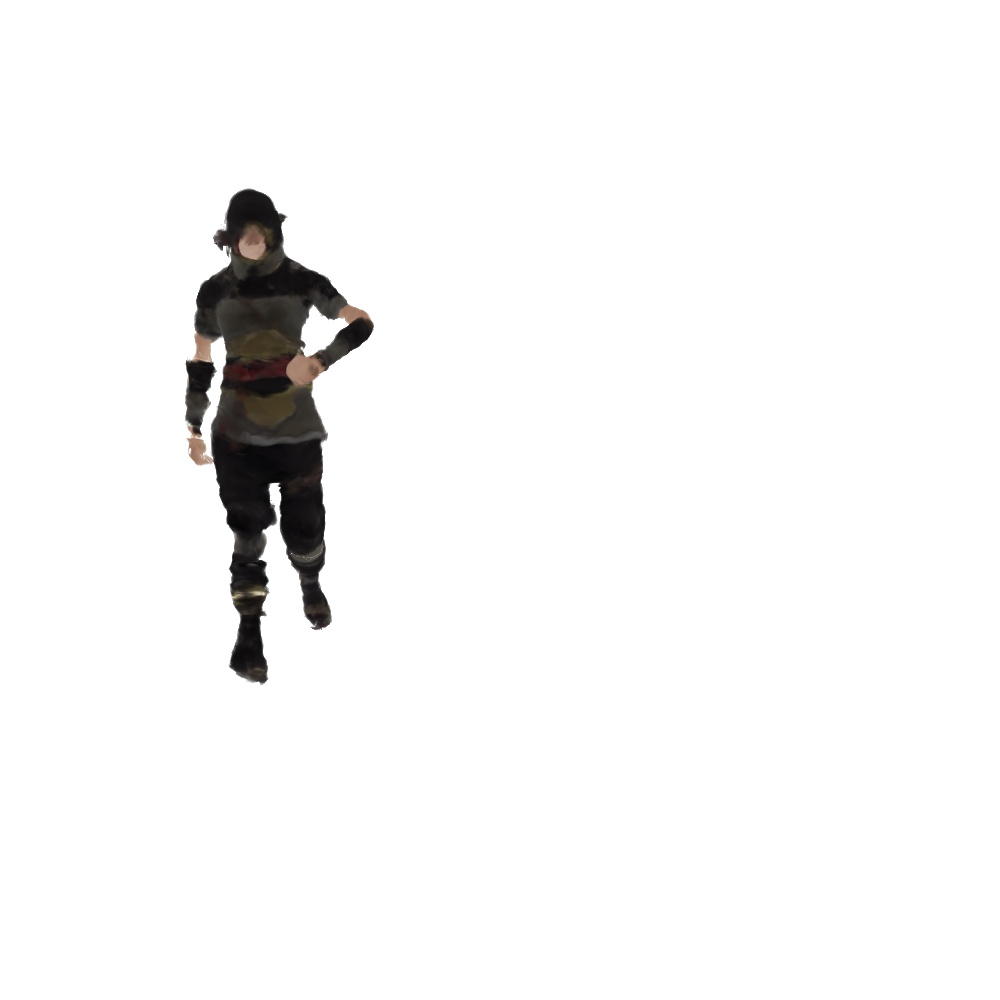}}\\%
\fbox{\includegraphics[width=\qrtpscale,trim=371 280 352 305,clip]{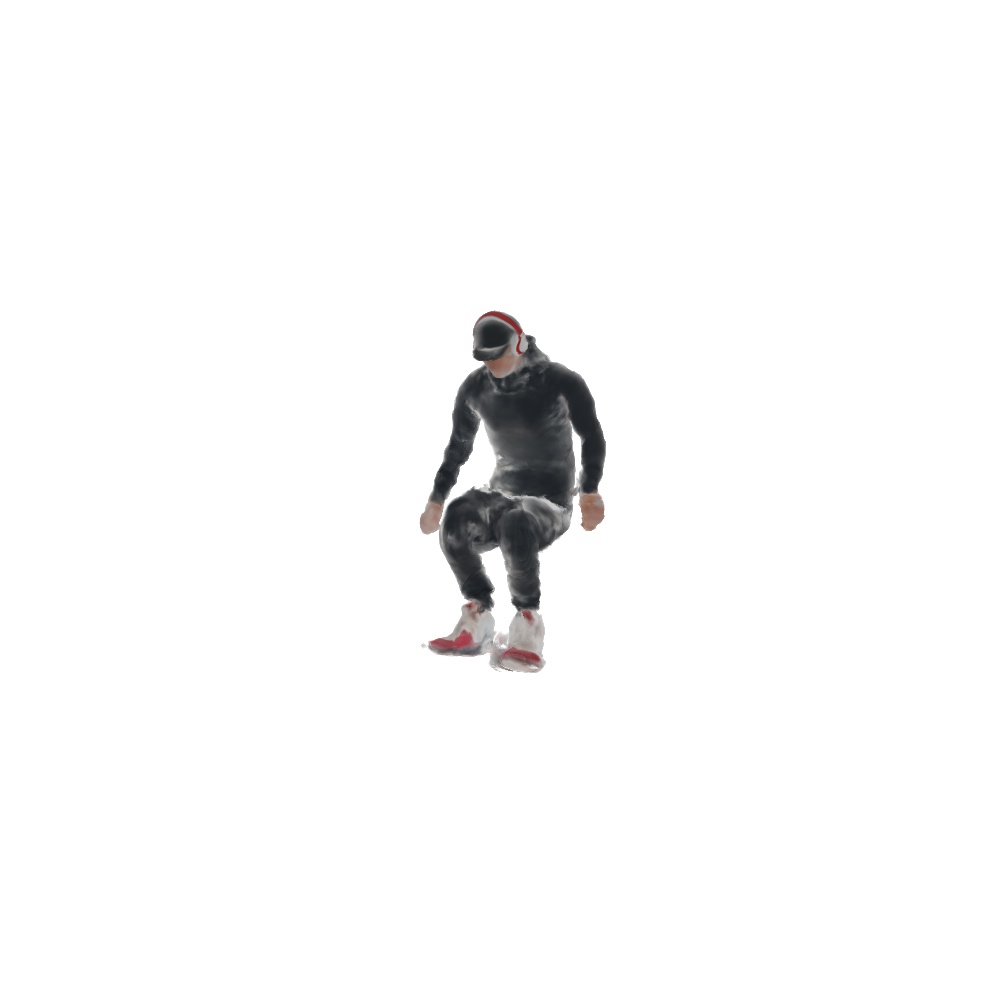}}\\%
{\small Ours }%
}%
\hfill%
\parbox[t]{\qrtpscale}{
\centering
\fbox{\includegraphics[width=\qrtpscale,trim=420 260 320 340,clip]{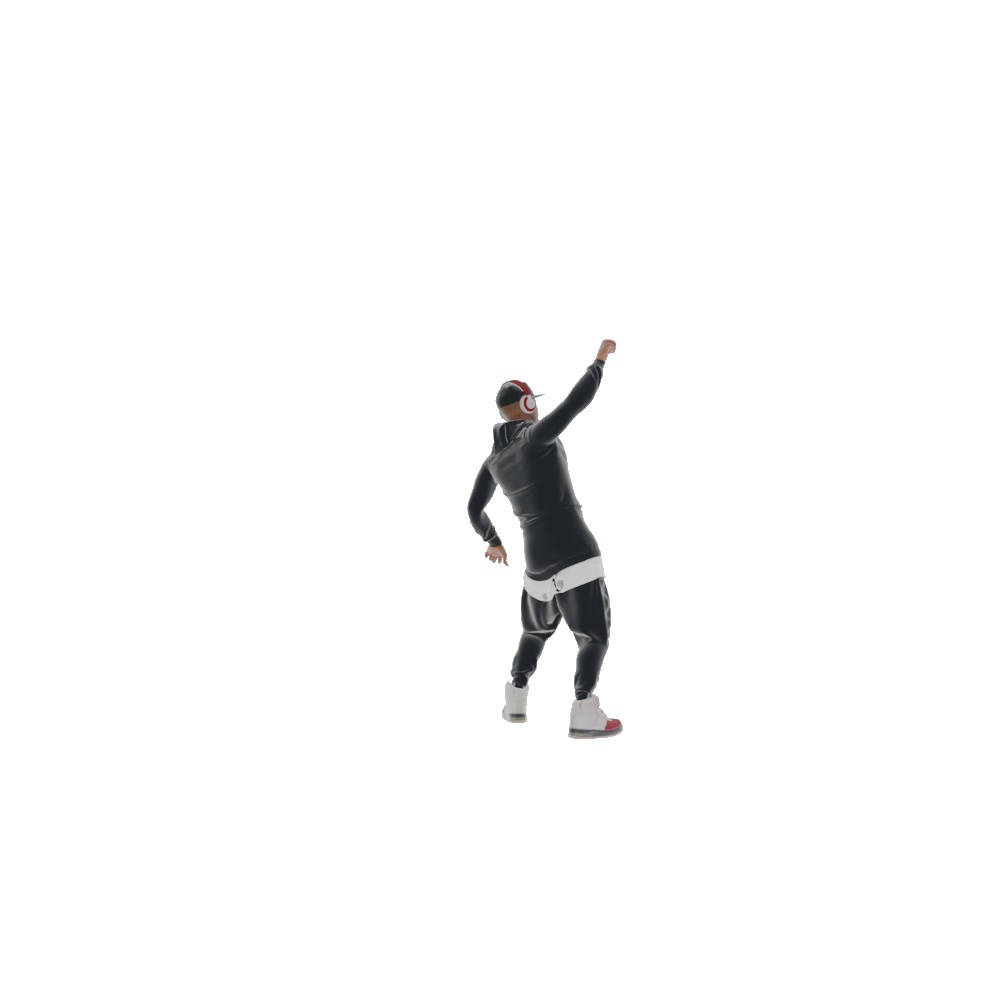}}\\%
\fbox{\includegraphics[width=\qrtpscale,trim=387 280 387 380,clip]{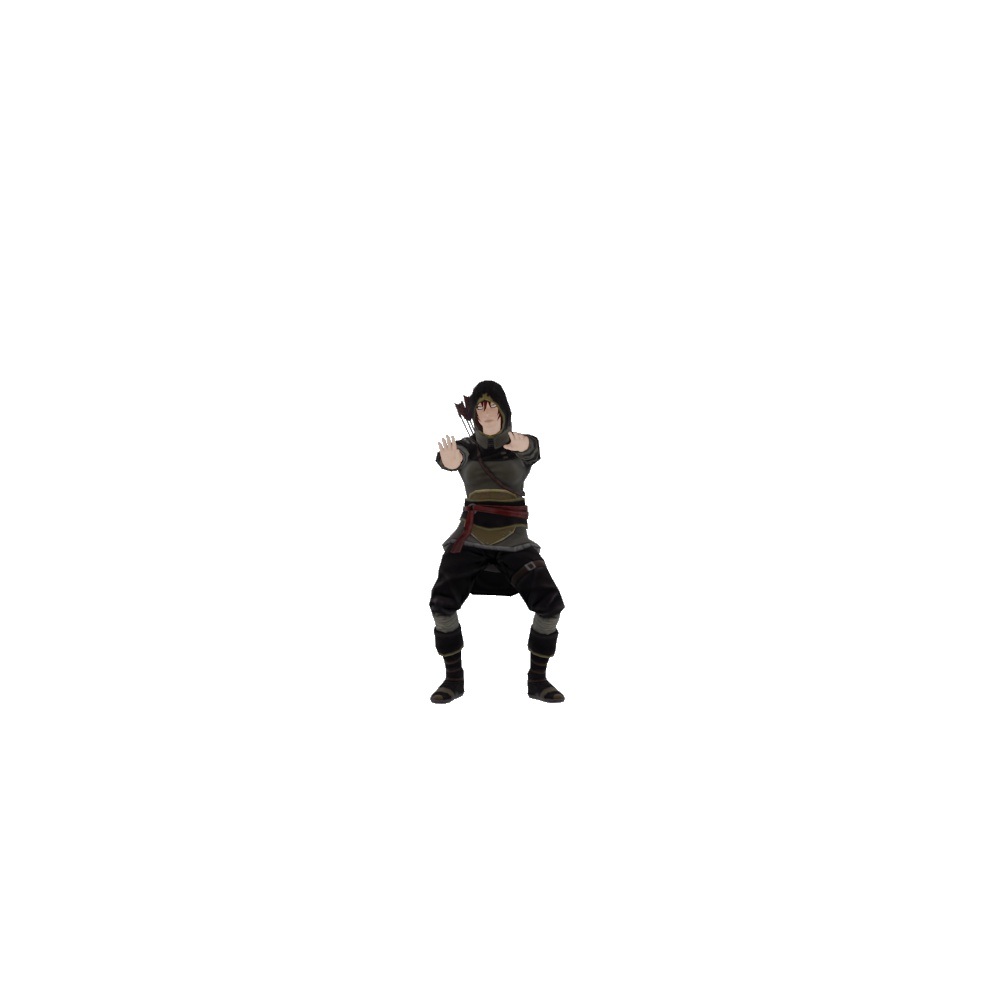}}\\%
{\small Source}
}%
\hfill%
\parbox[t]{\qrtpscale}{
\centering
\fbox{\includegraphics[width=\qrtpscale,trim=353 280 373 300,clip]{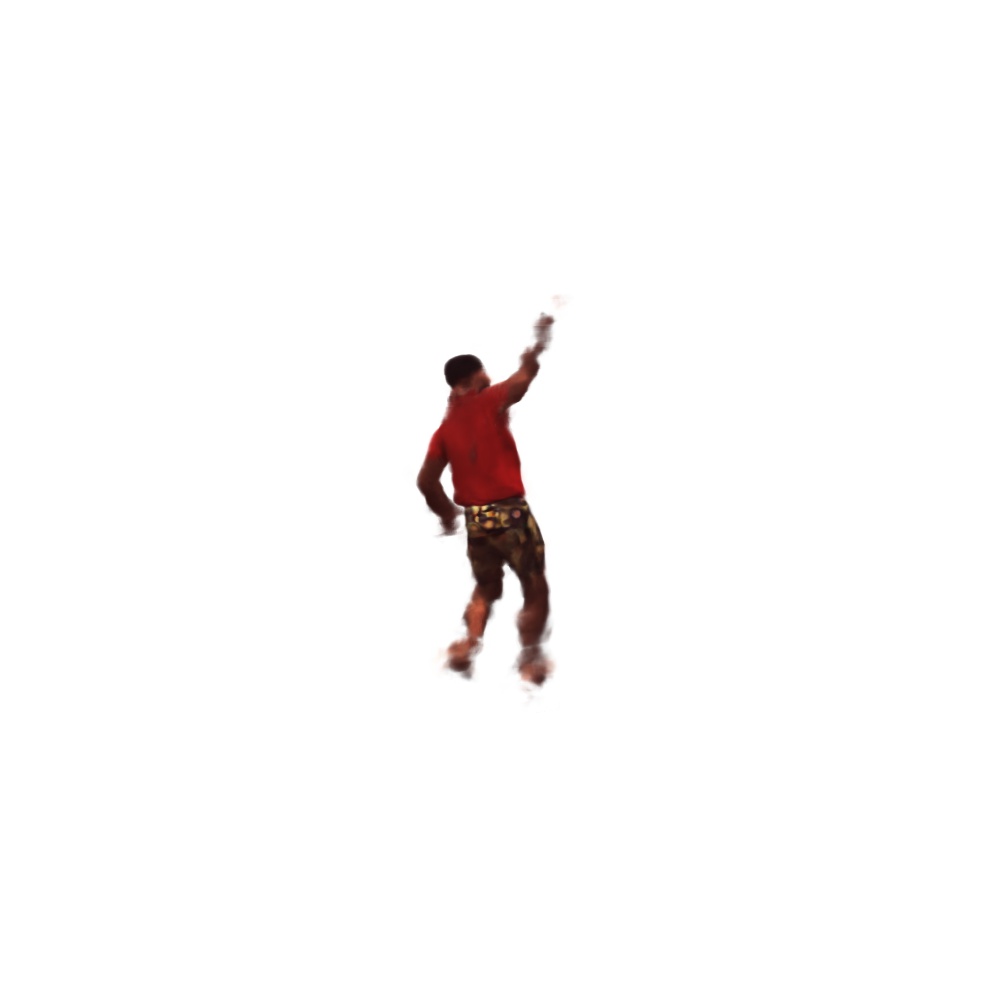}}\\%
\fbox{\includegraphics[width=\qrtpscale,trim=387 280 387 380,clip]{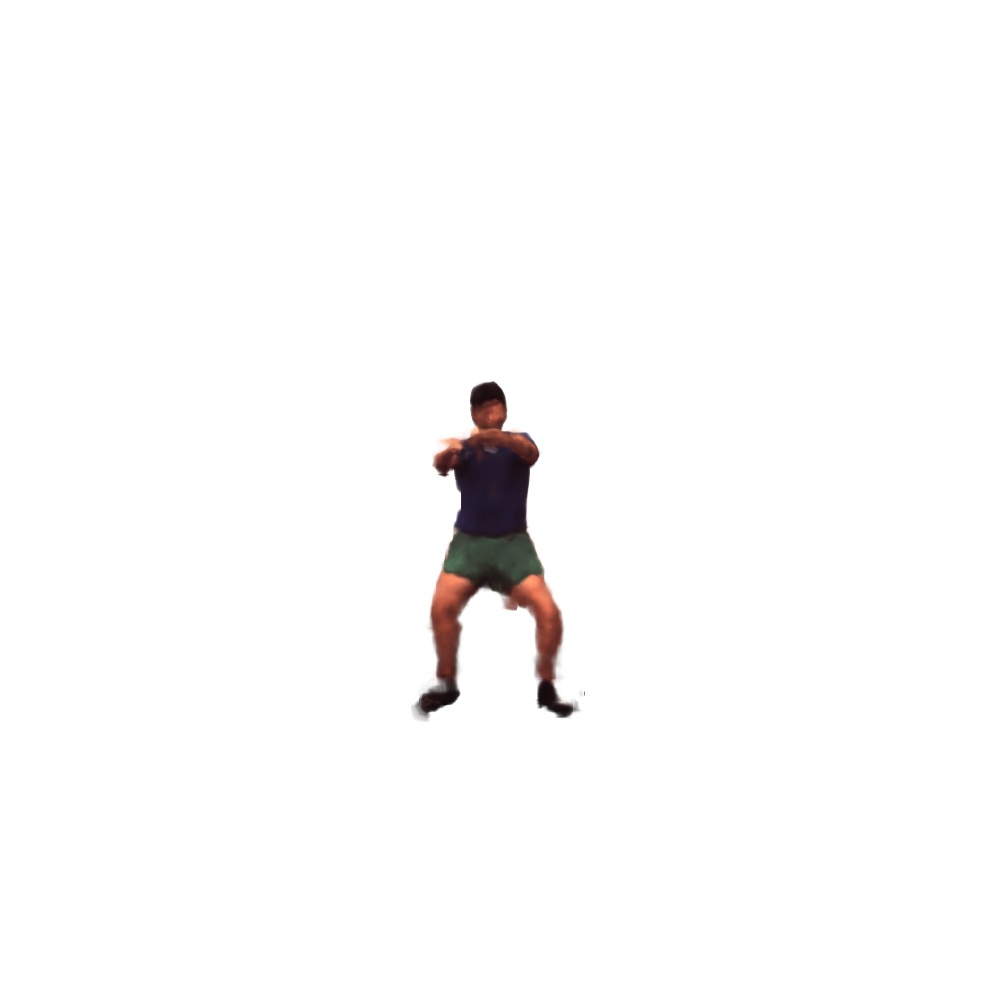}}\\%
{\small NeuralBody}%
}%
\hfill%
\parbox[t]{\qrtpscale}{
\centering
\fbox{\includegraphics[width=\qrtpscale,trim=376 320 376 300,clip]{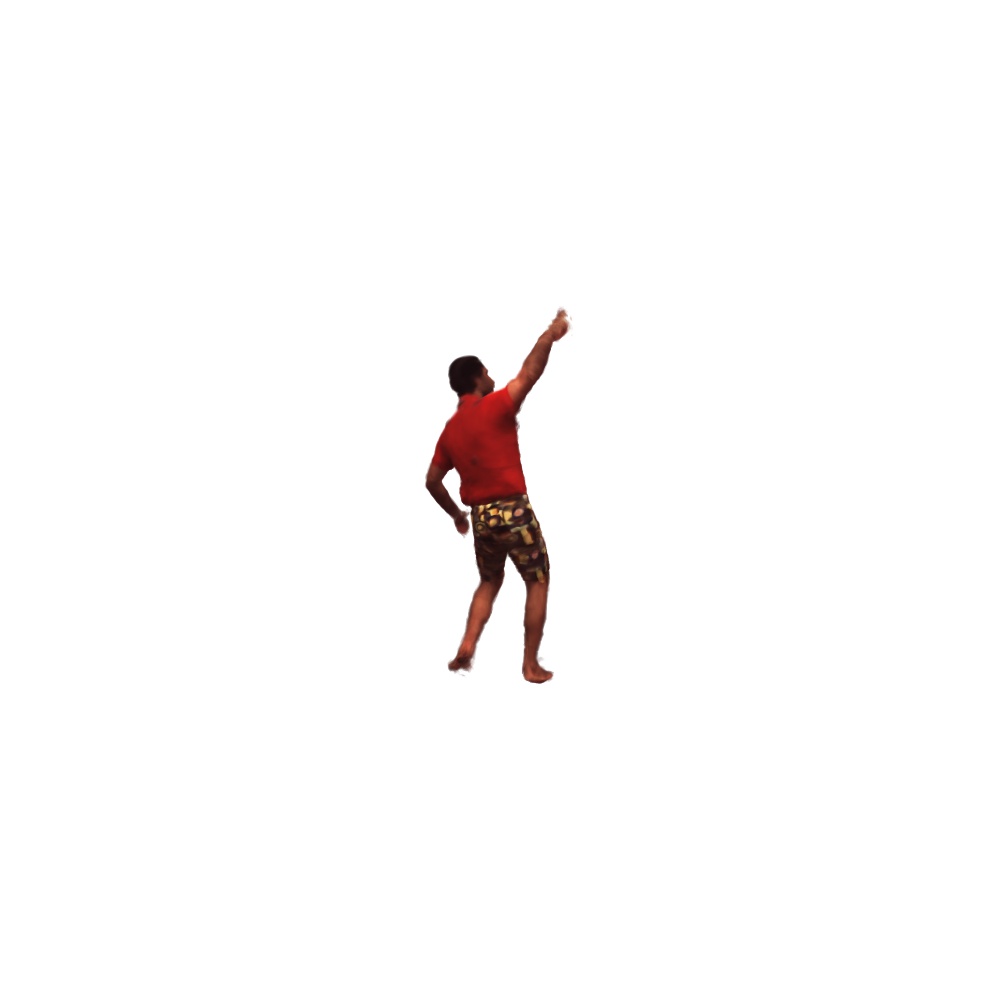}}\\%
\fbox{\includegraphics[width=\qrtpscale,trim=387 280 387 380,clip]{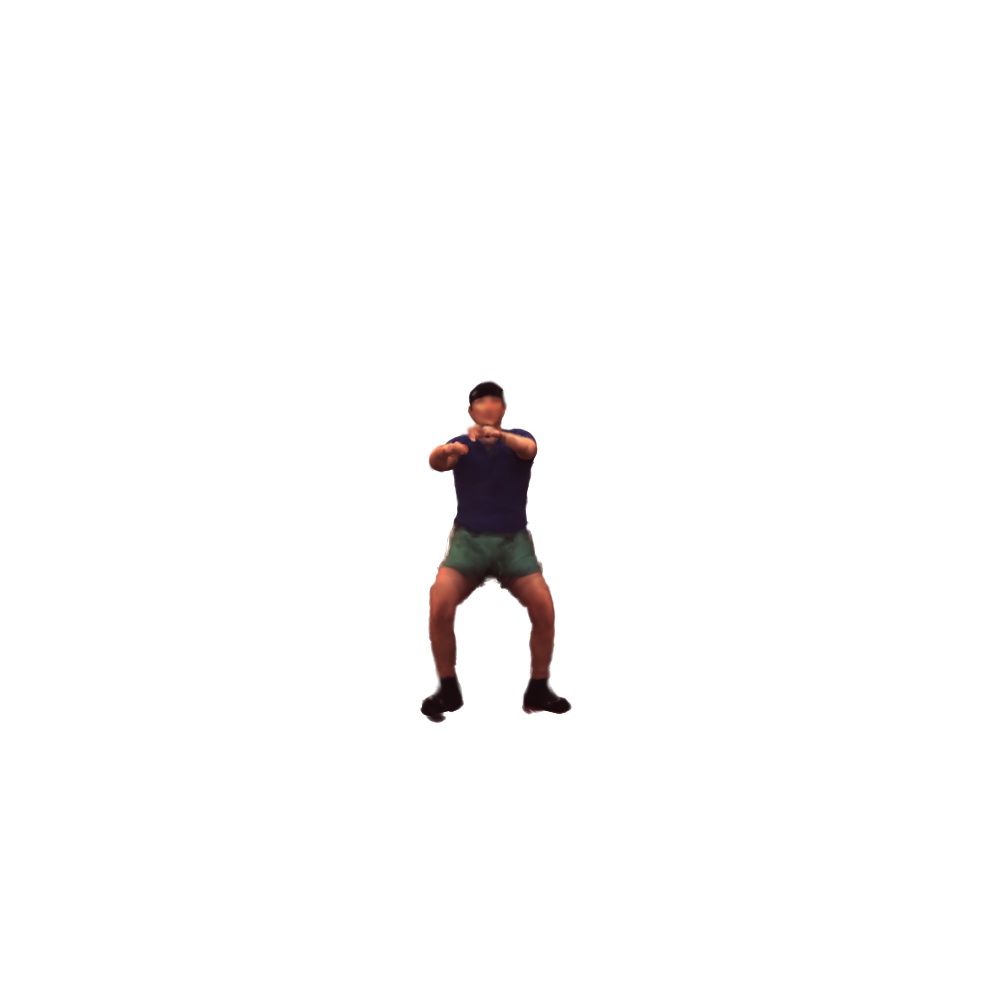}}\\%
{\small Ours }%
}\\%
\centering
\parbox[t]{\qaniscalec}{
\centering
\fbox{\includegraphics[width=\qaniscalec]{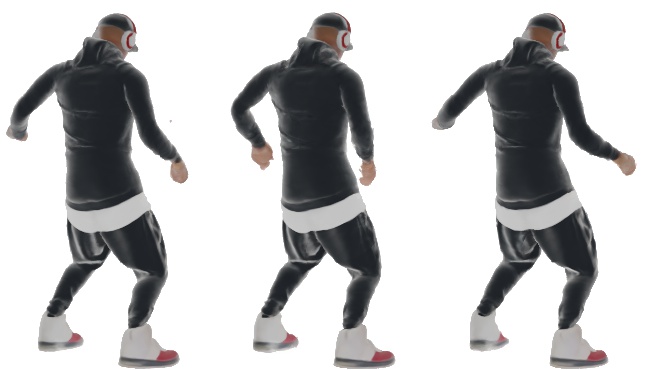}}\\%
}\quad\quad%
\parbox[t]{\qaniscaleb}{
\centering
\fbox{\includegraphics[width=\qaniscaleb]{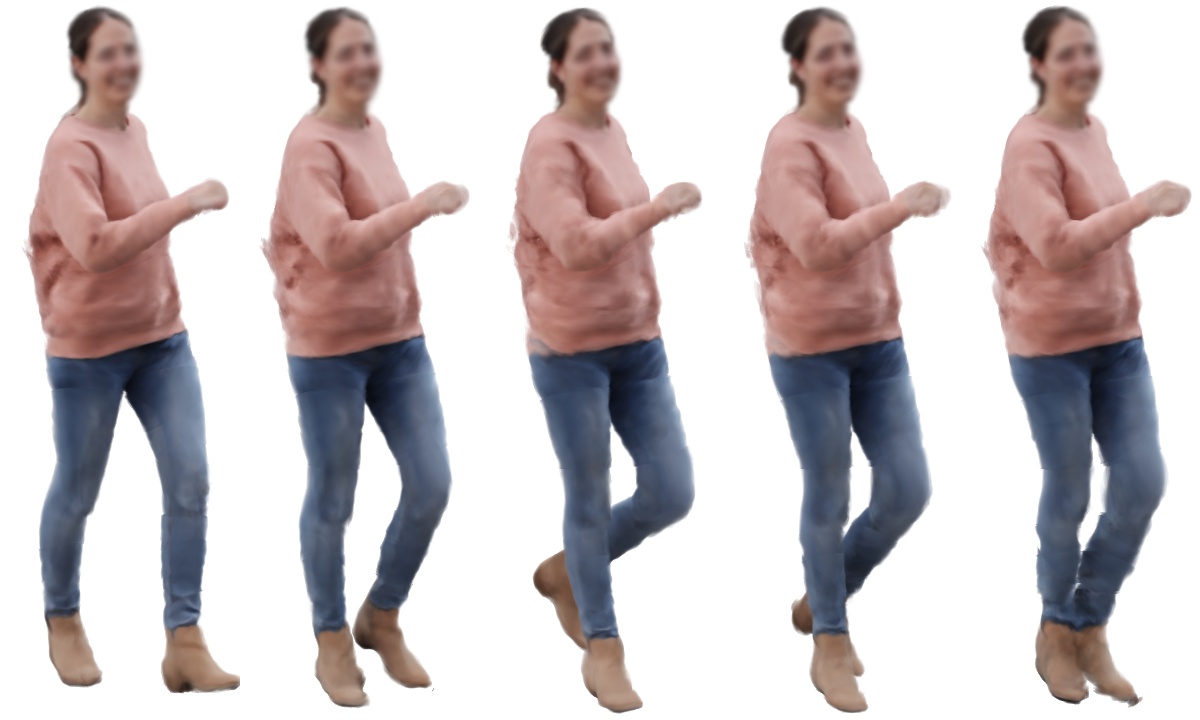}}\\%
}\\%
\caption{\textbf{Motion retargeting and animation.} \textbf{Top rows:} Pose transfer, with the source pose reconstructed by A-NeRF and rendered with different target body models. \textbf{Last row:} Animating the A-NeRF model while keeping the lower or upper part of the body fixed.
}
\label{fig:motion-retarget}
\end{figure*}

\paragraph{Novel-View-Synthesis and Character Animation.}
Our body model is generative, which allows us to train on a single or multiple uncalibrated videos and alter viewpoint and human pose.
\figref{fig:novel-view-synthesis} shows renderings of the same persons from a new camera angle; novel-view-synthesis. 
Likewise, \figref{fig:motion-retarget} demonstrates character animation, where the view is fixed, and the underlying skeleton is re-posed by manually changing joint angles or by transferring the motion between characters.
A-NeRF is the first model that learns a detailed human body model with such capabilities without needing a 3D surface or multi-view supervision. %
While NeuralBody~\cite{peng2020arxiv_neuralbody} can also learn photo-realistic human models from monocular images, their model anchors its representation on the SMPL 3D surface. When trained on our single-view setup with the same noisy estimated poses $\hat{\theta}$ as input, NeuralBody suffers from artifacts and less detail when training, as it assumes 3D ground truth, which is only available in controlled conditions.
The importance of our joint body and pose optimization is further validated by the ablation with refinement disabled (Ours w/o rf.), which similarly produces ghosting artifacts around extremities. 
\begin{table}[t]
\caption{\textbf{Quantitative evaluation on Human3.6M~\cite{ionescu_human36_pami14} and MPI-INF-3DHP~\cite{mpi_3dhp}}. Our test-time pose refinement improves consistently upon the SPIN baseline, with largest improvements for extremities (PA-Wrist).}
\centering
\resizebox{0.980\linewidth}{!}{
\setlength{\tabcolsep}{3pt}
\begin{tabular}{lcccccc}
\toprule
        &  \multicolumn{4}{c}{Human 3.6M} & \multicolumn{2}{c}{MPI-INF-3DHP} \\
        & Protocol \RNum{1} & Protocol \RNum{2} & Prot.~\RNum{2} Wrist & Prot.~\RNum{2} Multi-view & & \\
        \cmidrule{2-7}
Method  & PA-MPJPE$\downarrow$ & PA-MPJPE$\downarrow$ & PA-Wrist$\downarrow$ & PA-MPJPE$\downarrow$ & PA-MPJPE$\downarrow$ & PCK$\uparrow$ \\
\midrule
MotioNet \cite{shi2020motionet} & 54.6 & - & - & - & - & -  \\
\rowcolor{Gray}
HoloPose \cite{guler2019holopose} & 46.5 & - & - & - & - & -\\
VIBE \cite{kocabas2020vibe} & 41.4  & n/a & n/a & n/a & \bf 64.6 & \bf 89.3\\
\rowcolor{Gray}
SPIN \cite{kolotouros2019learning_spin} & \bf 41.1  & - &- & -& 67.5 & 76.4 \\ %
\midrule
Baseline (SPIN github~\cite{kolotouros2019learning_spin}) &  42.7$^\star$  & 41.9$^\star$ & 66.5$^\star$ & 34.0$^\star$ & 68.2$^\star$ & 79.3$^\star$\\
\rowcolor{Gray}
SPIN-SMPLify$^{\dagger}$\cite{bogo2016keep_smplify,kolotouros2019learning_spin} & 57.7 & 59.2 & 100.9 & -& - & - \\ %
\bf Ours (w/o smoothness prior) &  39.4  & \bf 39.6 &\bf 57.3 & \bf 28.0 & 66.9 & 80.4\\
\rowcolor{Gray}
\bf Ours  & \bf 39.3 & n/a & n/a & n/a& \bf 66.8 & \bf 80.4 \\

\bottomrule
\multicolumn{6}{l}{\footnotesize$^\star$ Reevaluation of publicly-available model because missing evaluation protocols or not reproducible.}\\%
\multicolumn{6}{l}{\footnotesize$^\dagger$ We refine the SPIN estimated pose using SMPLify. We adopt the implementation from the SPIN repository.}%

\end{tabular}%
\label{tab:h36m-quant}
}%
\end{table}

\paragraph{Human Pose Estimation.}
Training A-NeRF includes a form of test-time optimization (see \figref{fig:teaser}, right), only 
the initialization from \cite{kolotouros2019learning_spin} is trained supervised on the Human3.6M training set. 
\tabref{tab:h36m-quant} shows that on Human 3.6M, A-NeRF reaches comparable results with other single-view approaches, and achieves a $8.0\%$ improvement in PA-MPJPE upon the baseline used for pose initialization on Protocol~\RNum{1} ($42.7\rightarrow39.3$) and $5.5\%$ on Protocol~\RNum{2} ($41.9\rightarrow39.6$). %
Note that these are average numbers across all joints, including easy-to-predict hip, shoulder, and head joints. Our largest gains are on the extremities, e.g., with an improvement of $14\%$ (9.2 mm) for the wrist joint on Protocol II (PA-Wrist).
We also compare to applying SMPLify~\cite{bogo2016keep_smplify}, a method that refines 3D poses using 2D joint locations (estimated with~\cite{cao2018openpose}) as constraints, at test time. It tends to explain the 2D joint estimates perfectly but degrades the 3D pose. The final pose estimations become less accurate than the initial ones. In contrast, A-NeRF optimizes the poses by implicitly minimizing the disagreement among the 3D body representation in different images, and thus achieves better performance.
On MPI-INF-3DHP, \tabref{tab:h36m-quant} shows the results averaged over the 6 test subjects from MPI-INF-3DHP. Despite having a low number of frames and human poses available for learning the skeleton-relative encoding, A-NeRF still provides moderate improvements over the baseline estimations.
\begin{table}[!t]
\caption{\textbf{Visual quality evaluation on the Human3.6M~\cite{ionescu_human36_pami14} and MonoPerfCap~\cite{xu2018monoperfcap} held-out sets}. Our full A-NeRF model significantly improves the visual quality. The body model itself attains a higher quality than NeuralBody (2nd vs. 3rd row), and additional detail is gained with the proposed pose refinement (last row).}%
\centering
\resizebox{0.82\linewidth}{!}{
\setlength{\tabcolsep}{3pt}
\begin{tabular}{lcccc}
\toprule
  & \multicolumn{2}{c}{MonoPerfCap} & \multicolumn{2}{c}{Human 3.6M} \\
 \midrule
  &   PSNR~$\uparrow$ & SSIM~$\uparrow$  &   PSNR~$\uparrow$ & SSIM~$\uparrow$  \\ 
\midrule
NeuralBody, driving motion from \cite{kolotouros2019learning_spin}  & 21.80 & 0.8476 & 22.08  & 0.8766 \\ %
\rowcolor{Gray}
NeuralBody, driving motion from A-NeRF refinement  & 21.75 & 0.8468 & 22.55  & 0.8782 \\
A-NeRF w/o pose refinement & 21.99  & 0.8405 & 23.33  & 0.8776\\
\rowcolor{Gray}
A-NeRF (Our full model) & \bf 24.39 &  \bf 0.8851  & \bf 27.45  & \bf 0.9277  \\
    \bottomrule
\end{tabular}
\label{tab:ablation-refinement}
}
\end{table}

\paragraph{Video-based volumetric reconstruction.} Figure~\ref{fig:geometry main} visualizes the learned density using Marching Cubes~\cite{lorensen1987marching}, with voxel grid resolution of 256 and density threshold 10.
Despite only learned from monocular videos (no stereo, depth camera, or multi-view constraints) and without using a pre-defined template model, A-NeRF reconstructs a detailed volumetric body with details that could not be captured by offsets to a parametric surface model, such as the head phones and basebal cap in the last row of Figure~\ref{fig:geometry main}. Note that no geometric smoothness term is enforced.
\newlength\qualgeoscale
\setlength\qualgeoscale{0.132\textwidth}
\begin{figure}
\setlength{\fboxrule}{0pt}%
\centering
\parbox[t]{\qualgeoscale}{%
    \vspace{0mm}\centering%
    \fbox{\includegraphics[width=\qualgeoscale]{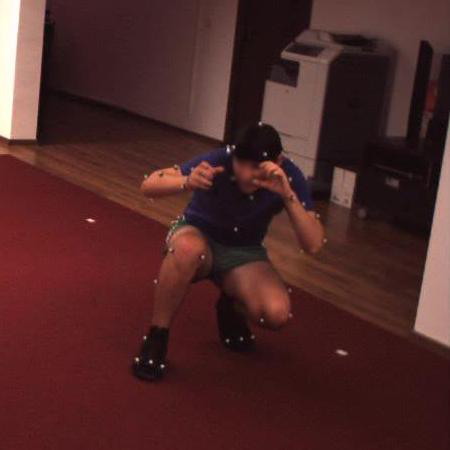}}\\%
    \fbox{\includegraphics[width=\qualgeoscale]{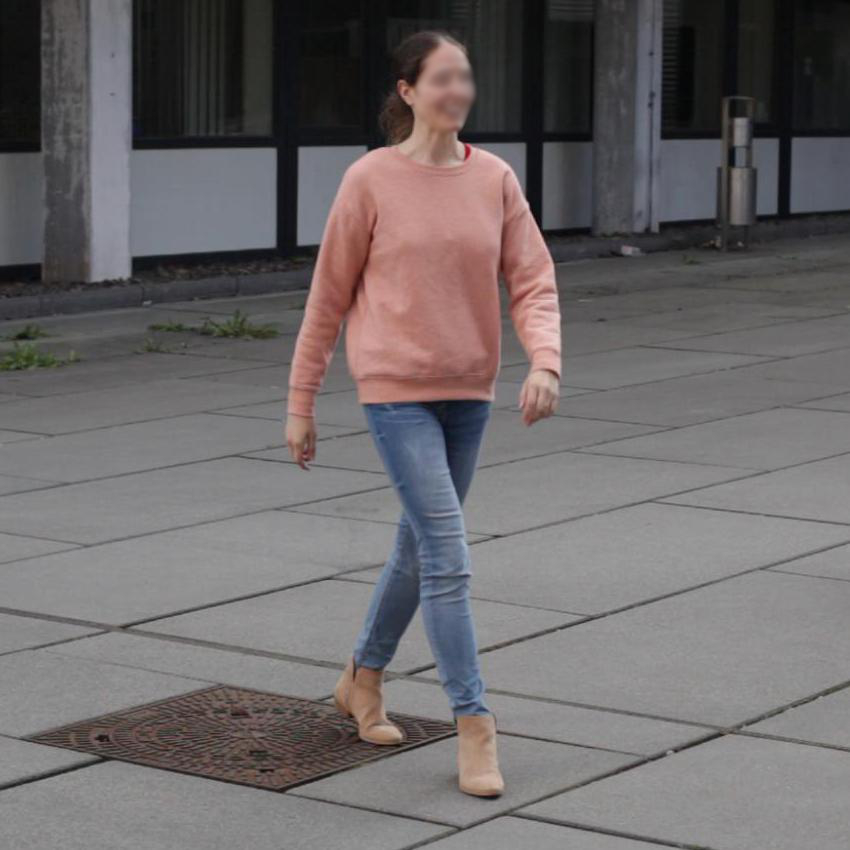}}\\%
    \fbox{\includegraphics[width=\qualgeoscale]{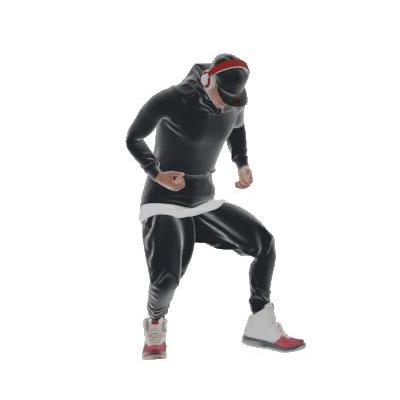}}\\%
    {\footnotesize Ref. Image}\\
}%
\hfill%
\parbox[t]{\qualgeoscale}{%
    \vspace{0mm}\centering%
    \fbox{\includegraphics[width=\qualgeoscale]{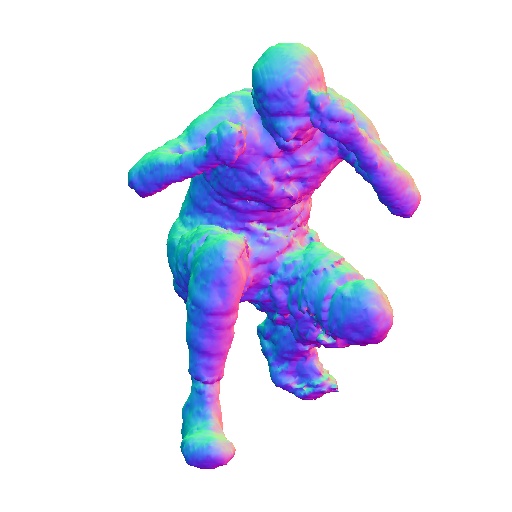}}\\%
    \fbox{\includegraphics[width=\qualgeoscale]{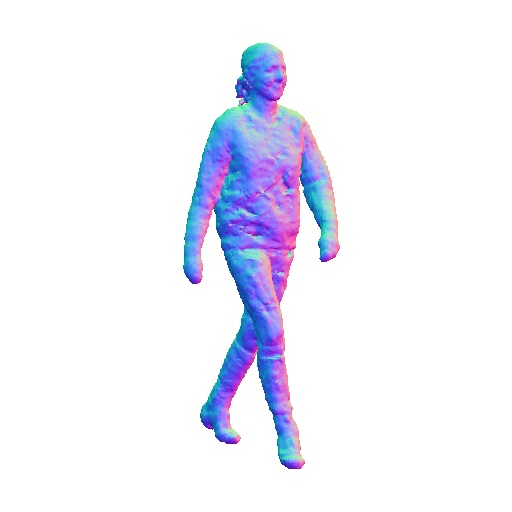}}\\%
    \fbox{\includegraphics[width=\qualgeoscale]{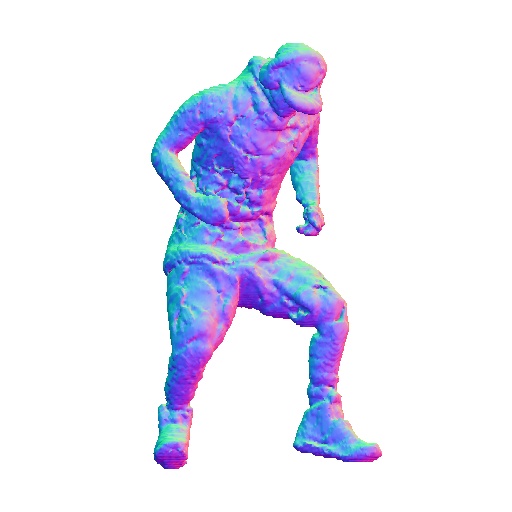}}\\%
    {{\footnotesize Geometry $\rightarrow$}}
}%
\hfill%
\parbox[t]{\qualgeoscale}{%
    \vspace{0mm}\centering%
    \fbox{\includegraphics[width=\qualgeoscale]{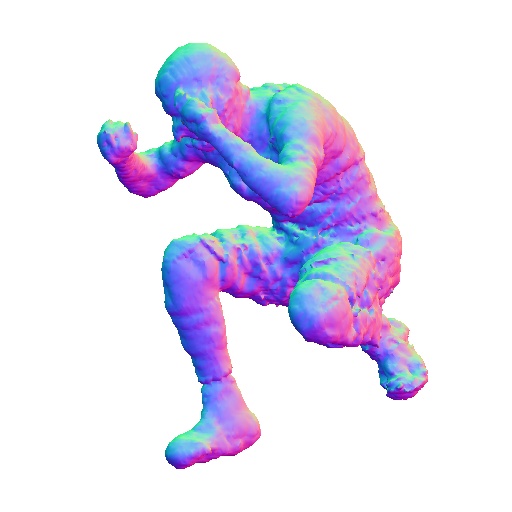}}\\%
    \fbox{\includegraphics[width=\qualgeoscale]{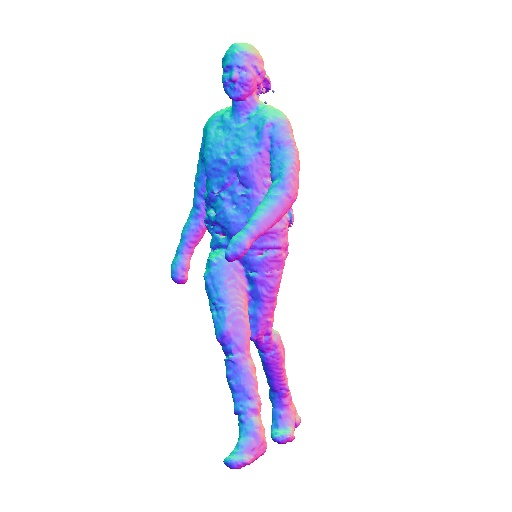}}\\%
    \fbox{\includegraphics[width=\qualgeoscale]{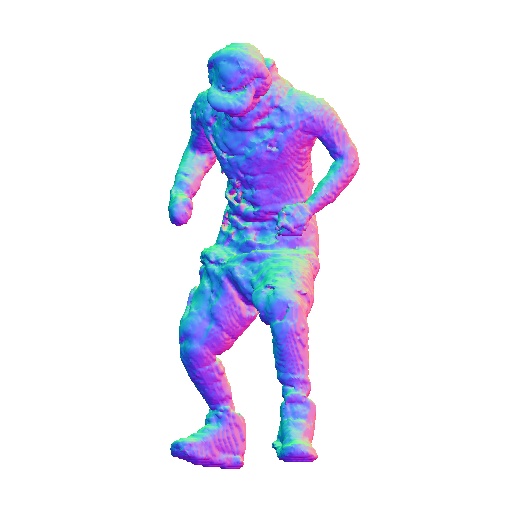}}\\%
}%
\hfill%
\parbox[t]{\qualgeoscale}{%
    \vspace{0mm}\centering%
    \fbox{\includegraphics[width=\qualgeoscale]{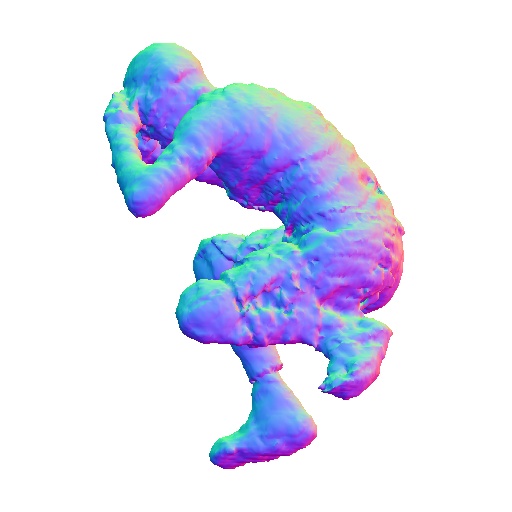}}\\%
    \fbox{\includegraphics[width=\qualgeoscale]{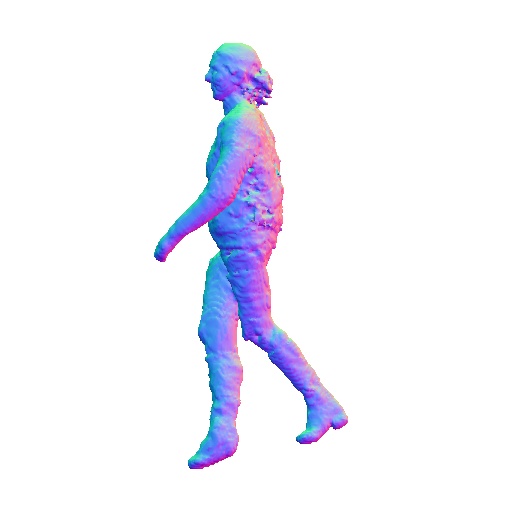}}\\%
    \fbox{\includegraphics[width=\qualgeoscale]{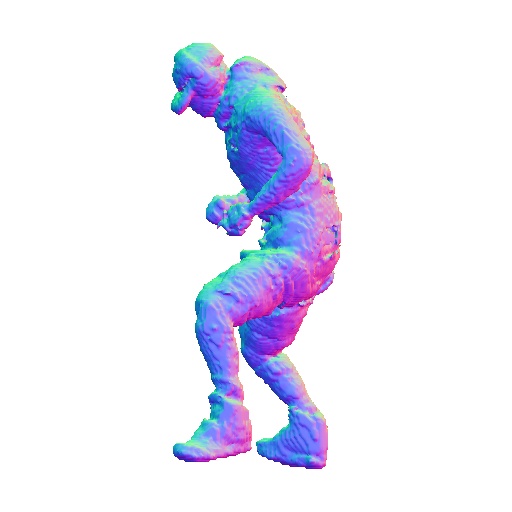}}\\%
    {\footnotesize Novel views}
}%
\hfill%
\parbox[t]{\qualgeoscale}{%
    \vspace{0mm}\centering%
    \fbox{\includegraphics[width=\qualgeoscale]{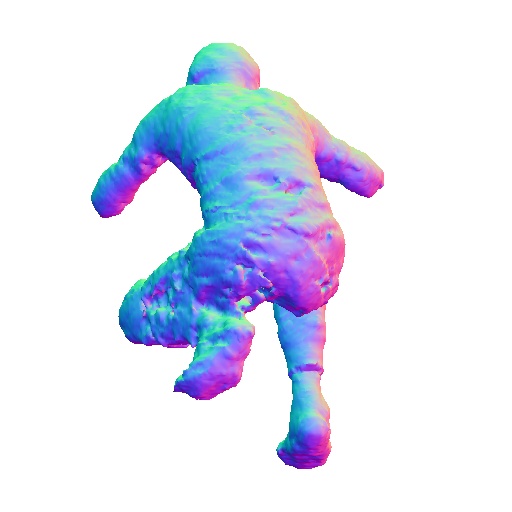}}\\%
    \fbox{\includegraphics[width=\qualgeoscale]{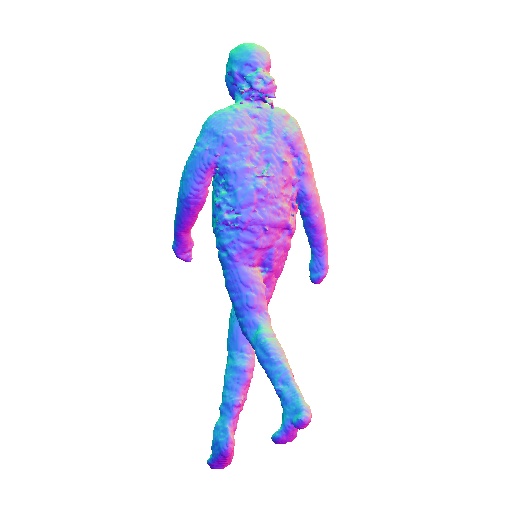}}\\%
    \fbox{\includegraphics[width=\qualgeoscale]{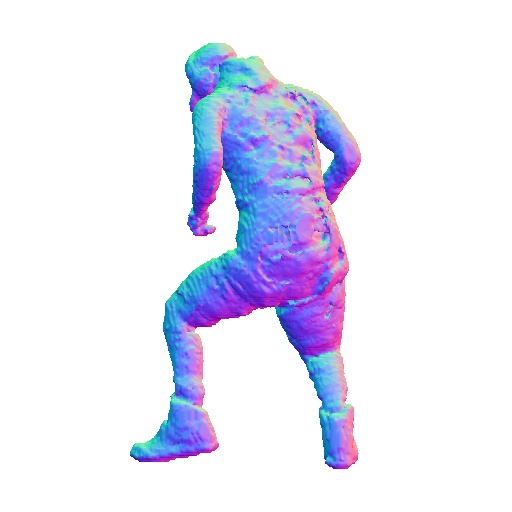}}\\%
}%
\hfill%
\parbox[t]{\qualgeoscale}{%
    \vspace{0mm}\centering%
    \fbox{\includegraphics[width=\qualgeoscale]{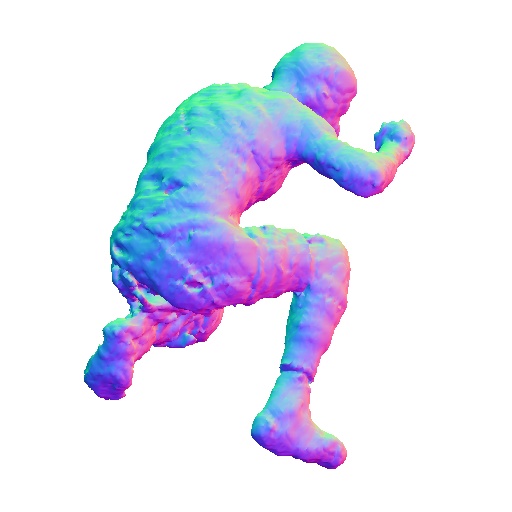}}\\%
    \fbox{\includegraphics[width=\qualgeoscale]{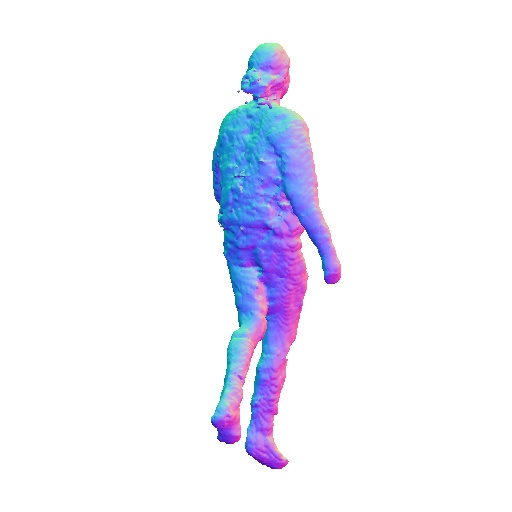}}\\%
    \fbox{\includegraphics[width=\qualgeoscale]{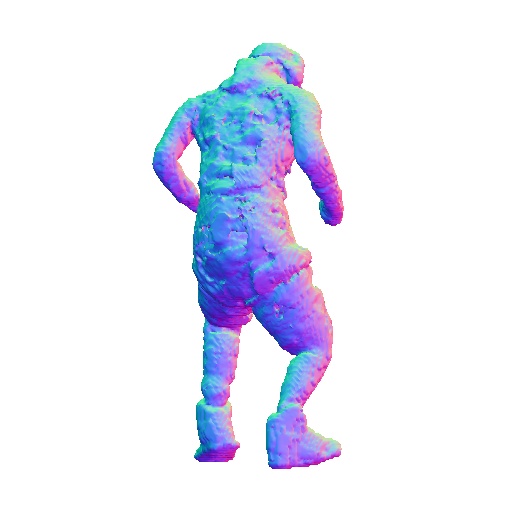}}\\%
}%
\hfill%
\parbox[t]{\qualgeoscale}{%
    \vspace{0mm}\centering%
    \fbox{\includegraphics[width=\qualgeoscale]{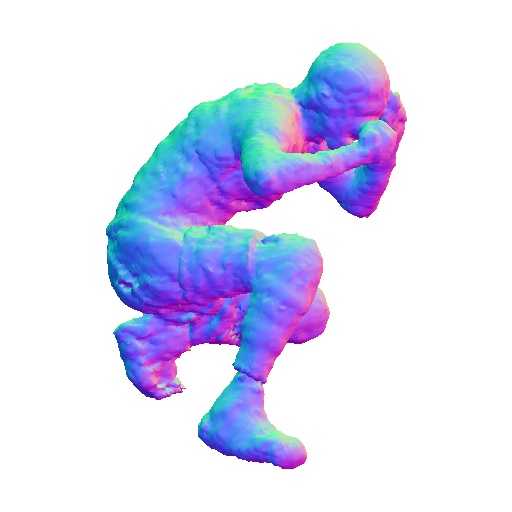}}\\%
    \fbox{\includegraphics[width=\qualgeoscale]{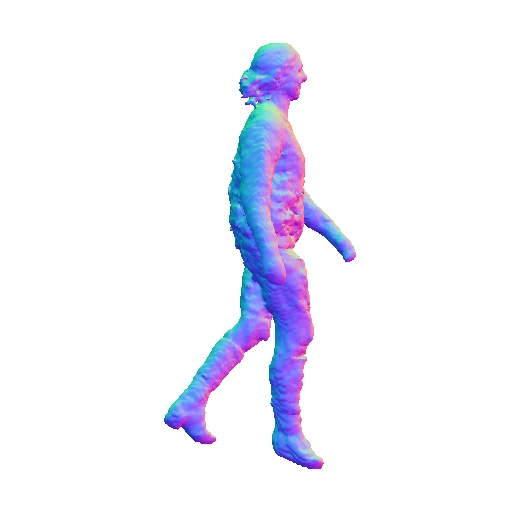}}\\%
    \fbox{\includegraphics[width=\qualgeoscale]{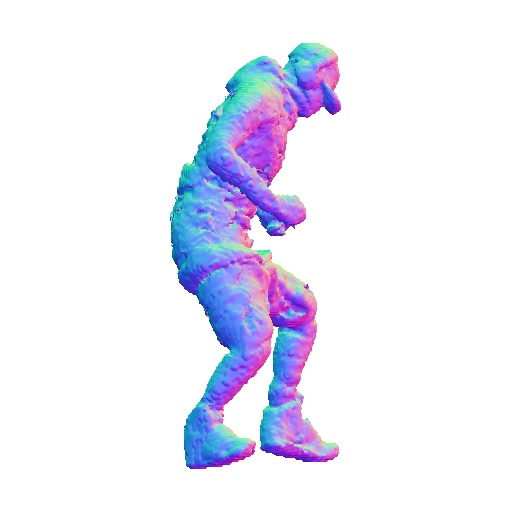}}\\%
}%
\caption{\tb{Geometry.} The isosurface of our density model is rendered from multiple unseen views. A-NeRF learns plausible geometry without using explicit surface templates and accurate initial poses. }
\label{fig:geometry main}
\end{figure}

\paragraph{Multi-view extension.} A-NeRF can also leverage multi-view refinement (\tabref{tab:h36m-quant}, Protocol~\RNum{2} Multi-view), even without access to ground truth camera calibration (see supplementary).

\paragraph{Visual Quality Comparison.}
We report the results in~\tabref{tab:ablation-refinement}. We compare to NeuralBody with both the initial estimates from~\cite{kolotouros2019learning_spin} and our refined pose since NeuralBody has no refinement step. Compared to both variants, our A-NeRF shows significantly better reconstruction performance on held-out poses. 
We observe that NeuralBody models can retain the facial features and hands as they anchor their representation on a 3D surface model.  However, the rendered limbs and faces are blurry and distorted.
Results are similar to training A-NeRF without refinement (see~\figref{fig:qual-heldout}). As the estimation for these joints is often inaccurate and noisy, the models without pose refinement simply learn to predict mean pixel values. To conclude, it is important to train with pose refinement, with which A-NeRF suffers less from artifacts with overall better visual quality. %
\newlength\qualheldoutscale
\setlength\qualheldoutscale{0.08\textheight}
\newlength\qualheldoutvspace
\setlength\qualheldoutvspace{-2.0mm}
\begin{figure}
\setlength{\fboxrule}{0pt}%
\centering
\hfill\hspace{1mm}%
\parbox[t]{\qualheldoutscale}{%
    \vspace{0mm}\centering%
    \fbox{\includegraphics[height=\qualheldoutscale]{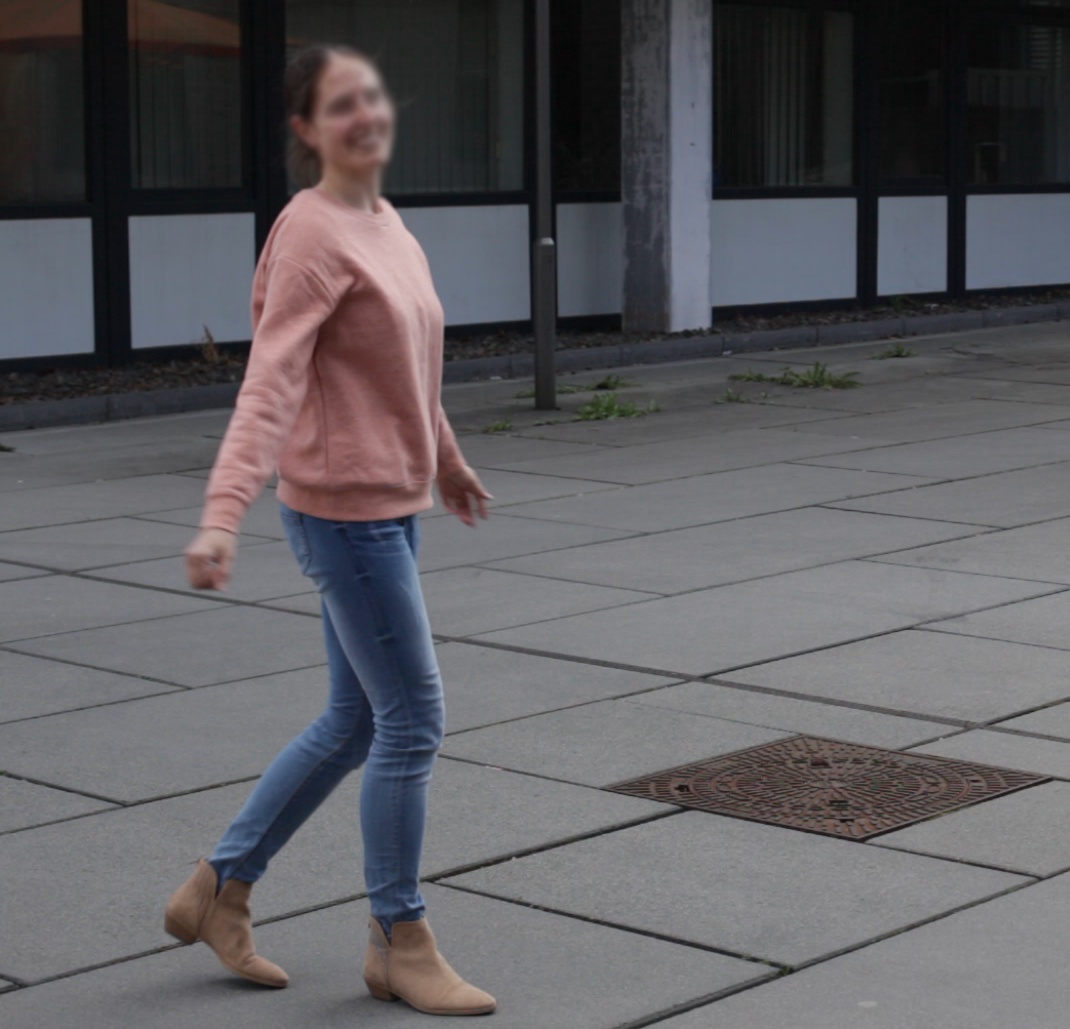}}\\%
     {\footnotesize Ground Truth}\\
    \vspace{0.090\qualheldoutscale}
    {\footnotesize Photometric error\\($\rightarrow$)}\\
    \vspace{0.070\qualheldoutscale}
    \fbox{\includegraphics[height=\qualheldoutscale]{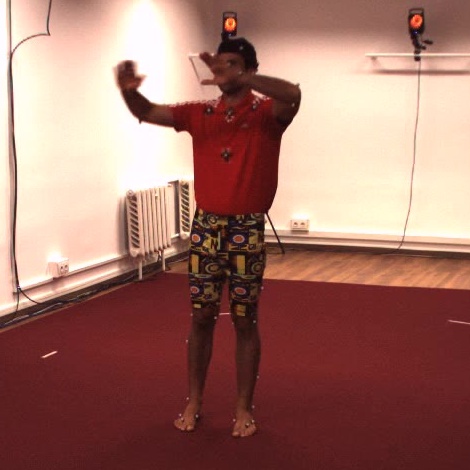}}\\%
    {\footnotesize Ground Truth}\\
    \vspace{0.090\qualheldoutscale}
    {\footnotesize Photometric error\\($\rightarrow$)}\\
    \vspace{0.000\qualheldoutscale}
}%
\hfill%
\parbox[t]{\qualheldoutscale}{%
    \vspace{0mm}\centering%
    \fbox{\includegraphics[height=\qualheldoutscale]{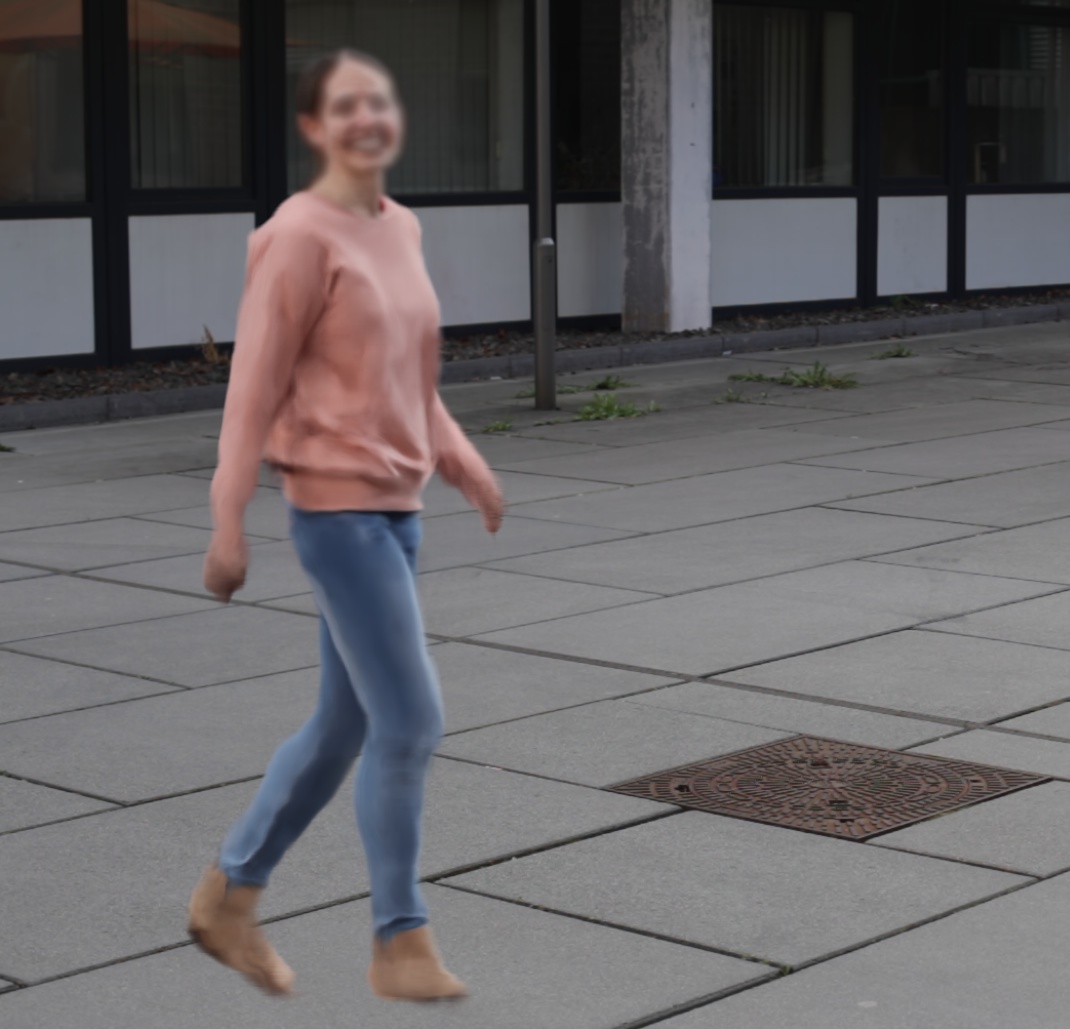}}\\%
    \vspace{\qualheldoutvspace}%
    \fbox{\includegraphics[height=\qualheldoutscale]{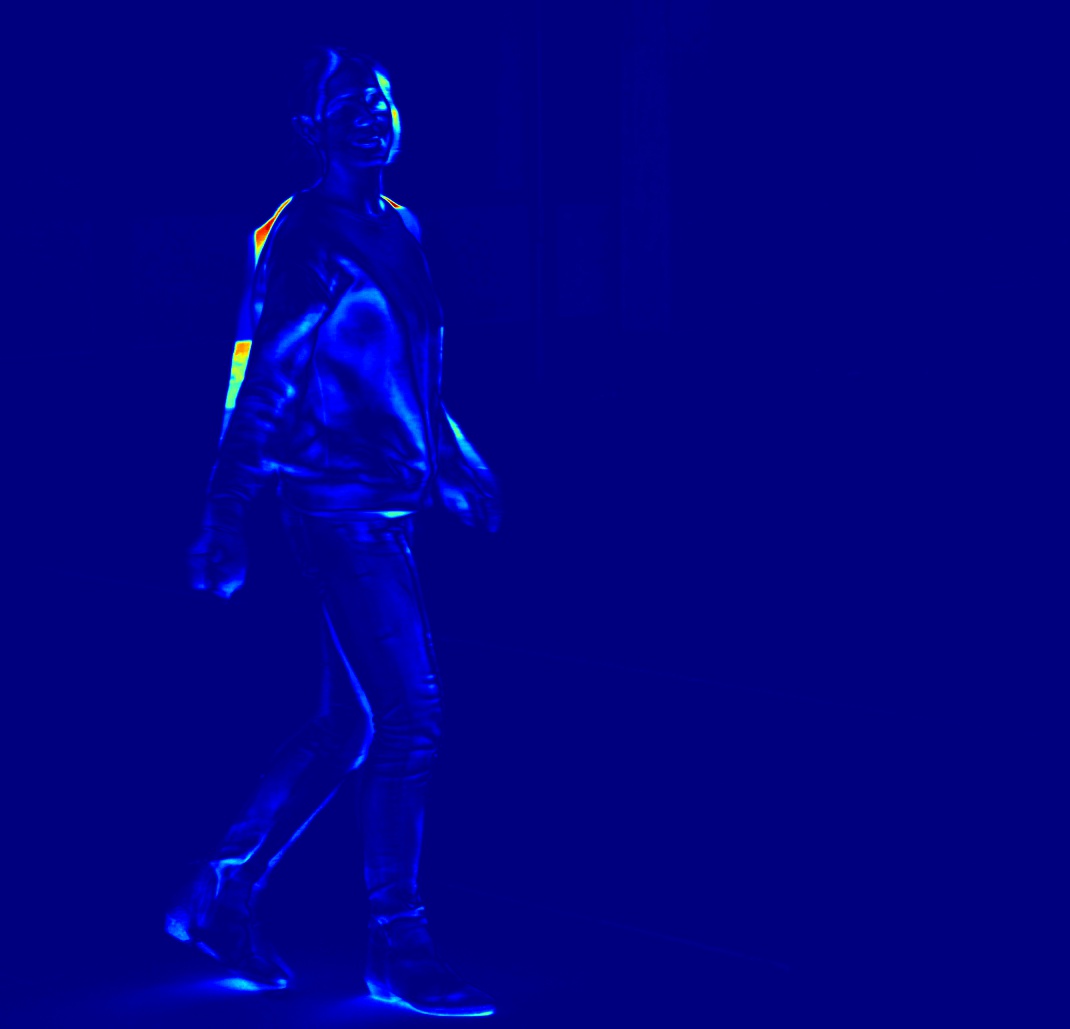}}\\%
    \vspace{\qualheldoutvspace}%
    \fbox{\includegraphics[height=\qualheldoutscale]{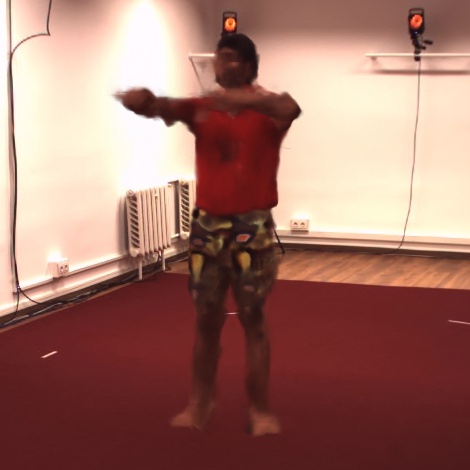}}\\%
    \vspace{\qualheldoutvspace}%
    \fbox{\includegraphics[height=\qualheldoutscale]{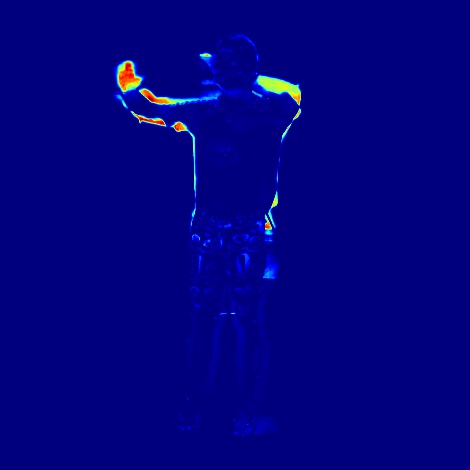}}\\%
    {{\footnotesize NeuralBody} w/ pose from~\cite{kolotouros2019learning_spin}}
}%
\hfill%
\parbox[t]{\qualheldoutscale}{%
    \vspace{0mm}\centering%
    \fbox{\includegraphics[height=\qualheldoutscale]{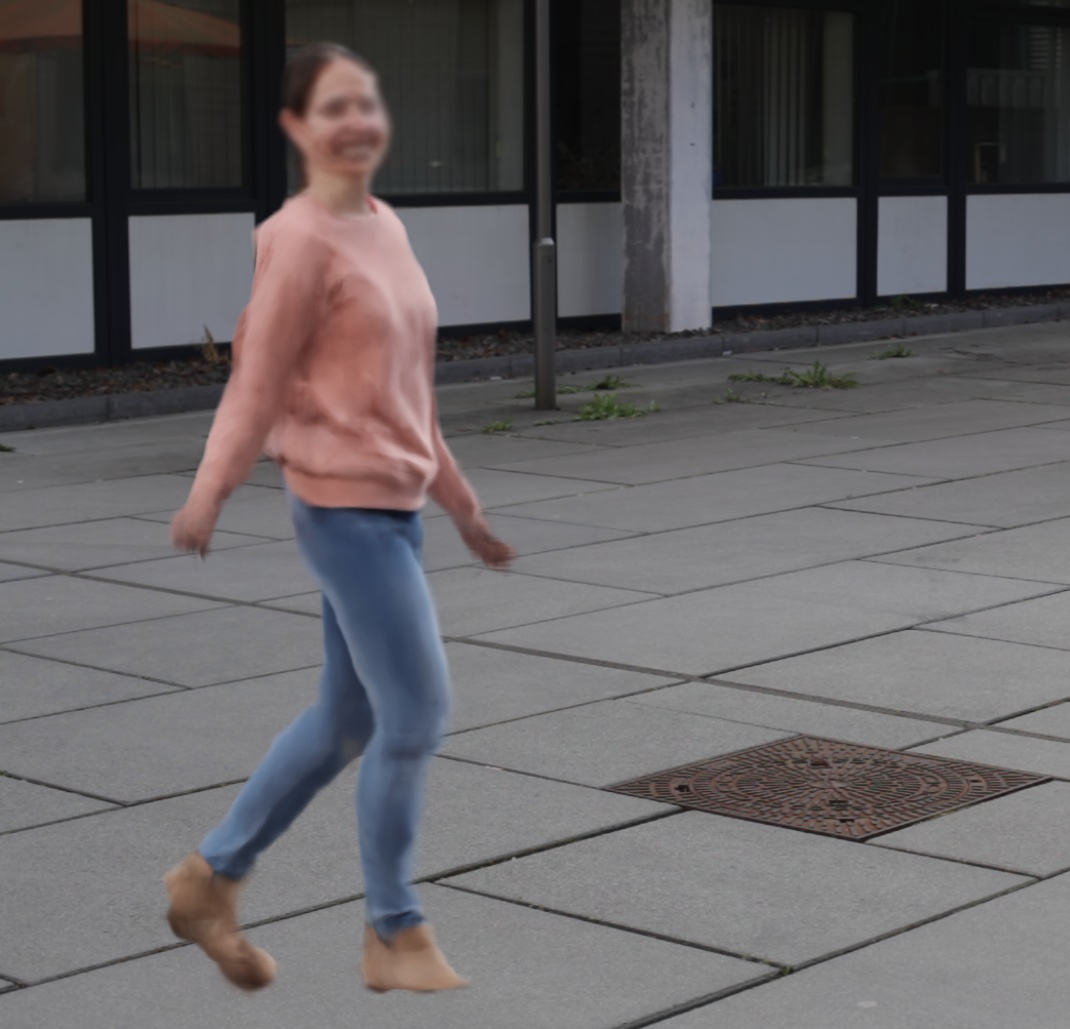}}\\%
    \vspace{\qualheldoutvspace}%
    \fbox{\includegraphics[height=\qualheldoutscale]{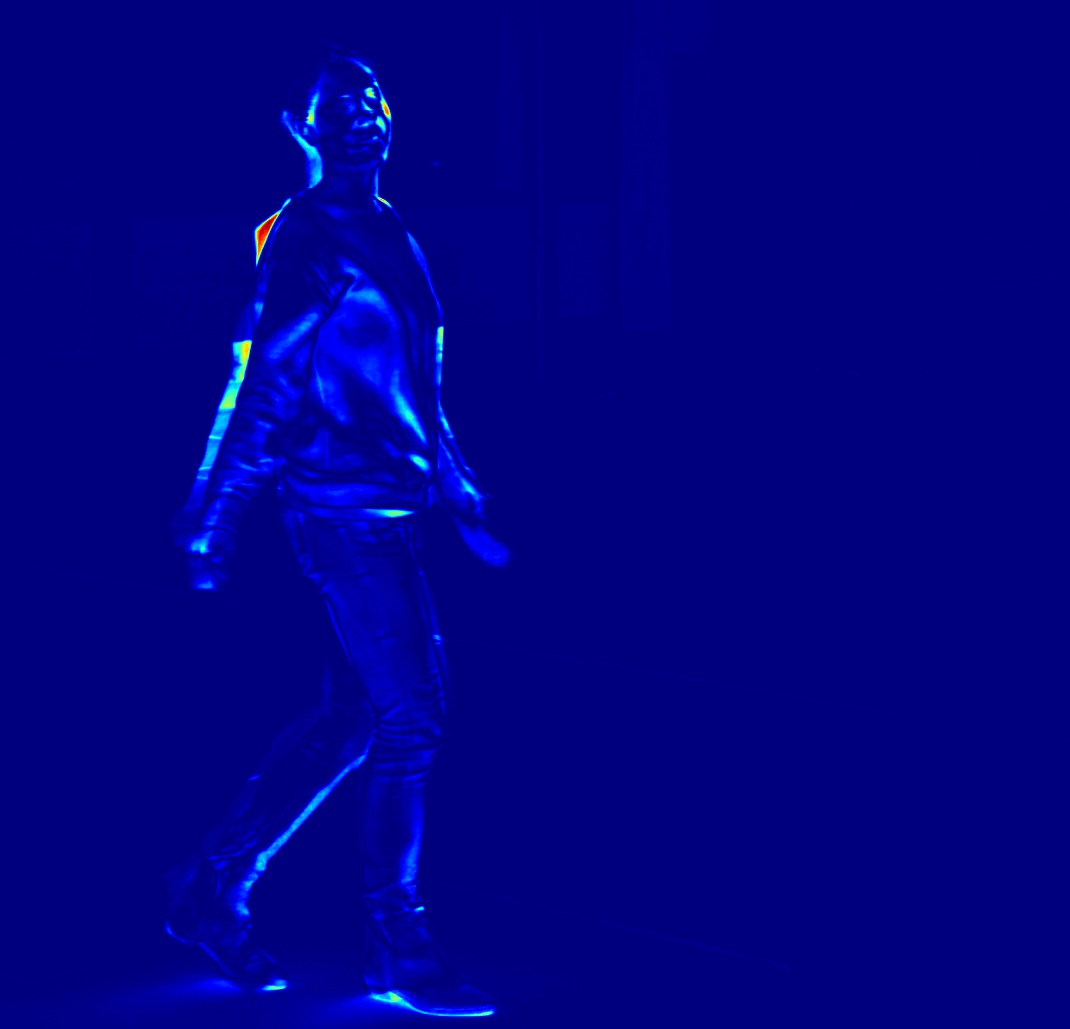}}\\%
    \vspace{\qualheldoutvspace}%
    \fbox{\includegraphics[height=\qualheldoutscale]{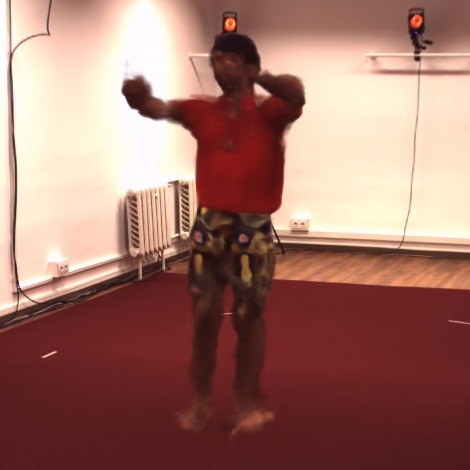}}\\%
    \vspace{\qualheldoutvspace}%
    \fbox{\includegraphics[height=\qualheldoutscale]{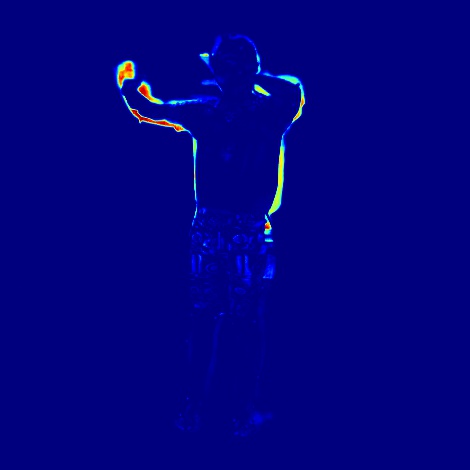}}\\%
    {\footnotesize NeuralBody w/ pose from A-NeRF rf.
    }
}%
\hfill%
\parbox[t]{\qualheldoutscale}{%
    \vspace{0mm}\centering%
    \fbox{\includegraphics[height=\qualheldoutscale]{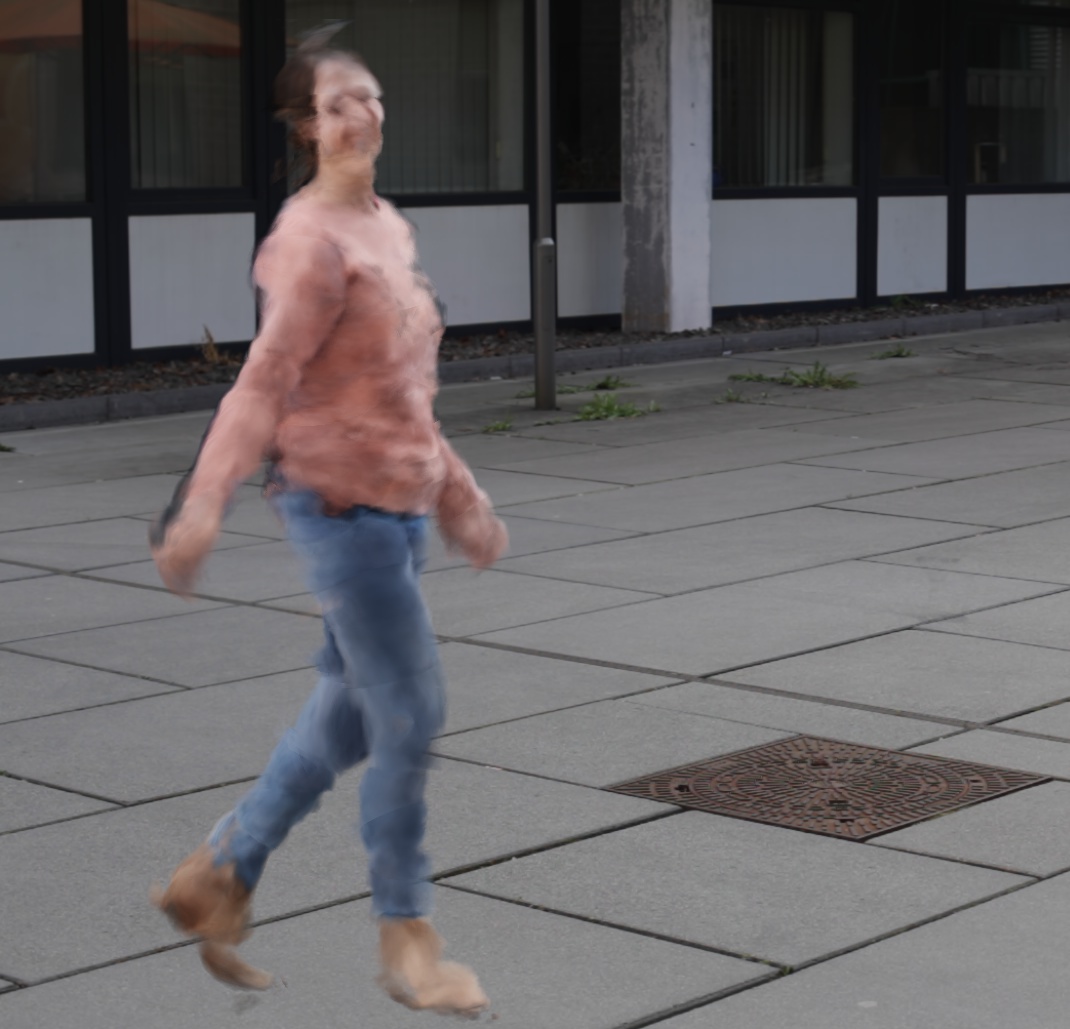}}\\%
    \vspace{\qualheldoutvspace}%
    \fbox{\includegraphics[height=\qualheldoutscale]{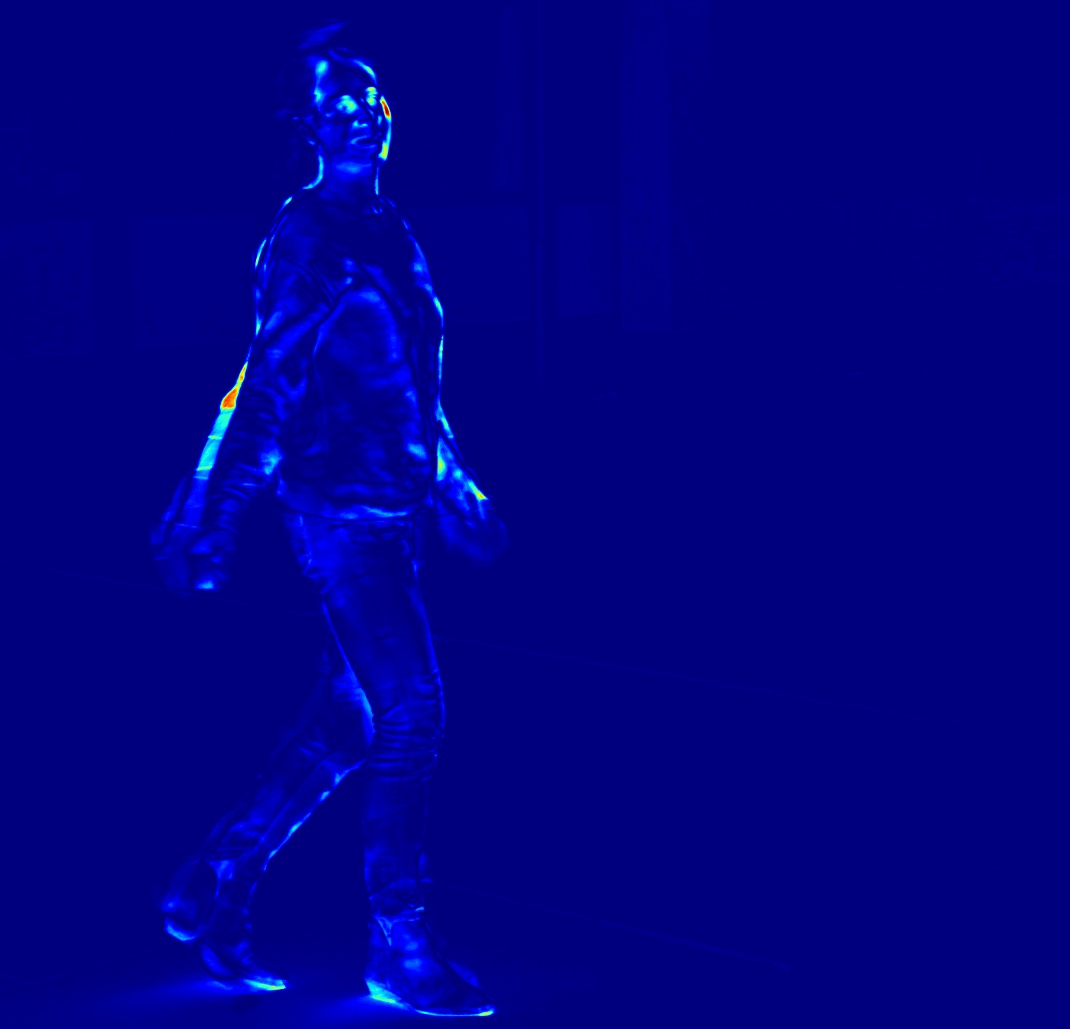}}\\%
    \vspace{\qualheldoutvspace}%
    \fbox{\includegraphics[height=\qualheldoutscale]{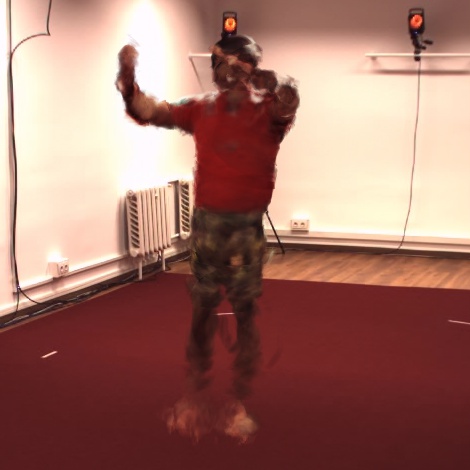}}\\%
    \vspace{\qualheldoutvspace}%
    \fbox{\includegraphics[height=\qualheldoutscale]{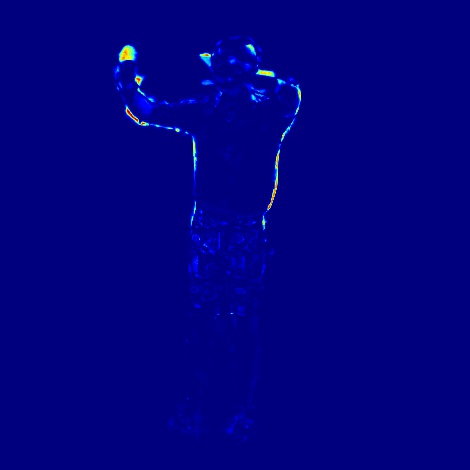}}\\%
    {\footnotesize A-NeRF w/o rf.}
}%
\hfill%
\parbox[t]{\qualheldoutscale}{%
    \vspace{0mm}\centering%
    \fbox{\includegraphics[height=\qualheldoutscale]{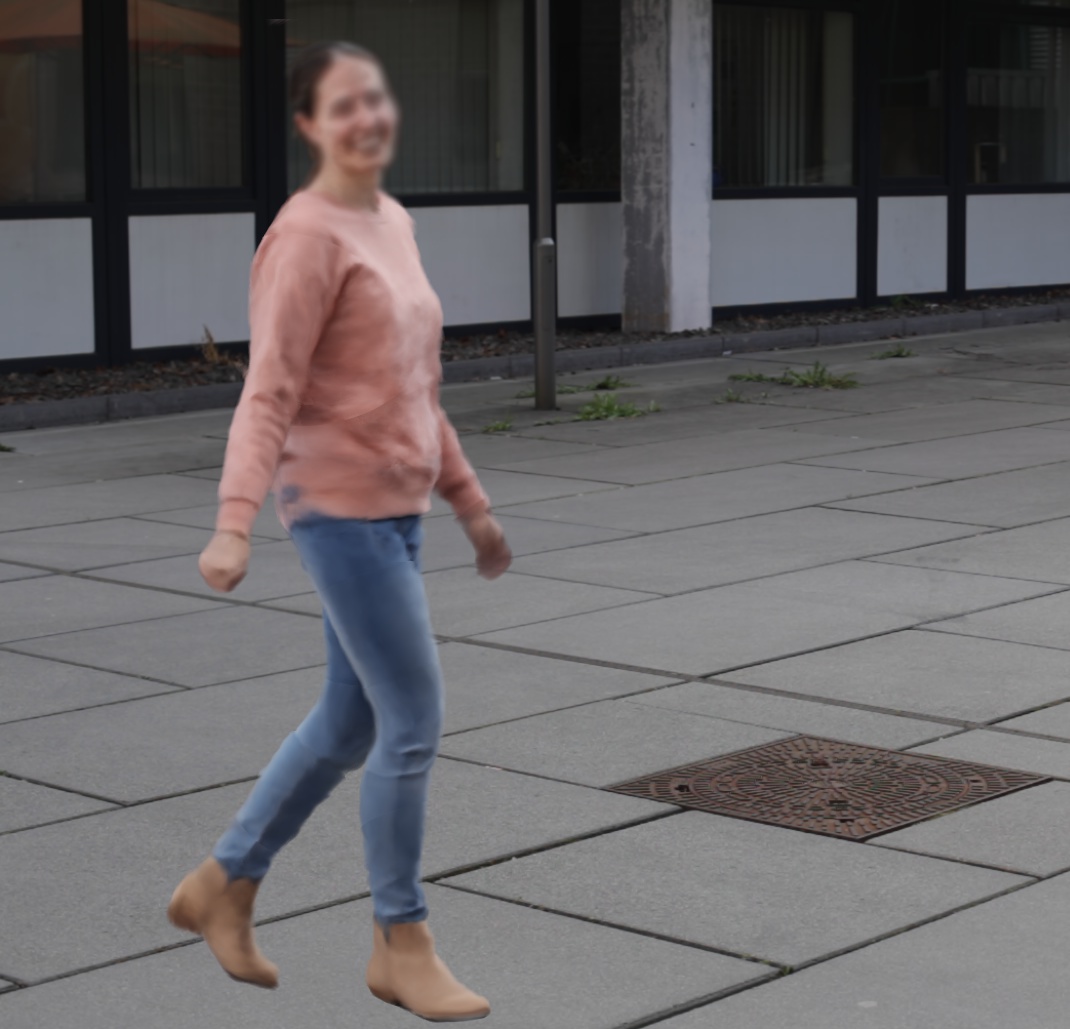}}\\%
    \vspace{\qualheldoutvspace}%
    \fbox{\includegraphics[height=\qualheldoutscale]{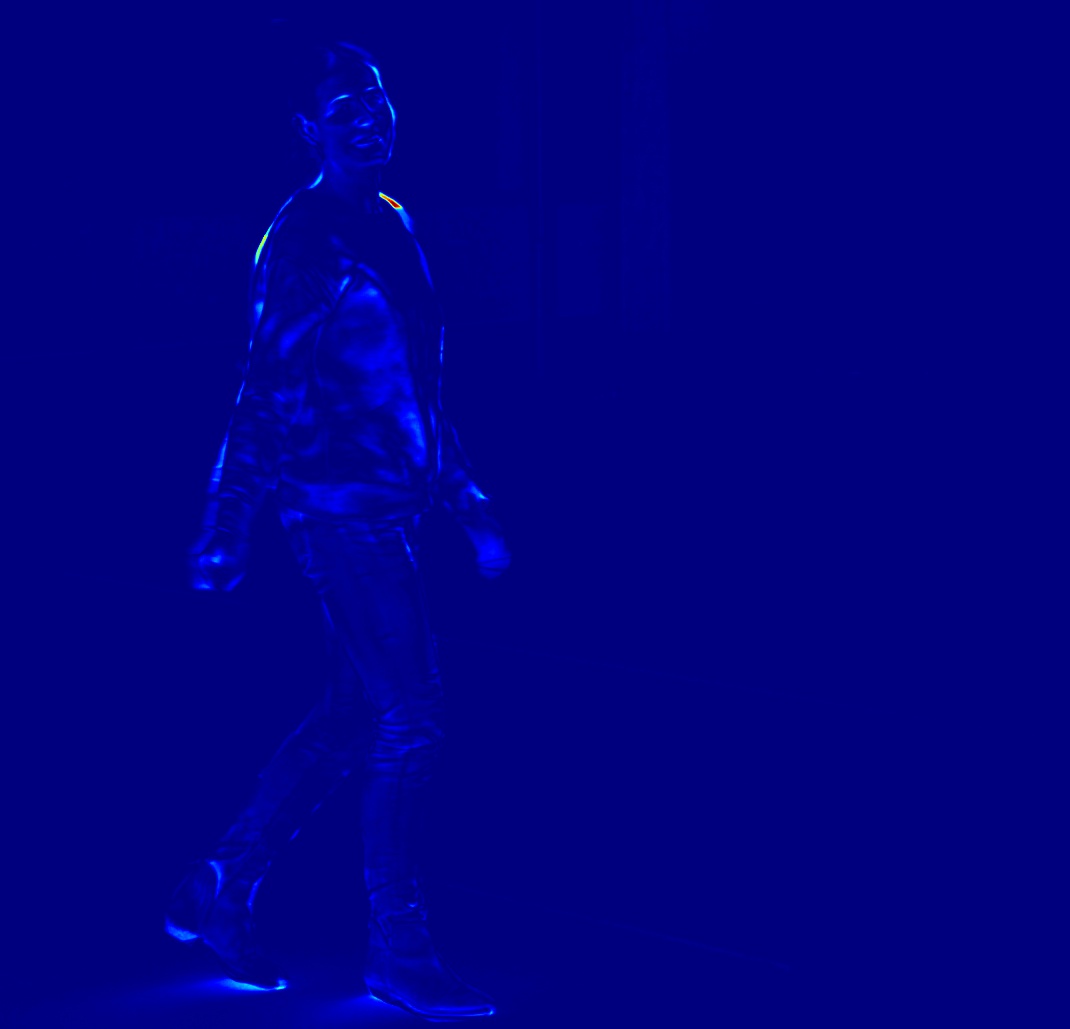}}\\%
    \vspace{\qualheldoutvspace}%
    \fbox{\includegraphics[height=\qualheldoutscale]{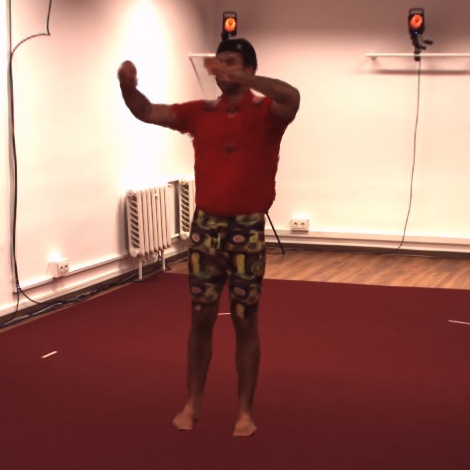}}\\%
    \vspace{\qualheldoutvspace}%
    \fbox{\includegraphics[height=\qualheldoutscale]{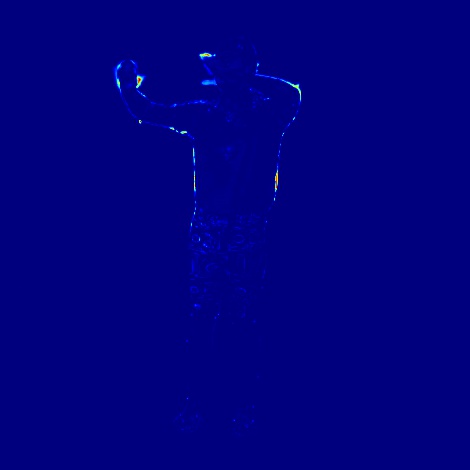}}\\%
    {\footnotesize A-NeRF}
}%
\hfill\hspace{1mm}%
\caption{\tb{A-NeRF with pose refinement generalizes better to test sequences.} We visualize both the rendered images, as well as the photometric error (squared distance, normalized to $\left[0,1\right]$) between the rendered and the ground truth images (warmer color indicates higher error) from our MonoPerfCap (1-2th rows) and Human 3.6M (3-4th rows) held-out sets. %
NeuralBody models produce artifacts around body contours.
A-NeRF w/o refinement cannot reproduce facial features and limbs. With pose refinement, A-NeRF can produce both appearances and shapes more plausibly.}
\label{fig:qual-heldout}
\end{figure}

\paragraph{Ablation study.} Our detailed ablation studies are reported in the supplemental document. In summary, they reveal:
i) Embedding relative 3D position, $\local{\query}_k$, instead of our proposed radial embedding $\querydist$ yields only half as good pose refinements.
ii) Our embedding choices keep the dimensionality moderate while improving on or matching the PSNR and SSIM of higher-dimensional variants.
iii) For a fixed number of images with accurate poses, learning from a long video with diverse poses has visual quality comparable to learning from multiple shorter clips. 

\paragraph{Limitations and Failure Cases.}

\newlength\failurelinescale
\setlength\failurelinescale{0.20\linewidth}
\newlength\failurescaler
\setlength\failurescaler{0.5\failurelinescale}
\begin{wrapfigure}[8]{r}{\failurelinescale}
\centering%
\setlength{\fboxrule}{0pt}%
\parbox[t]{0.5\failurelinescale}{%
\vspace{0mm}\centering%
\vspace{-0.6cm}
\fbox{\includegraphics[width=\failurescaler,trim=220 300 557 380,clip]{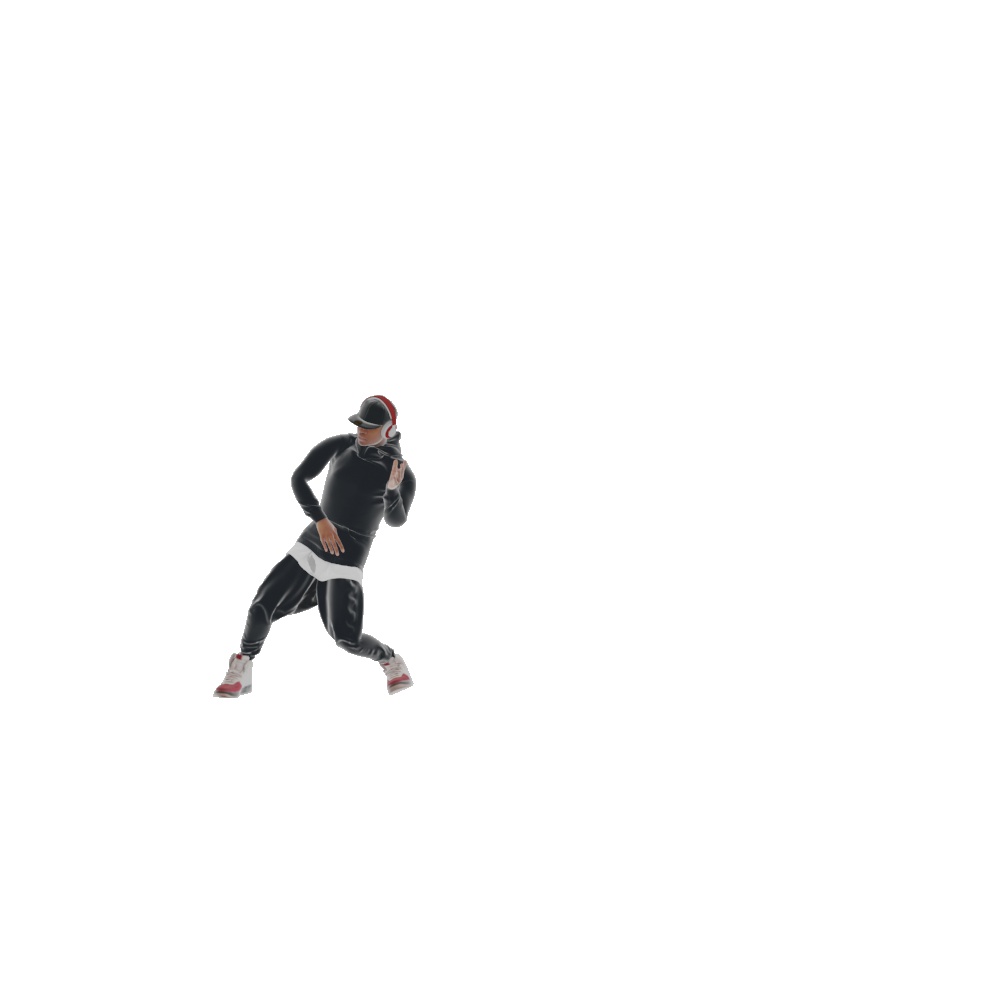}}\\%
{\small Reference}%
}%
\parbox[t]{0.5\failurelinescale}{%
\vspace{0mm}\centering%
\vspace{-0.6cm}
\fbox{\includegraphics[width=0.93\failurescaler,trim=230 320 630 420,clip]{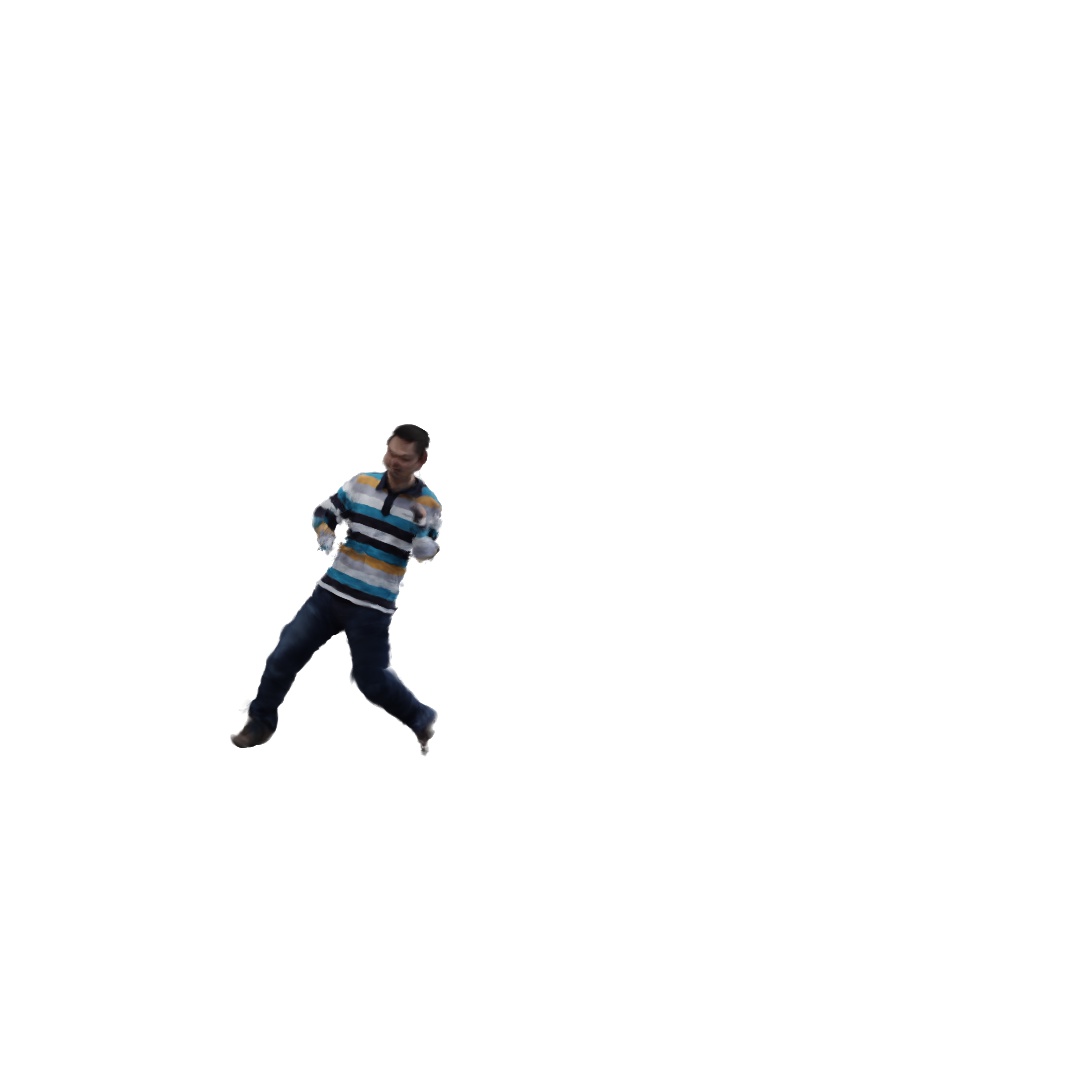}}\\
{\small Failure}%
}%
\label{fig:example-failure}
\end{wrapfigure}

Our computation time is the biggest bottleneck in extending A-NeRF to long sequences and multiple actors. 
Although a single static camera suffices, A-NeRF requires to see the person from all sides in varying poses to learn pose dependencies from data. The inlet shows a rendering of an extreme breakdance pose retargeted to a model trained on normal walking motions. Hence, the source pose is unseen during training and far from the data distribution, which leads to artifacts.

\section{Conclusion}
We propose a new way for integrating articulated skeleton models and implicit functions via an overcomplete re-parametrization. It includes learning an interpretable 3D representation from 2D images; a personalized volumetric density field with texture detail and time-varying poses of the actor depicted in the input.
The underlying ill-posed problem of mapping a single query location to multiple parts is addressed with an overparametrization over nearby parts. To the best of our knowledge, A-NeRF is the first approach to define NeRF models for extreme and articulated motion on unconstrained video and this new approach scores high on the Human 3.6M benchmark.
Importantly, it works from a single video and naturally extends to multi-view and does not require camera calibration in either scenario.
This is an important step towards making motion capture more accurate and practical.
In future work, we will learn a general human model from a database of subjects instead of individuals. 

\paragraph{Funding in direct support of this work:} NSERC Discovery grant, UBC Advanced Research Computing (ARC) GPU cluster, Compute Canada GPU servers, and a gift by Facebook Reality Labs.

\bibliographystyle{abbrvnat}    
\bibliography{bib/string,bib/bibliography,bib/XNect_article,bib/XNect_eccv,bib/vision,bib/learning,bib/biomed}
\begin{appendices}
\clearpage
\setcounter{table}{0}
\renewcommand{\thetable}{A\arabic{table}}
\setcounter{figure}{0}
\renewcommand{\thefigure}{A\arabic{figure}}
\setcounter{section}{0}
\renewcommand{\thesection}{\Alph{section}}
\section*{Appendix}
In this document, we present visualizations of the learned A-NeRF body geometry~(\secref{sec:visual-geo}), and show additional qualitative results on novel view synthesis~(\secref{sec:add-qual-anerf}) and pose refinement~(\secref{sec:add-qual-pose}). 
We show comprehensive ablation studies on A-NeRF (\secref{sec:ablation-std}), and then make quantitative comparisons between A-NeRF and the most related work~\cite{peng2020arxiv_neuralbody} on a dataset with perfect training poses (\secref{sec:surreal-visual-comp}). 
We justify our choice of view encoding in the context of modeling the view-dependent effects of the human body (\secref{sec:view-dependent}).
Finally, we describe additional dataset information (\secref{sec:dataset-supp}) and implementation details (\secref{sec:imp-det}). The supplemental video shows results in motion. \tbi{Real faces and their reconstructions are blurred for anonymity}.

\newlength\qualgeosuppscale
\setlength\qualgeosuppscale{0.135\textwidth}
\begin{figure}
\setlength{\fboxrule}{0pt}%
\centering
\parbox[t]{\qualgeosuppscale}{%
    \vspace{0mm}\centering%
    \fbox{\includegraphics[width=\qualgeosuppscale]{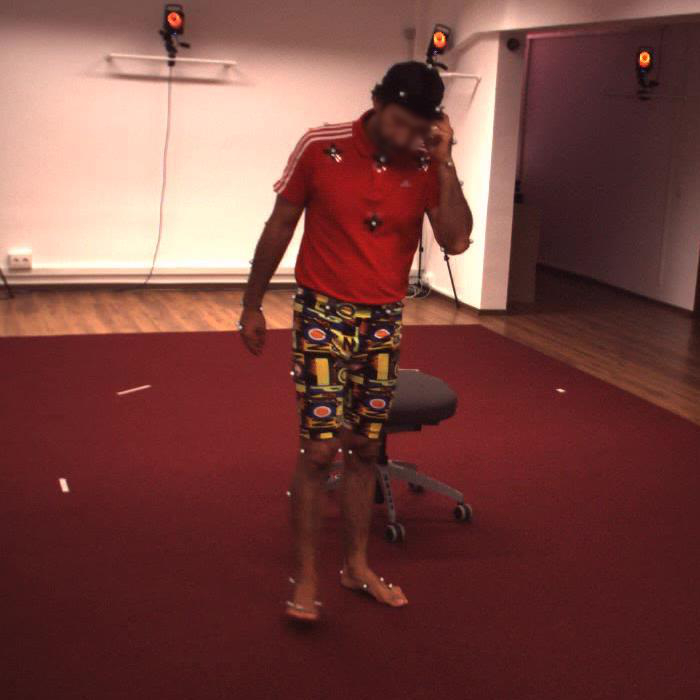}}\\%
    \fbox{\includegraphics[width=\qualgeosuppscale]{images/supp/geometry/s11/ref_036_blur}}\\%
    \fbox{\includegraphics[width=\qualgeosuppscale]{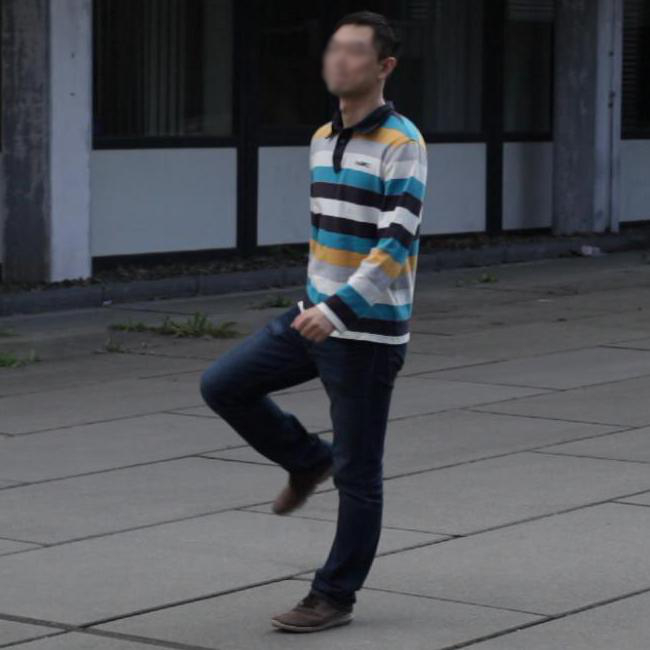}}\\%
    \fbox{\includegraphics[width=\qualgeosuppscale]{images/supp/geometry/nd/ref_018_blur}}\\%
    \fbox{\includegraphics[width=\qualgeosuppscale]{images/supp/geometry/james/ref_012}}\\%
    \fbox{\includegraphics[width=\qualgeosuppscale]{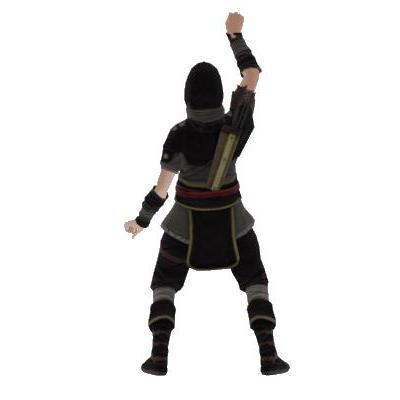}}\\%
    {\footnotesize Ref. Image}\\
}%
\hfill%
\parbox[t]{\qualgeosuppscale}{%
    \vspace{0mm}\centering%
    \fbox{\includegraphics[width=\qualgeosuppscale]{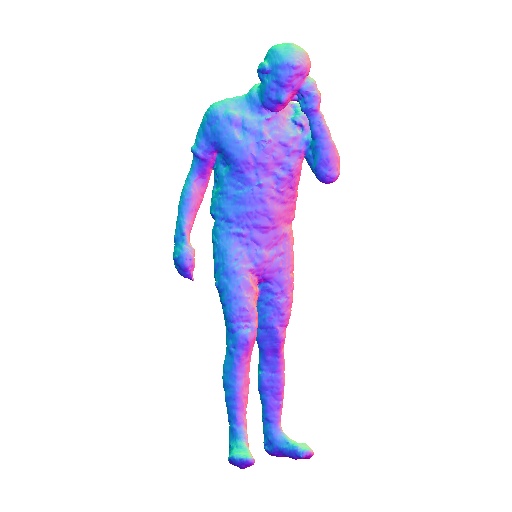}}\\%
    \fbox{\includegraphics[width=\qualgeosuppscale]{images/supp/geometry/s11/000}}\\%
    \fbox{\includegraphics[width=\qualgeosuppscale]{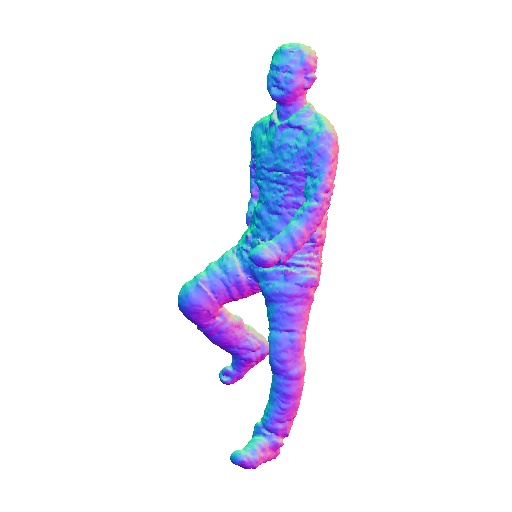}}\\%
    \fbox{\includegraphics[width=\qualgeosuppscale]{images/supp/geometry/nd/000}}\\%
    \fbox{\includegraphics[width=\qualgeosuppscale]{images/supp/geometry/james/000}}\\%
    \fbox{\includegraphics[width=\qualgeosuppscale]{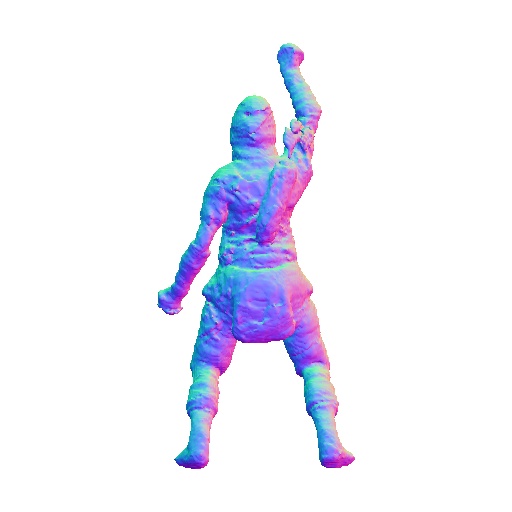}}\\%
    {{\footnotesize Geometry $\rightarrow$}}
}%
\hfill%
\parbox[t]{\qualgeosuppscale}{%
    \vspace{0mm}\centering%
    \fbox{\includegraphics[width=\qualgeosuppscale]{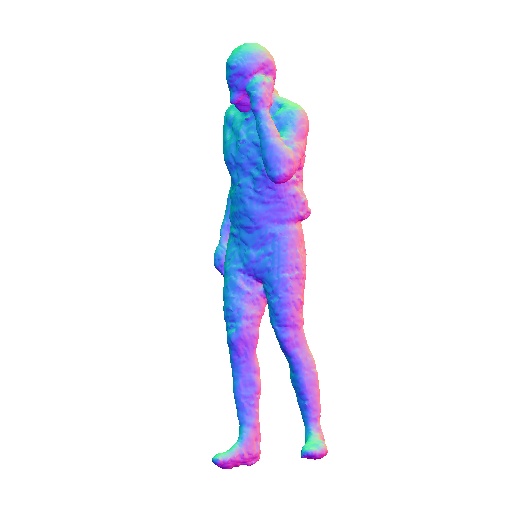}}\\%
    \fbox{\includegraphics[width=\qualgeosuppscale]{images/supp/geometry/s11/001}}\\%
    \fbox{\includegraphics[width=\qualgeosuppscale]{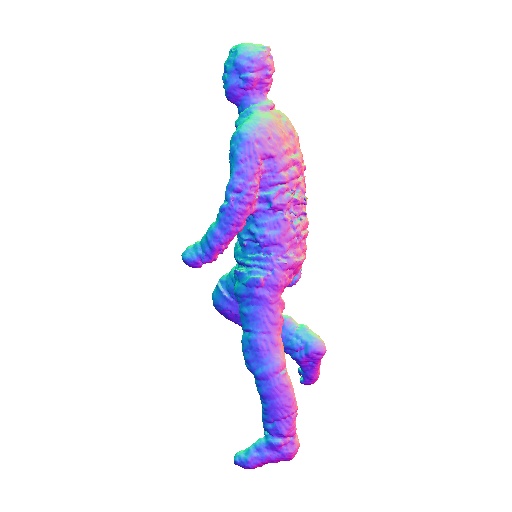}}\\%
    \fbox{\includegraphics[width=\qualgeosuppscale]{images/supp/geometry/nd/001}}\\%
    \fbox{\includegraphics[width=\qualgeosuppscale]{images/supp/geometry/james/001}}\\%
    \fbox{\includegraphics[width=\qualgeosuppscale]{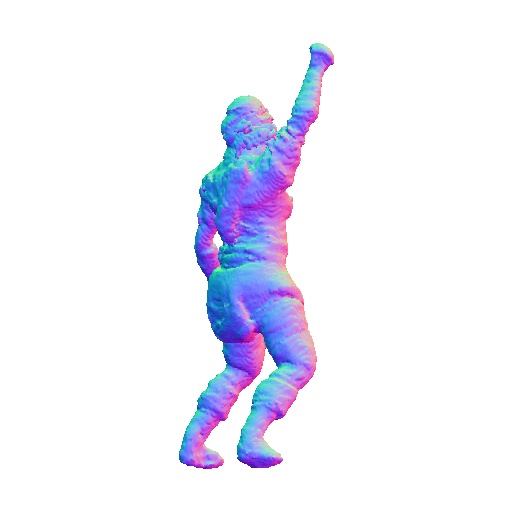}}\\%
}%
\hfill%
\parbox[t]{\qualgeosuppscale}{%
    \vspace{0mm}\centering%
    \fbox{\includegraphics[width=\qualgeosuppscale]{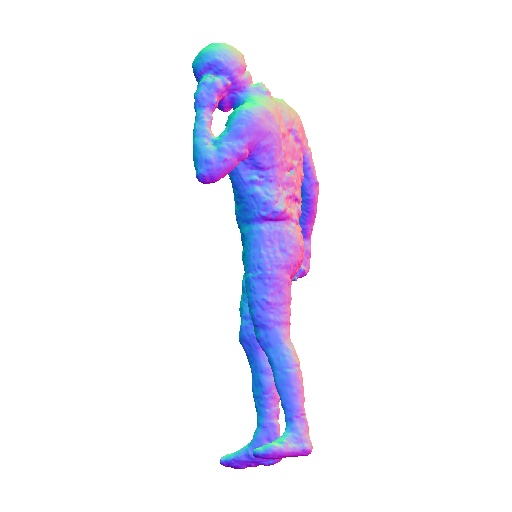}}\\%
    \fbox{\includegraphics[width=\qualgeosuppscale]{images/supp/geometry/s11/002}}\\%
    \fbox{\includegraphics[width=\qualgeosuppscale]{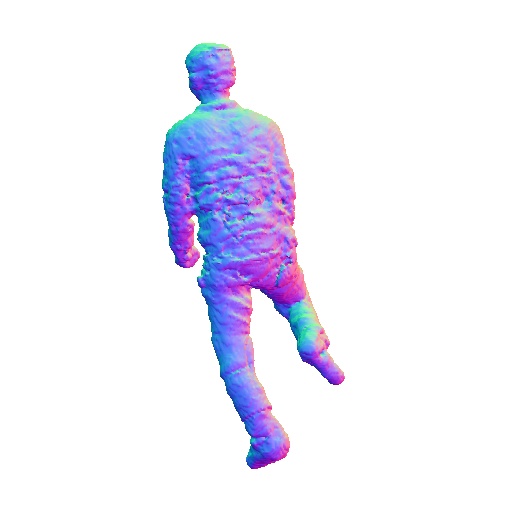}}\\%
    \fbox{\includegraphics[width=\qualgeosuppscale]{images/supp/geometry/nd/002}}\\%
    \fbox{\includegraphics[width=\qualgeosuppscale]{images/supp/geometry/james/002}}\\%
    \fbox{\includegraphics[width=\qualgeosuppscale]{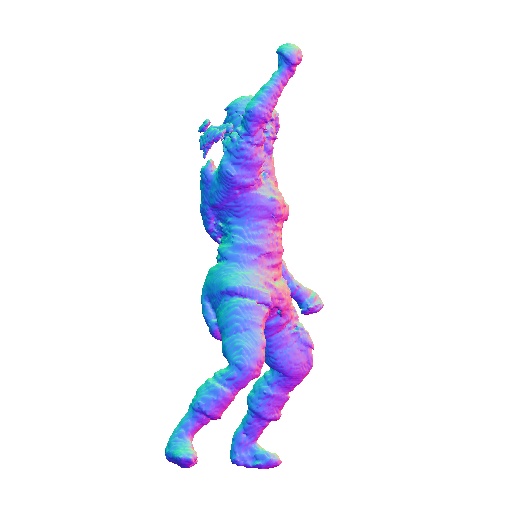}}\\%
}%
\hfill%
\parbox[t]{\qualgeosuppscale}{%
    \vspace{0mm}\centering%
    \fbox{\includegraphics[width=\qualgeosuppscale]{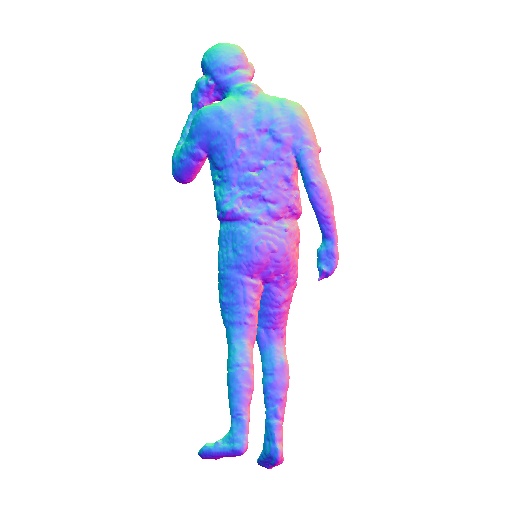}}\\%
    \fbox{\includegraphics[width=\qualgeosuppscale]{images/supp/geometry/s11/003}}\\%
    \fbox{\includegraphics[width=\qualgeosuppscale]{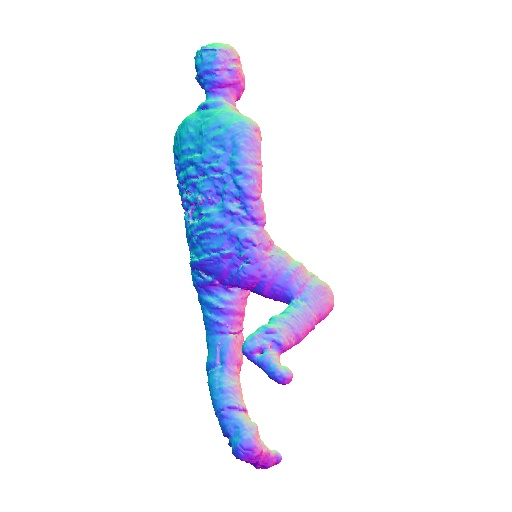}}\\%
    \fbox{\includegraphics[width=\qualgeosuppscale]{images/supp/geometry/nd/003}}\\%
    \fbox{\includegraphics[width=\qualgeosuppscale]{images/supp/geometry/james/003}}\\%
    \fbox{\includegraphics[width=\qualgeosuppscale]{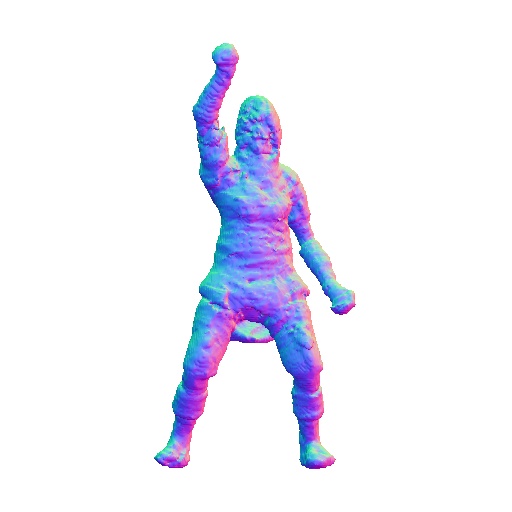}}\\%
}%
\hfill%
\parbox[t]{\qualgeosuppscale}{%
    \vspace{0mm}\centering%
    \fbox{\includegraphics[width=\qualgeosuppscale]{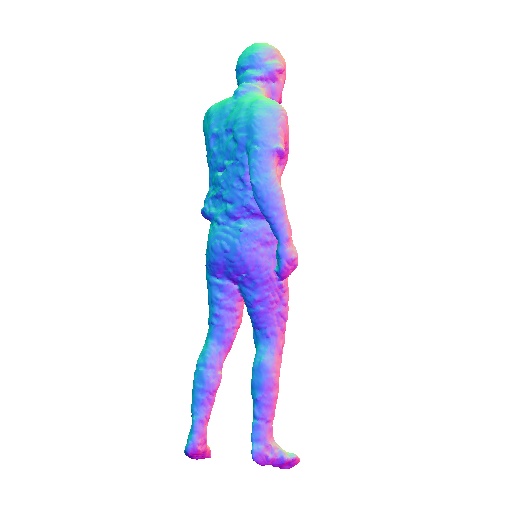}}\\%
    \fbox{\includegraphics[width=\qualgeosuppscale]{images/supp/geometry/s11/004}}\\%
    \fbox{\includegraphics[width=\qualgeosuppscale]{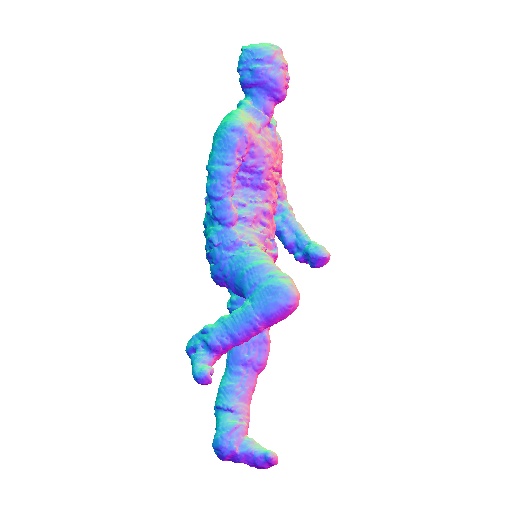}}\\%
    \fbox{\includegraphics[width=\qualgeosuppscale]{images/supp/geometry/nd/004}}\\%
    \fbox{\includegraphics[width=\qualgeosuppscale]{images/supp/geometry/james/004}}\\%
    \fbox{\includegraphics[width=\qualgeosuppscale]{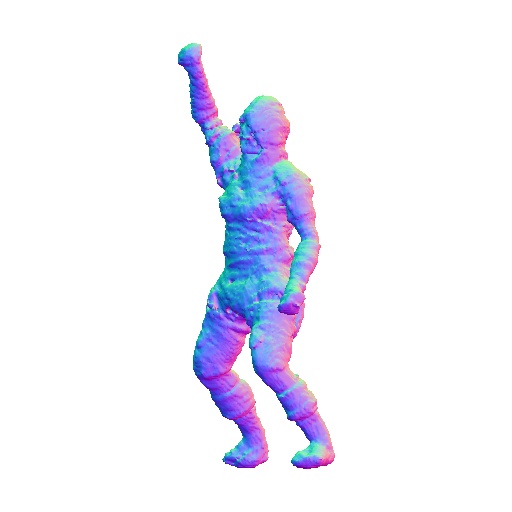}}\\%
}%
\hfill%
\parbox[t]{\qualgeosuppscale}{%
    \vspace{0mm}\centering%
    \fbox{\includegraphics[width=\qualgeosuppscale]{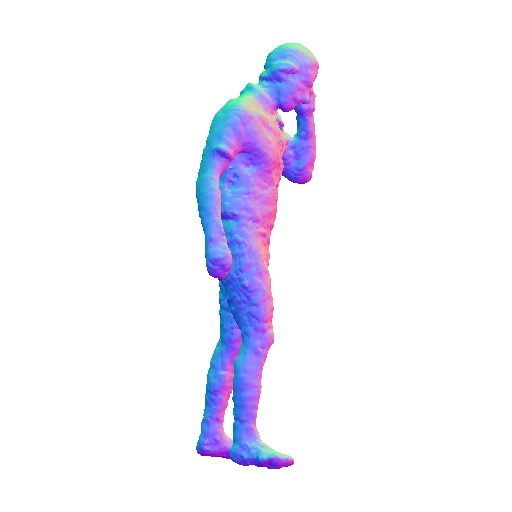}}\\%
    \fbox{\includegraphics[width=\qualgeosuppscale]{images/supp/geometry/s11/005}}\\%
    \fbox{\includegraphics[width=\qualgeosuppscale]{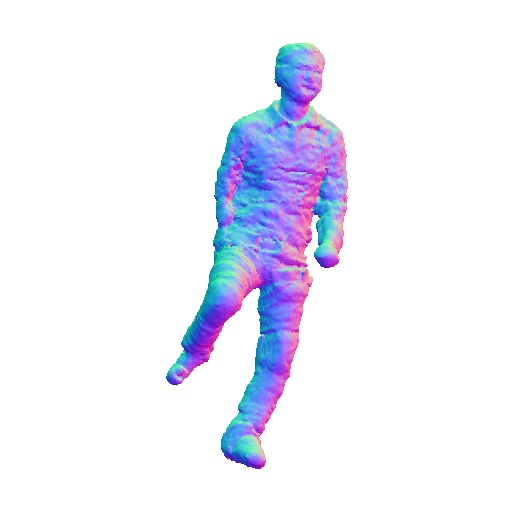}}\\%
    \fbox{\includegraphics[width=\qualgeosuppscale]{images/supp/geometry/nd/005}}\\%
    \fbox{\includegraphics[width=\qualgeosuppscale]{images/supp/geometry/james/005}}\\%
    \fbox{\includegraphics[width=\qualgeosuppscale]{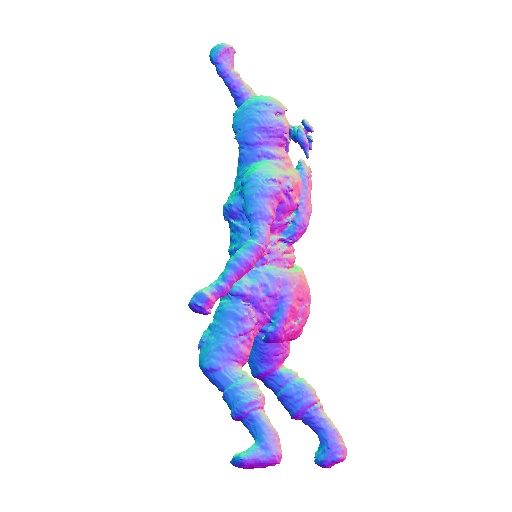}}\\%
}%
\caption{\tb{A-NeRF learns plausible geometry without relying on explicit surface templates and accurate initial poses.} We visualize the geometry using Marching Cubes~\cite{lorensen1987marching}, with the resolution of the voxel grid set to 256.}
\label{fig:geometry}
\end{figure}
\section{Geometry Visualization}
\label{sec:visual-geo}
In ~\figref{fig:geometry}, we show that A-NeRF can learn convincing body geometry without relying on pre-defined template meshes. The surface-free property also allows A-NeRF to model accessories like headphones, caps and quivers (the second last and the last rows in~\figref{fig:qual-nvs-extra}), which are often not included in human template meshes. 
Moreover, the geometry is consistent between front and back even though learned only from a monocular video, so long as the person is seen from the back. However, very fine details such as the thin arrow shaft are not captured and the surface is not regularized to be smooth.
\newlength\qualnvsextrascale
\setlength\qualnvsextrascale{0.16\linewidth}

\begin{figure}[h]
\setlength{\fboxrule}{0pt}%
\centering
\parbox[t]{\qualnvsextrascale}{%
    \vspace{0mm}\centering%
    \fbox{\includegraphics[width=\qualnvsextrascale]{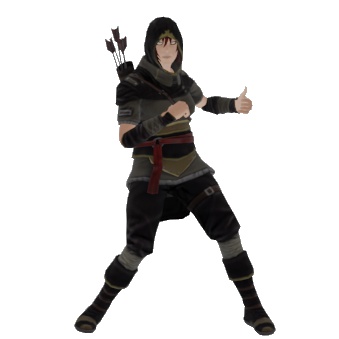}}\\%
    \fbox{\includegraphics[width=\qualnvsextrascale]{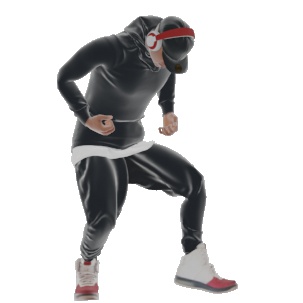}}\\%
    \fbox{\includegraphics[width=\qualnvsextrascale]{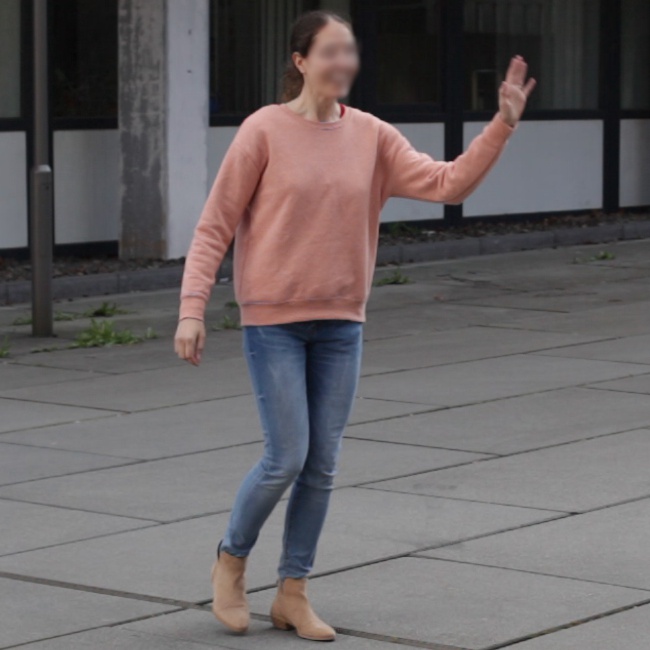}}\\%
    \fbox{\includegraphics[width=\qualnvsextrascale]{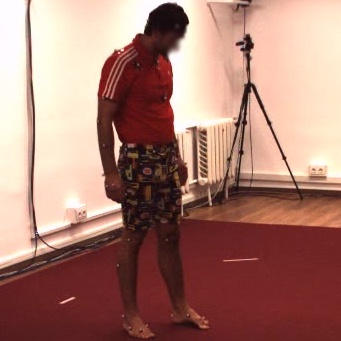}}\\%
    \fbox{\includegraphics[width=\qualnvsextrascale]{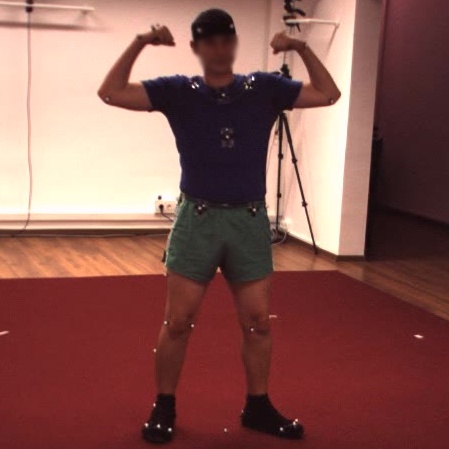}}\\%
    \fbox{\includegraphics[width=\qualnvsextrascale]{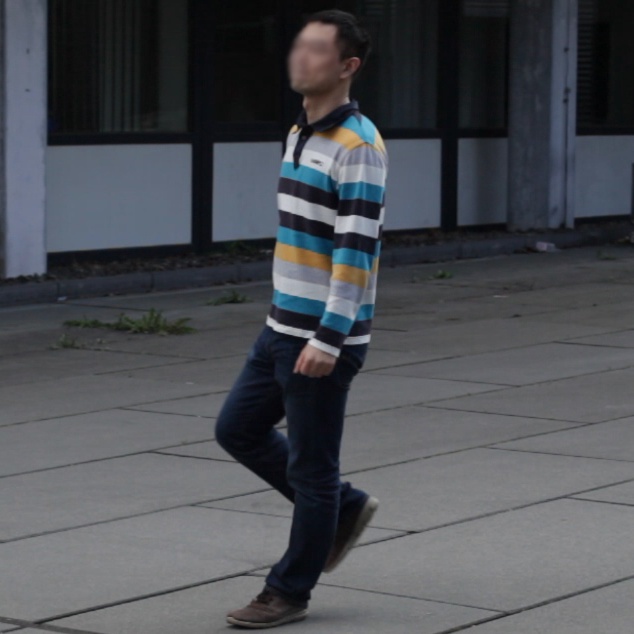}}\\%
    {\footnotesize Reference}
}%
\hfill%
\parbox[t]{\qualnvsextrascale}{%
    \vspace{0mm}\centering%
    \fbox{\includegraphics[width=\qualnvsextrascale]{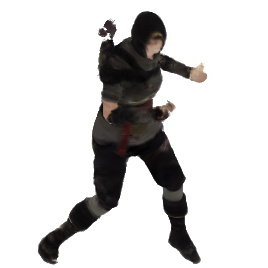}}\\%
    \fbox{\includegraphics[width=\qualnvsextrascale]{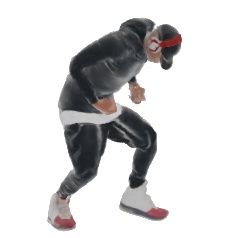}}\\%
    \fbox{\includegraphics[width=\qualnvsextrascale]{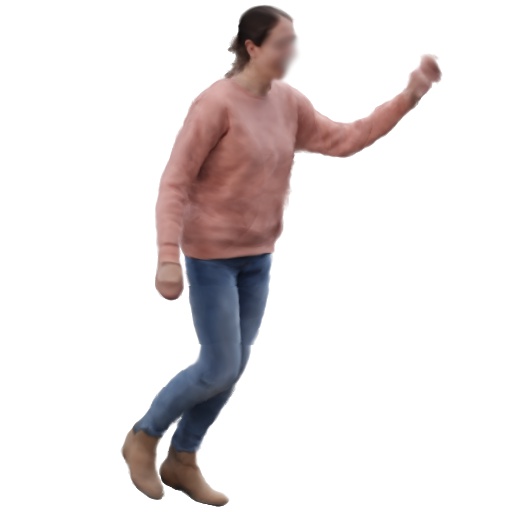}}\\%
    \fbox{\includegraphics[width=\qualnvsextrascale]{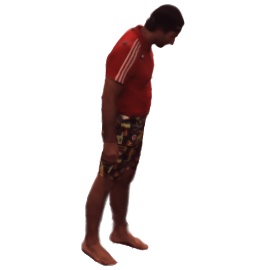}}\\%
    \fbox{\includegraphics[width=\qualnvsextrascale]{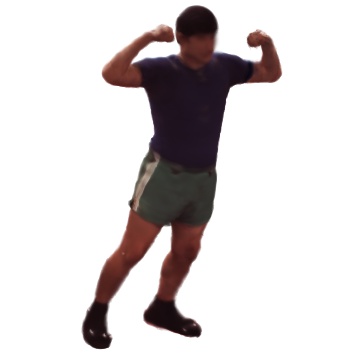}}\\%
    \fbox{\includegraphics[width=\qualnvsextrascale]{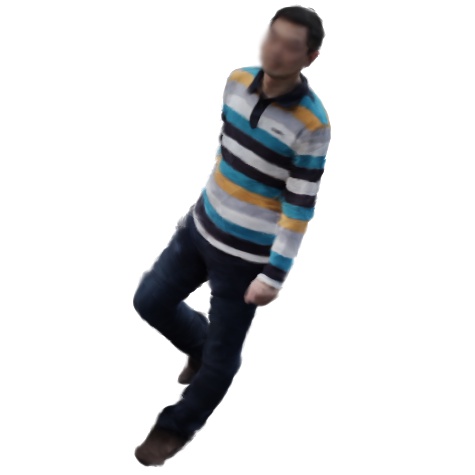}}\\%
    {\footnotesize Novel Views ($\rightarrow$)}
}%
\hfill%
\parbox[t]{\qualnvsextrascale}{%
    \vspace{0mm}\centering%
    \fbox{\includegraphics[width=\qualnvsextrascale]{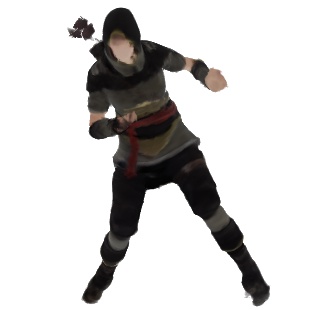}}\\%
    \fbox{\includegraphics[width=\qualnvsextrascale]{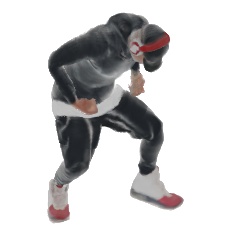}}\\%
    \fbox{\includegraphics[width=\qualnvsextrascale]{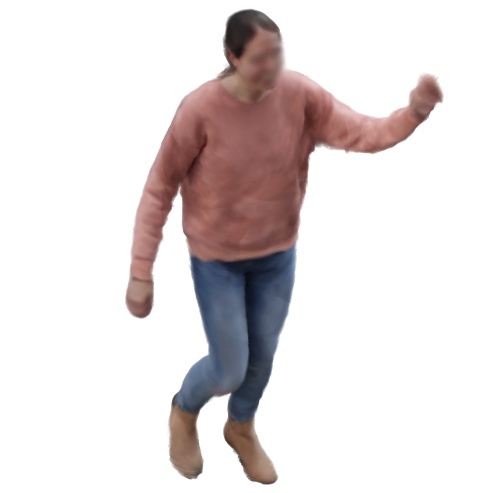}}\\%
    \fbox{\includegraphics[width=\qualnvsextrascale]{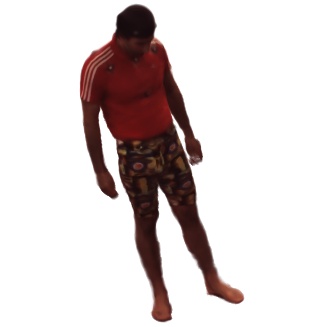}}\\%
    \fbox{\includegraphics[width=\qualnvsextrascale]{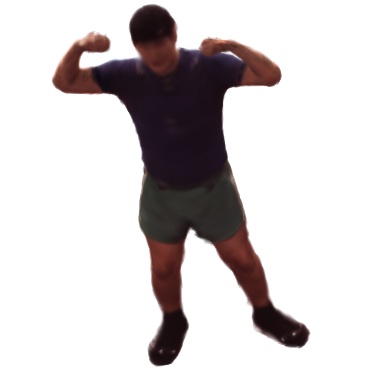}}\\%
    \fbox{\includegraphics[width=\qualnvsextrascale]{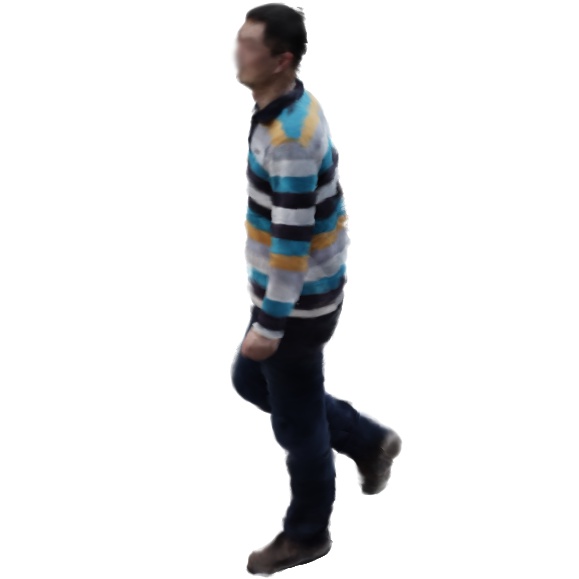}}\\%
}%
\hfill%
\parbox[t]{\qualnvsextrascale}{%
    \vspace{0mm}\centering%
    \fbox{\includegraphics[width=\qualnvsextrascale]{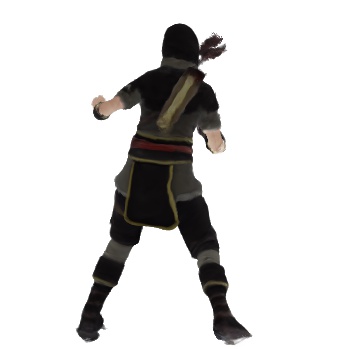}}\\%
    \fbox{\includegraphics[width=\qualnvsextrascale ]{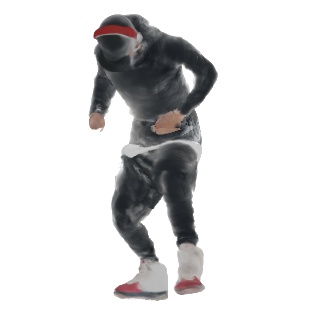}}\\%
    \fbox{\includegraphics[width=\qualnvsextrascale ]{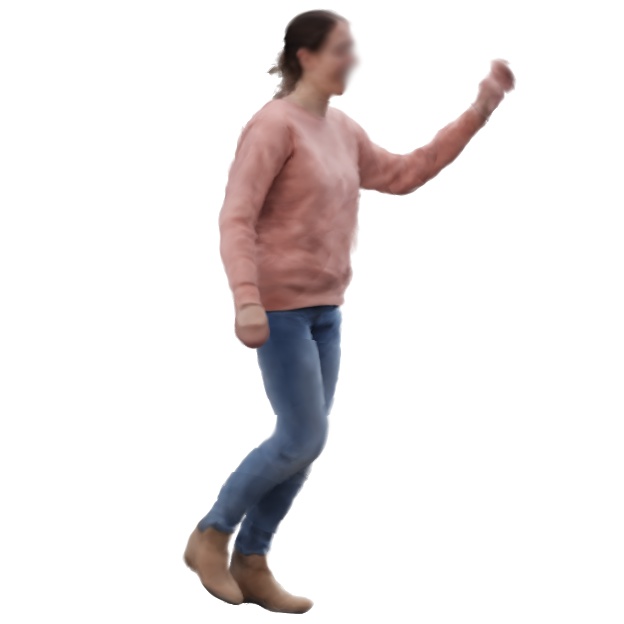}}\\%
    \fbox{\includegraphics[width=\qualnvsextrascale]{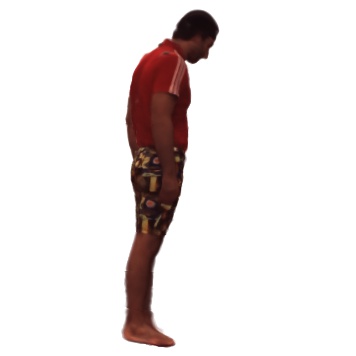}}\\%
    \fbox{\includegraphics[width=\qualnvsextrascale]{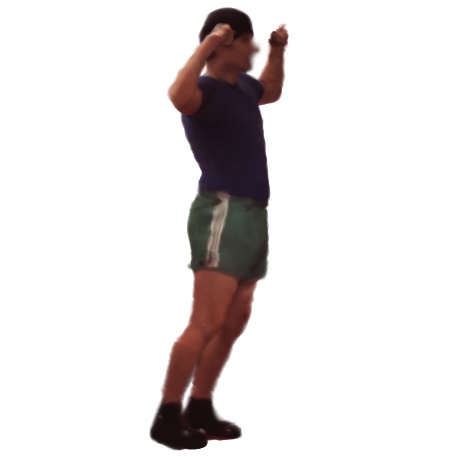}}\\%
    \fbox{\includegraphics[width=\qualnvsextrascale]{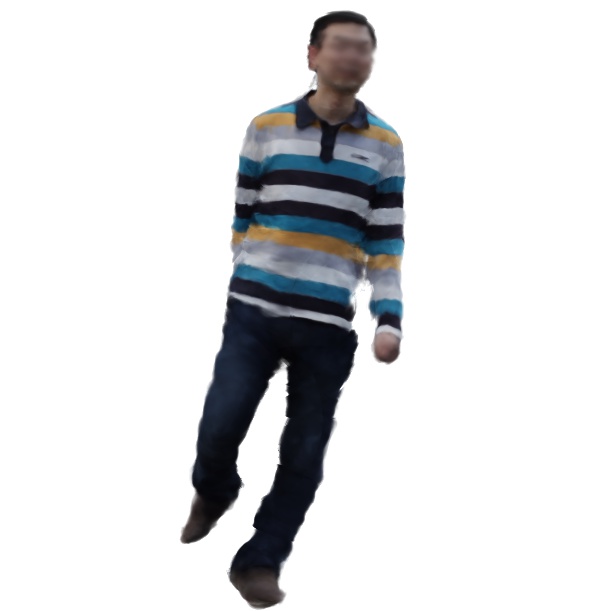}}\\%
}%
\hfill%
\parbox[t]{\qualnvsextrascale}{%
    \vspace{0mm}\centering%
    \fbox{\includegraphics[width=\qualnvsextrascale]{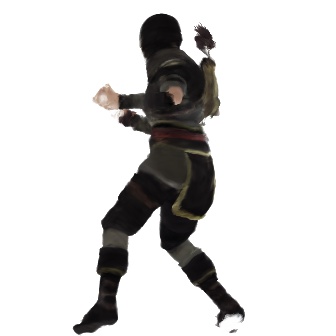}}\\%
    \fbox{\includegraphics[width=\qualnvsextrascale ]{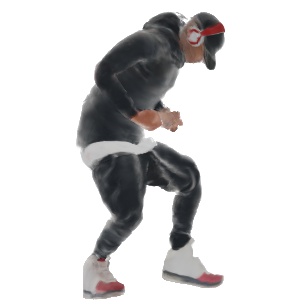}}\\%
    \fbox{\includegraphics[width=\qualnvsextrascale ]{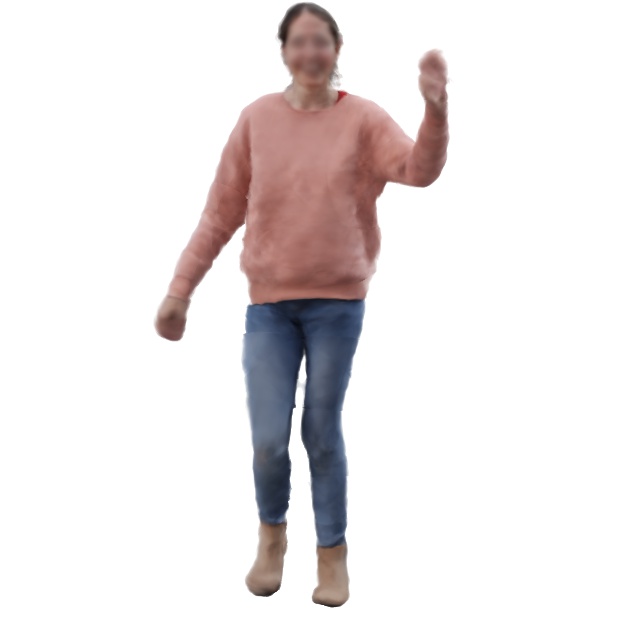}}\\%
    \fbox{\includegraphics[width=\qualnvsextrascale]{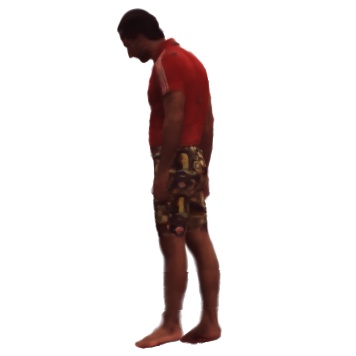}}\\%
    \fbox{\includegraphics[width=\qualnvsextrascale]{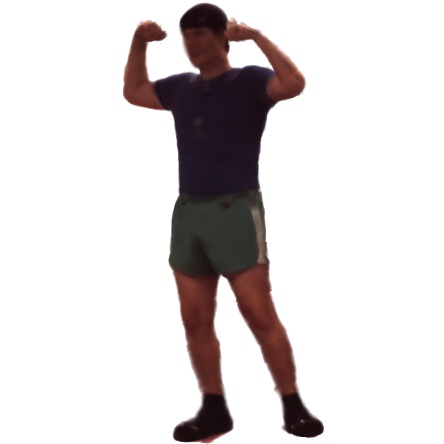}}\\%
    \fbox{\includegraphics[width=\qualnvsextrascale]{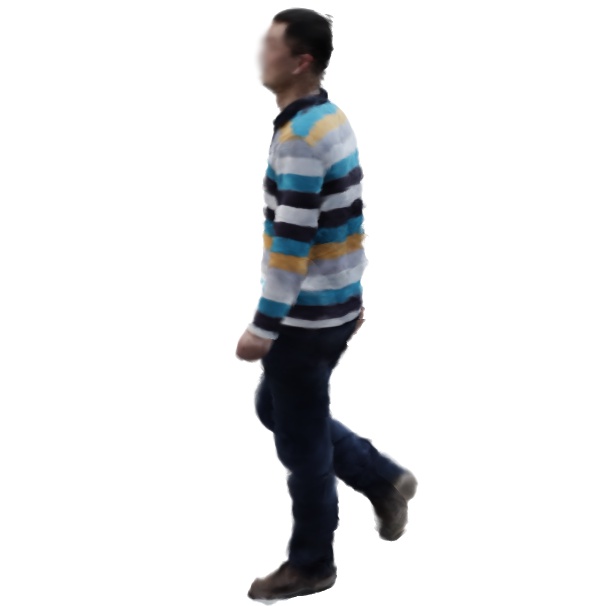}}\\%
}%
\caption{\tb{Additional novel view results.} The learned volumetric body model can be viewed from different directions with consistent geometry. Furthermore, A-NeRF also captures details such as arrows, quivers, and caps (1st and 2nd row), which are usually not included in human surface models; but such thin structures remain challenging. \tbi{Real faces and their reconstructions are blurred in all figures for anonymity.}
}
\label{fig:qual-nvs-extra}
\end{figure}

\section{Additional Qualitative Results for A-NeRF}
\label{sec:add-qual-anerf}
We include additional novel view synthesis results in~\figref{fig:qual-nvs-extra}. A-NeRF shows plausible synthesis results from different angles. Besides the geometric detail of the density field analyzed in~\secref{sec:visual-geo}, A-NeRF also reconstructs the appearance of small structures that are not modeled by human surface models and are most difficult to reconstruct, such as arrows, quivers, and caps (first two rows in~\figref{fig:qual-nvs-extra}).

\section{Additional Pose Refinement Results for A-NeRF}
\label{sec:add-qual-pose}
\newlength\qualposerefine
\setlength\qualposerefine{0.164\textwidth}
\begin{figure}
\setlength{\fboxrule}{0pt}%
\centering
\parbox[t]{\qualposerefine}{%
    \vspace{0mm}\centering%
    \fbox{\includegraphics[width=\qualposerefine]{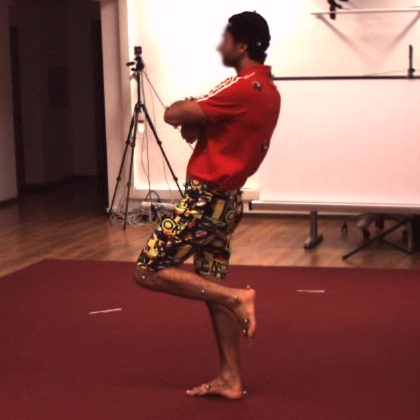}}\\%
    \fbox{\includegraphics[width=\qualposerefine]{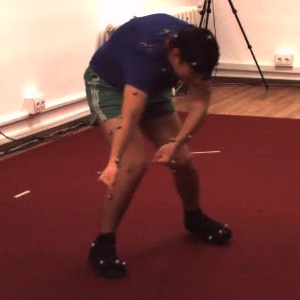}}\\%
    \fbox{\includegraphics[width=\qualposerefine]{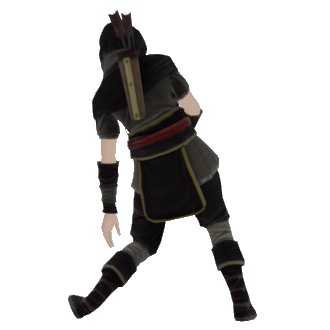}}\\%
    \fbox{\includegraphics[width=\qualposerefine]{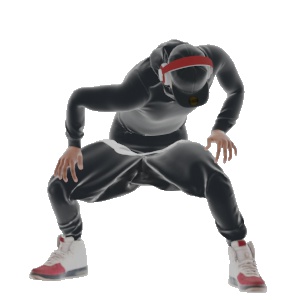}}\\%
    \fbox{\includegraphics[width=\qualposerefine]{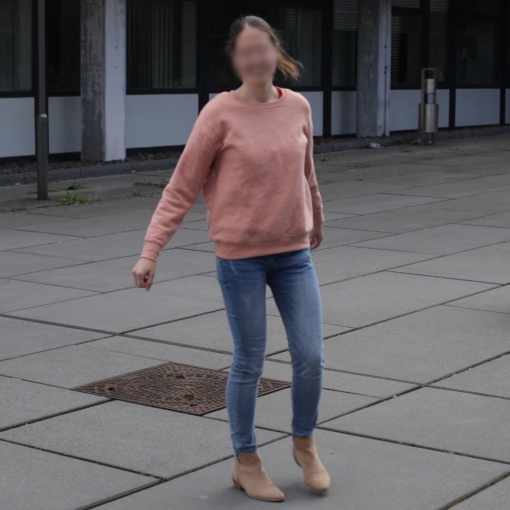}}\\%
    \fbox{\includegraphics[width=\qualposerefine]{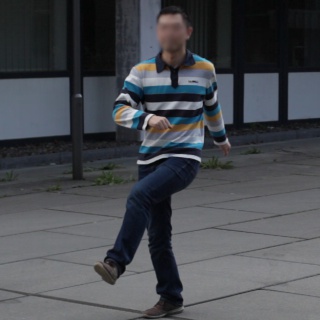}}\\%
    \fbox{\includegraphics[width=\qualposerefine]{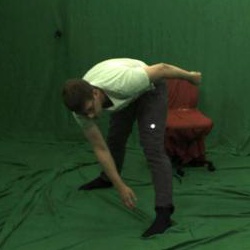}}\\%
    \fbox{\includegraphics[width=\qualposerefine]{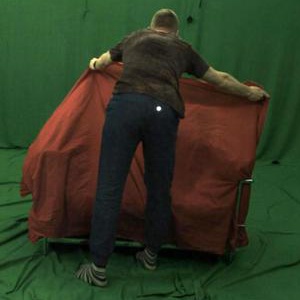}}\\%
    {\footnotesize Ref. Image}\\
}%
\hfill%
\parbox[t]{\qualposerefine}{%
    \vspace{0mm}\centering%
    \fbox{\includegraphics[width=\qualposerefine]{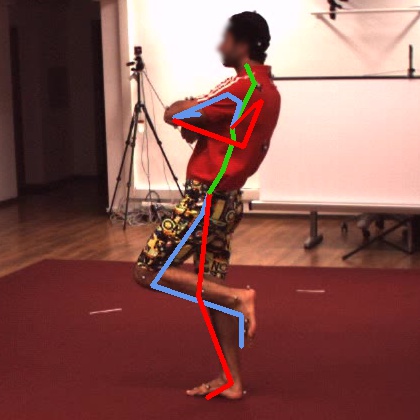}}\\%
    \fbox{\includegraphics[width=\qualposerefine]{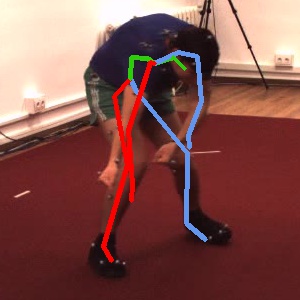}}\\%
    \fbox{\includegraphics[width=\qualposerefine]{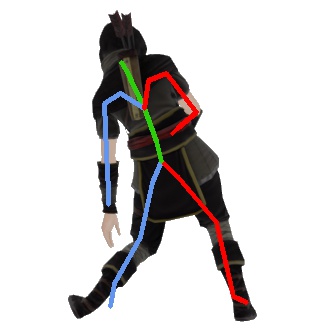}}\\%
    \fbox{\includegraphics[width=\qualposerefine]{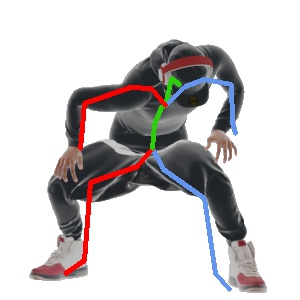}}\\%
    \fbox{\includegraphics[width=\qualposerefine]{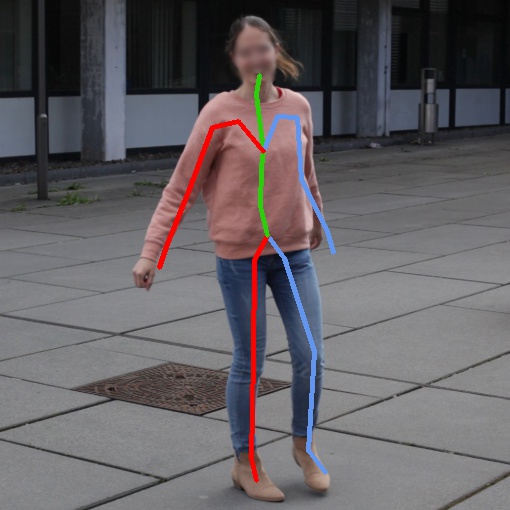}}\\%
    \fbox{\includegraphics[width=\qualposerefine]{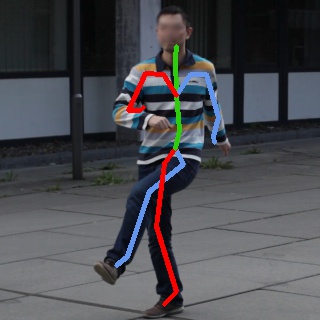}}\\%
    \fbox{\includegraphics[width=\qualposerefine]{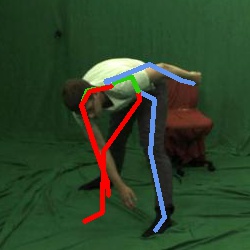}}\\%
    \fbox{\includegraphics[width=\qualposerefine]{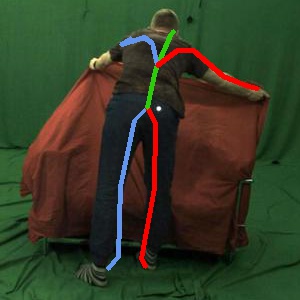}}\\%
    {\footnotesize SPIN~\cite{kolotouros2019learning_spin}}
}%
\hfill%
\parbox[t]{\qualposerefine}{%
    \vspace{0mm}\centering%
    \fbox{\includegraphics[width=\qualposerefine]{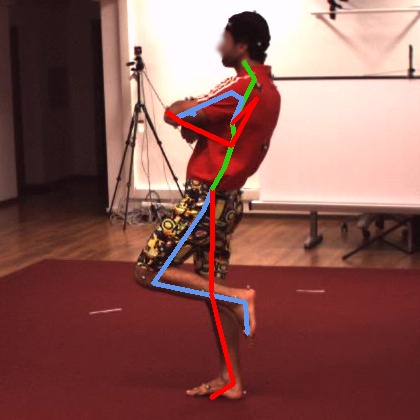}}\\%
    \fbox{\includegraphics[width=\qualposerefine]{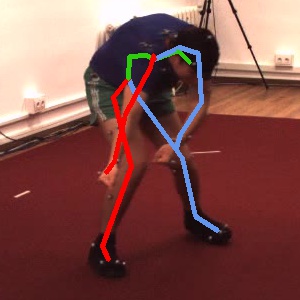}}\\%
    \fbox{\includegraphics[width=\qualposerefine]{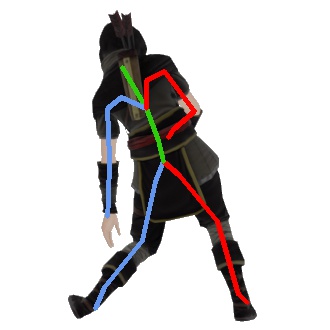}}\\%
    \fbox{\includegraphics[width=\qualposerefine]{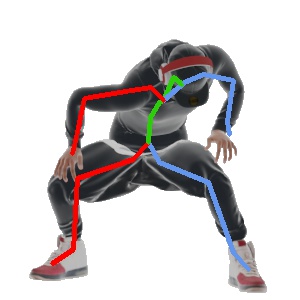}}\\%
    \fbox{\includegraphics[width=\qualposerefine]{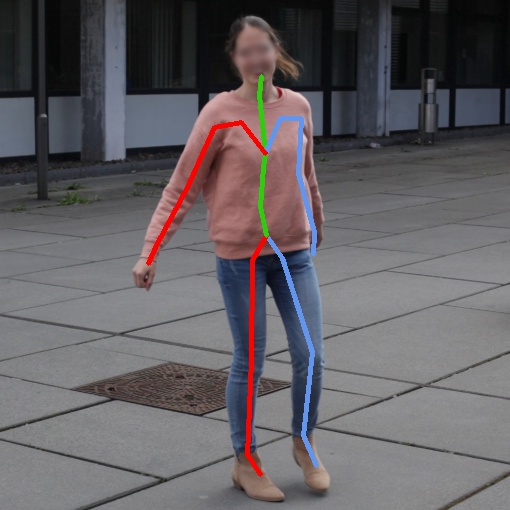}}\\%
    \fbox{\includegraphics[width=\qualposerefine]{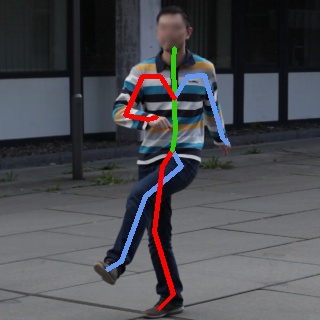}}\\%
    \fbox{\includegraphics[width=\qualposerefine]{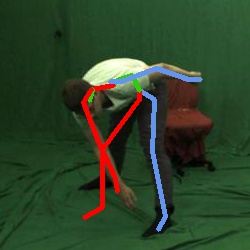}}\\%
    \fbox{\includegraphics[width=\qualposerefine]{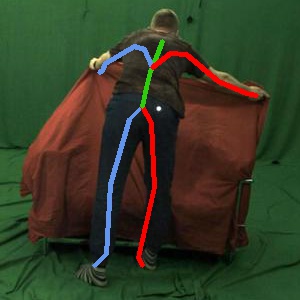}}\\%
    {\footnotesize Refined (Ours)}
}%
\hfill%
\parbox[t]{\qualposerefine}{%
    \vspace{0mm}\centering%
    \fbox{\includegraphics[width=\qualposerefine]{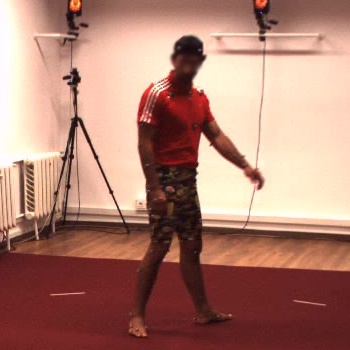}}\\%
    \fbox{\includegraphics[width=\qualposerefine]{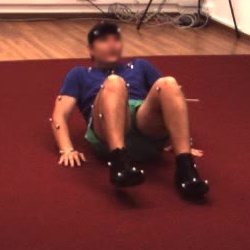}}\\%
    \fbox{\includegraphics[width=\qualposerefine]{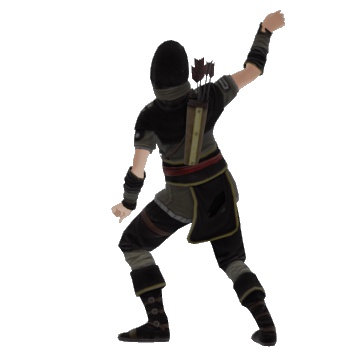}}\\%
    \fbox{\includegraphics[width=\qualposerefine]{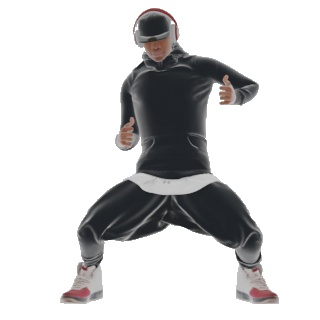}}\\%
    \fbox{\includegraphics[width=\qualposerefine]{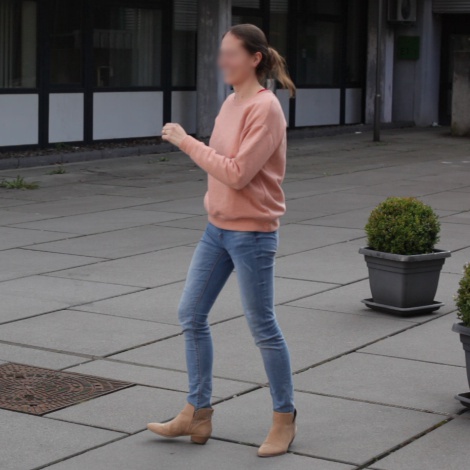}}\\%
    \fbox{\includegraphics[width=\qualposerefine]{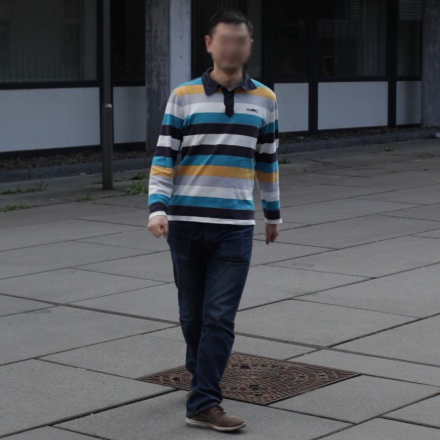}}\\%
    \fbox{\includegraphics[width=\qualposerefine]{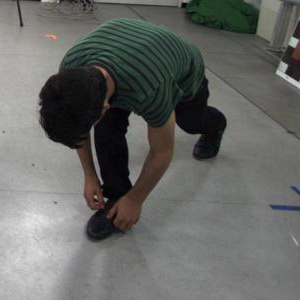}}\\%
    \fbox{\includegraphics[width=\qualposerefine]{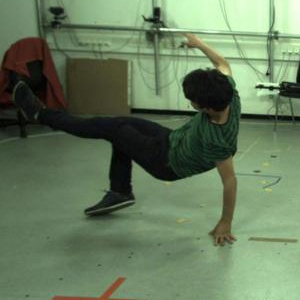}}\\%
    {\footnotesize Ref. Image}\\
}%
\hfill%
\parbox[t]{\qualposerefine}{%
    \vspace{0mm}\centering%
    \fbox{\includegraphics[width=\qualposerefine]{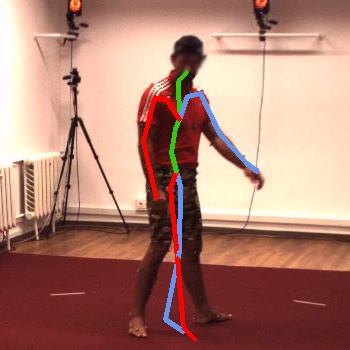}}\\%
    \fbox{\includegraphics[width=\qualposerefine]{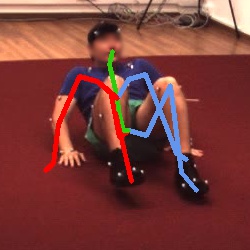}}\\%
    \fbox{\includegraphics[width=\qualposerefine]{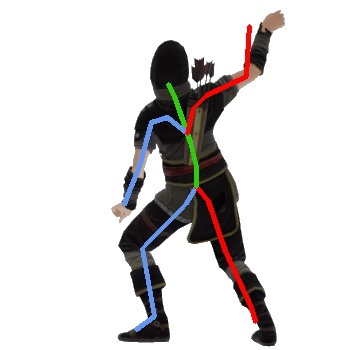}}\\%
    \fbox{\includegraphics[width=\qualposerefine]{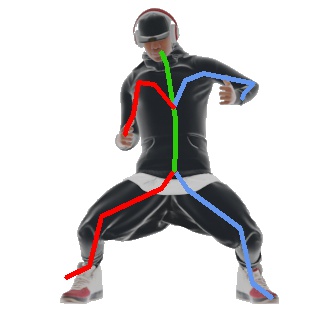}}\\%
    \fbox{\includegraphics[width=\qualposerefine]{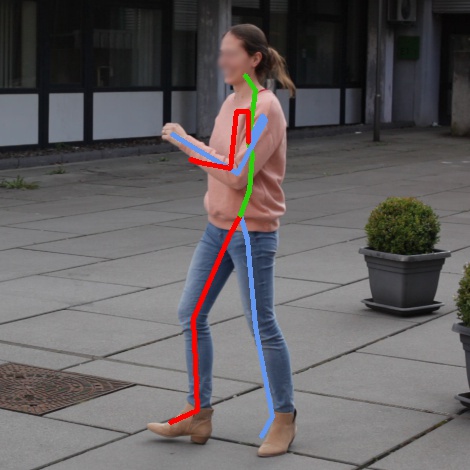}}\\%
    \fbox{\includegraphics[width=\qualposerefine]{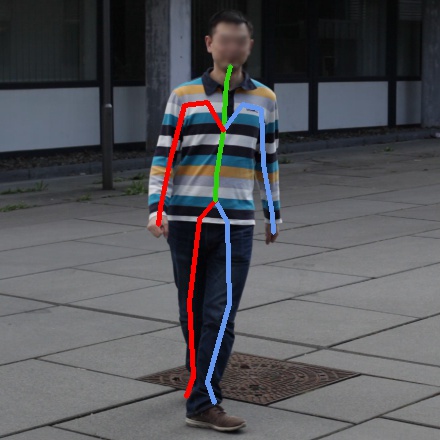}}\\%
    \fbox{\includegraphics[width=\qualposerefine]{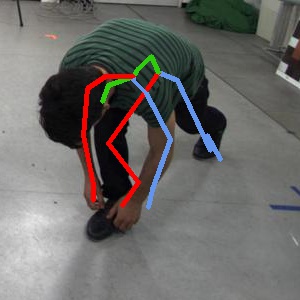}}\\%
    \fbox{\includegraphics[width=\qualposerefine]{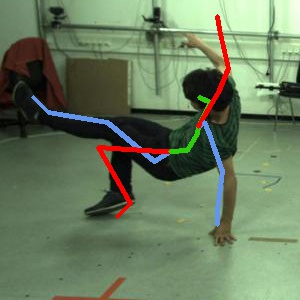}}\\%
    {\footnotesize SPIN~\cite{kolotouros2019learning_spin}}
}%
\hfill%
\parbox[t]{\qualposerefine}{%
    \vspace{0mm}\centering%
    \fbox{\includegraphics[width=\qualposerefine]{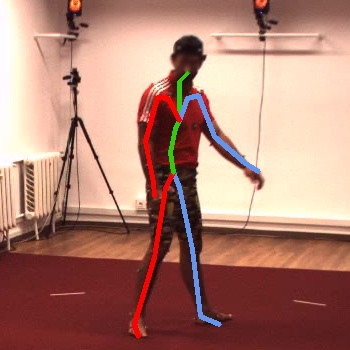}}\\%
    \fbox{\includegraphics[width=\qualposerefine]{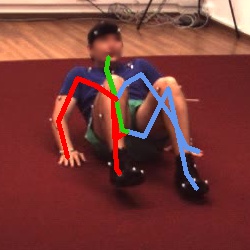}}\\%
    \fbox{\includegraphics[width=\qualposerefine]{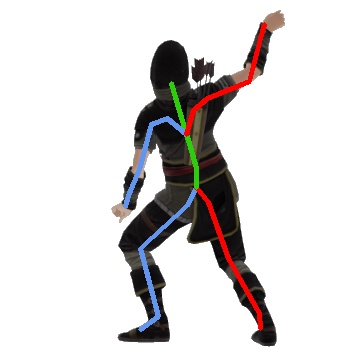}}\\%
    \fbox{\includegraphics[width=\qualposerefine]{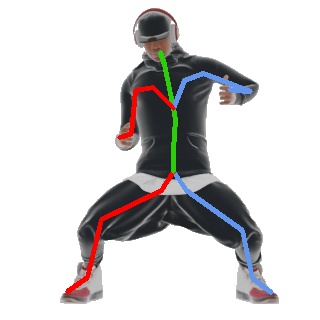}}\\%
    \fbox{\includegraphics[width=\qualposerefine]{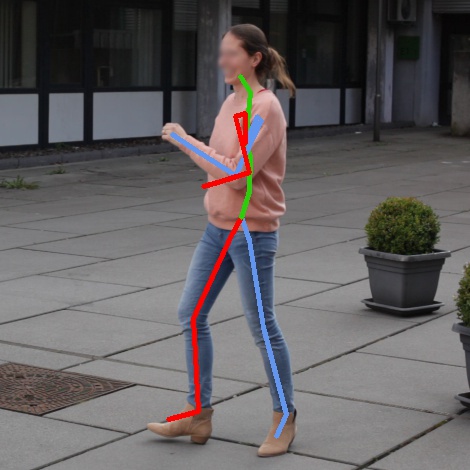}}\\%
    \fbox{\includegraphics[width=\qualposerefine]{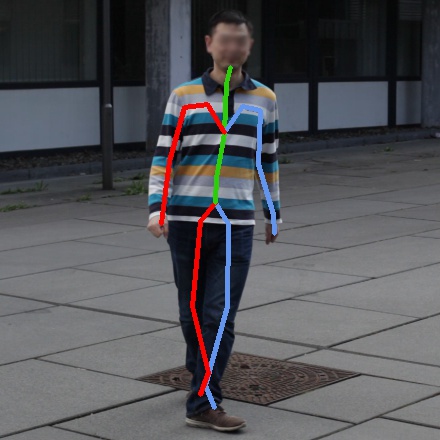}}\\%
    \fbox{\includegraphics[width=\qualposerefine]{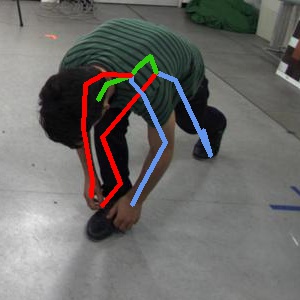}}\\%
    \fbox{\includegraphics[width=\qualposerefine]{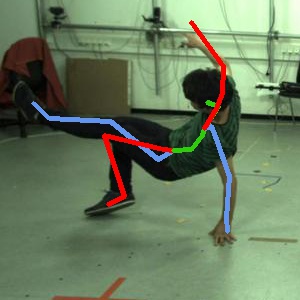}}\\%
    {\footnotesize Refined (Ours)}
}%
\caption{\tb{A-NeRF optimizes the body poses to align better with the images.} \tbi{Real faces and their reconstructions are blurred in all figures for anonymity.}}
\label{fig:qual-pose-extra}
\end{figure}
We show additional pose refinement results in~\figref{fig:qual-pose-extra}. A-NeRF estimates human poses that align better with the training images. 

\section{Ablation studies}
\label{sec:ablation-std}

In the following, we quantify the improvements brought about by our contributions concerning pose estimation accuracy and image generation quality. The results support the ablation summary point i), iii) and iv) given in the main document.

\begin{table}[t]
\caption{\textbf{PA-MPJPE improvements for different joints on Human 3.6M Protocol~\RNum{1}.} The higher, the better. A-NeRF helps improve pose accuracy on joints with high initial errors (elbows, wrists, knees, and ankles).}
\centering
\resizebox{\linewidth}{!}{
\setlength{\tabcolsep}{3pt}
\begin{tabular}{lccccccccccc}
\toprule
 & head top $\uparrow$ & neck $\uparrow$ &  shoulders $\uparrow$ & elbows $\uparrow$ & wrists $\uparrow$ & hips $\uparrow$ & knees $\uparrow$ & ankles $\uparrow$ & pelvis $\uparrow$ & spine $\uparrow$ & head $\uparrow$\\ 
\midrule
A-NeRF w/o smoothness prior & 1.90 &2.35 & 0.49& 3.77&9.04 &0.90 &3.10 &3.78 & 0.87 &0.86 &3.32\\
\rowcolor{Gray}
A-NeRF (Our full model) & 2.18&2.16 &0.67 &3.97 &9.44 &1.04 &3.28 &3.66 &1.04 &1.14 &3.28\\
\bottomrule
\end{tabular}
\label{tab:ablation-perjoint}
}
\end{table}

\begin{table}[t]
\caption{\textbf{Pose refinement influence evaluated on Human 3.6M S9 Protocol~\RNum2{2}.} Using Rel. Dist. encoding helps refine the estimated poses, and \textbf{Cutoff} can further improve the refinement.}
\centering
\resizebox{0.65\linewidth}{!}{
\setlength{\tabcolsep}{3pt}
\small
\begin{tabular}{lcc}
\toprule
 &   PA-MPJPE$\downarrow$  & Wrist-Improve $\uparrow$  \\ 
\midrule
SPIN & 43.67 &  0.00\\
\rowcolor{Gray}
Rel. Pos. + Cutoff & 41.90& 4.98 \\
Rel. Dist. & 41.34& 7.85\\
\rowcolor{Gray}
Rel. Dist. + Cutoff + No Reg & 41.21& 8.86 \\
Rel. Dist. + Cutoff (Ours) & \bf 40.97 & \bf 9.02\\
\bottomrule
\end{tabular}
\label{tab:ablation-refine}
}%
\end{table}

\begin{table}[t]
\caption{\textbf{Directional encoding impact.} The influence of the directional encoding is small but noticeable. It works best to transfer the ray direction relative to the bone coordinates.}
\centering
\resizebox{\linewidth}{!}{
\setlength{\tabcolsep}{3pt}
\begin{tabular}{llccc}
\toprule
Position Rep. & Direction Rep. &   PSNR~$\uparrow$ & SSIM~$\uparrow$ & \#dim for each Rep.  \\ 
\midrule
Rel. Dist. + Rel.~Dir. & World Ray Dir.    & 26.18 & 0.9456 & \bf\phantom{0}360 + 72 + \phantom{0}27 \\
\rowcolor{Gray}
Rel. Dist. + Rel.~Dir. &  Ray Ang.  & 30.13 & 0.9699 & \phantom{0}360 + 72 + 216 \\
Rel. Dist. + Rel.~Dir. &  Rel. Ray (our full model) & \bf{30.61} & \bf{0.9727} & \phantom{0}360 + 72 + 648 \\
    \bottomrule
\end{tabular}
\label{tab:ablation-view}
}
\end{table}

\begin{table}[t]
\caption{\textbf{Query encoding trade-off.} Encoding world coordinates does not succeed on motions and has a large memory consumption. Encoding positions relative to the skeleton works yet also has high dimensionality. Our full model that combines distance and direction performs well in both aspects. Note that we do not apply cutoff to these models. }
\centering
\resizebox{\linewidth}{!}{
\setlength{\tabcolsep}{3pt}
\begin{tabular}{llccc}
\toprule
Position Rep. & Direction Rep. &   PSNR~$\uparrow$ & SSIM~$\uparrow$ & \#dim for each Rep.   \\ 
\midrule
World Pos. + Joint Positions + $\smplpose$  & World Ray.  & 14.66 & 0.7554 & 1125 + 72 + \phantom{0}27 \\
\rowcolor{Gray}
World Pos. + Joint Positions + $\smplpose$  & Rel. Ang. & 15.90 & 0.7602 & 1125 + 72 + 216 \\
World Pos. + Joint Positions + $\smplpose$  & Rel. Ray. & 14.40 & 0.7185 & 1125 + 72 + 648 \\
\rowcolor{Gray}
Rel. Pos. & Ray Ang.         & 29.22 & 0.9638 & 1080 \phantom{+ 00 }+ 216 \\
Rel. Pos. & Rel. Ray.        & 29.25 & 0.9631 & 1080 \phantom{+ 00 }+ 648 \\
\rowcolor{Gray}
Rel. Dist. +Rel. Dir    & Ray Ang.         & \bf 29.99 & \bf 0.9702 &\bf \phantom{0}360 + 72 + 216 \\
Rel. Dist. +Rel. Dir (our model w/o cutoff) & Rel. Ray.        & 29.88 &  0.9692 & \phantom{0}360 + 72 + 648 \\
    \bottomrule
\end{tabular}
\label{tab:ablation-joint}
}
\end{table}

\begin{table}[t]
\caption{\textbf{Distance-based positional encoding} is compact (360 dim) but insufficient (lower PSNR and SSIM) to encode skeleton relative query locations unless paired with direction information in our full model (72 dim, w/o positional encoding). }
\centering
\resizebox{\linewidth}{!}{
\setlength{\tabcolsep}{3pt}
\begin{tabular}{llccc}
\toprule
Position Rep. &  Direction Rep. &  PSNR~$\uparrow$ & SSIM~$\uparrow$ & \#dim for each Rep.   \\ 
\midrule
Rel. Dist.          & Rel.~Ray & 23.76 & 0.9137 & \bf 360 + \phantom{0}0 + 648  \\
\rowcolor{Gray}
Rel. Dist. + $\theta$ & Rel.~Ray & 26.33 & 0.9340 & 360 + 72 + 648 \\
Rel. Dist. + Rel. Dir. (our full model)  & Rel.~Ray & \bf 30.61 & \bf 0.9727 & 360 + 72 + 648  \\
    \bottomrule
\end{tabular}
\label{tab:ablation-bone}
}
\end{table}

\begin{table}[t]
\caption{\textbf{Monocular vs. multi-view reconstruction.} The PSNR and SSIM scores show that our model can learn as well from a single view as from multiple ones. Having more views does not help if the poses are not diverse enough, as evidenced by rows 2 and 3.}
\centering
\resizebox{0.45\linewidth}{!}{
\setlength{\tabcolsep}{3pt}
\begin{tabular}{lcccc}
\toprule
 \#views & \#poses & \#imgs & PSNR~$\uparrow$ & SSIM~$\uparrow$ \\
  \midrule
1 &  1200 & 1200  & 28.32 & 0.9607\\
\rowcolor{Gray}
3 &  400 & 1200   & 29.10 & 0.9655\\
9 & 134 & 1206   & 29.00 & 0.9644\\
\rowcolor{Gray}
9 &  1200 & 10800  & 31.30 & 0.9750 \\
    \bottomrule
\end{tabular}
\label{tab:viewsyn-diffimg}
}
\end{table}

\begin{table}[t]
\caption{\textbf{Visual quality comparison between NeuralBody and A-NeRF.} .}
\centering
\resizebox{0.35\linewidth}{!}{
\setlength{\tabcolsep}{3pt}
\begin{tabular}{lcc}
\toprule
 Method & PSNR~$\uparrow$ & SSIM~$\uparrow$ \\
  \midrule
NeuralBody~\cite{peng2020arxiv_neuralbody}  & 23.86 & 0.9304\\
\rowcolor{Gray}
Ours & 29.10 & 0.9655\\

    \bottomrule
\end{tabular}
\label{tab:surreal_nb}
}
\end{table}

\begin{table}[t]
\caption{\textbf{Comparison of different strategies for encoding illumination effect on Human 3.6M S9 Protocol~\RNum{2} and validation split.} Using per-image codes and Rel. Ray. results in the best performance. World Ray causes overfitting in our static/estimated camera setting.}
\centering
\resizebox{0.85\linewidth}{!}{
\setlength{\tabcolsep}{3pt}
\small
\begin{tabular}{lcccc}
\toprule
 & w/ per-image code &  PSNR$\uparrow$ & SSIM $\uparrow$  & PA-MPJPE$\downarrow$   \\ 
\midrule
No Ray  & & 25.89 & 0.9132 & 41.99  \\
\rowcolor{Gray}
World Ray & & 26.04 & 0.9140 & 41.75\\
Rel. Ray. + World Ray & & 26.25 & 0.9170  & 41.00 \\
Rel. Ray. (our model w/o per-image code) & & 26.26 & 0.9167  & 41.03 \\
\midrule
\rowcolor{Gray}
No Ray & \checkmark & 26.60 & 0.9164 & 41.54  \\
World Ray & \checkmark & 26.55  & 0.9161 & 41.41  \\
Rel. Ray. + World Ray & \checkmark & 26.64 & 0.9192 & 41.00 \\
\rowcolor{Gray}
Rel. Ray (Ours) & \checkmark &  \bf27.27&  \bf0.9245 & \bf 40.97  \\
\bottomrule
\end{tabular}
\label{tab:view-dependent-effect}
}%
\end{table}

\begin{table}[t]
 \aboverulesep=0ex
 \belowrulesep=0ex
\caption{\textbf{Hyperparameters.} We tune the hyperparameters on Human 3.6M subject 1.}
\centering
\setlength{\tabcolsep}{3pt}
\resizebox{\linewidth}{!}{
\begin{tabular}{l|c|c|c|c|c}
\toprule
  & Mixamo  & Human 3.6M & MonoPerfCap & MPI-INF-3DHP  & SURREAL  \\ 
\midrule
$G$ (\#gradient accumulation) & 20 & 50,125$^\dagger$ & 20 & 10 & n/a$^\star$\\
\cmidrule{2-6}
\rowcolor{Gray}
$\lambda_\theta$ & \multicolumn{4}{c|}{2.0} & n/a$^\star$\\
$\lambda_t$ & \multicolumn{4}{c|}{0.05} & n/a$^\star$\\
\rowcolor{Gray}
$N_\text{batch}$  & \multicolumn{4}{c|}{3072} & 2048 \\
\cmidrule{2-6}
Coarse samples &  \multicolumn{5}{c}{64}\\
\rowcolor{Gray}
Fine samples & \multicolumn{5}{c}{16}\\
$L$ input (\#PE frequencies) & \multicolumn{5}{c}{7}\\
\rowcolor{Gray}
$L$ view (\#PE frequencies) & \multicolumn{5}{c}{4}\\
t (cutoff distance) &\multicolumn{5}{c}{500mm}\\
\rowcolor{Gray}
$\tau$ (cutoff temperature) &\multicolumn{5}{c}{-20$\xrightarrow{\text{exp dec.~250k}}$-200}\\
Learning rate $\phi$ &\multicolumn{5}{c}{5e-4$\xrightarrow{\text{exp dec.~500k}}$5e-5}\\
\cmidrule{2-6}
\rowcolor{Gray}
Learning $\phi$ rate (finetune) &\multicolumn{4}{c|}{2e-4$\xrightarrow{\text{exp dec.~200k}}$5e-5} & n/a$^\star$\\
\cmidrule{2-6}
Learning rate $\smplpose$ 
&\multicolumn{3}{c|}{5e-4} & 1e-4 & n/a$^\star$\\
\bottomrule
\multicolumn{6}{l}{\footnotesize$^\star$ When GT poses are used for training, we do not enable refinement and scratch the associated hyperparameters.}\\%
\multicolumn{6}{l}{\footnotesize$^\dagger$ for Protocol \RNum{1}, which has around 3 times more images than Protocol \RNum{2}.}\\%
\end{tabular}}
\label{tab:hyperparam}
\end{table}
\begin{figure*}[t]
\centering
\includegraphics[width=0.99\linewidth,trim=280 0 80 0,clip]{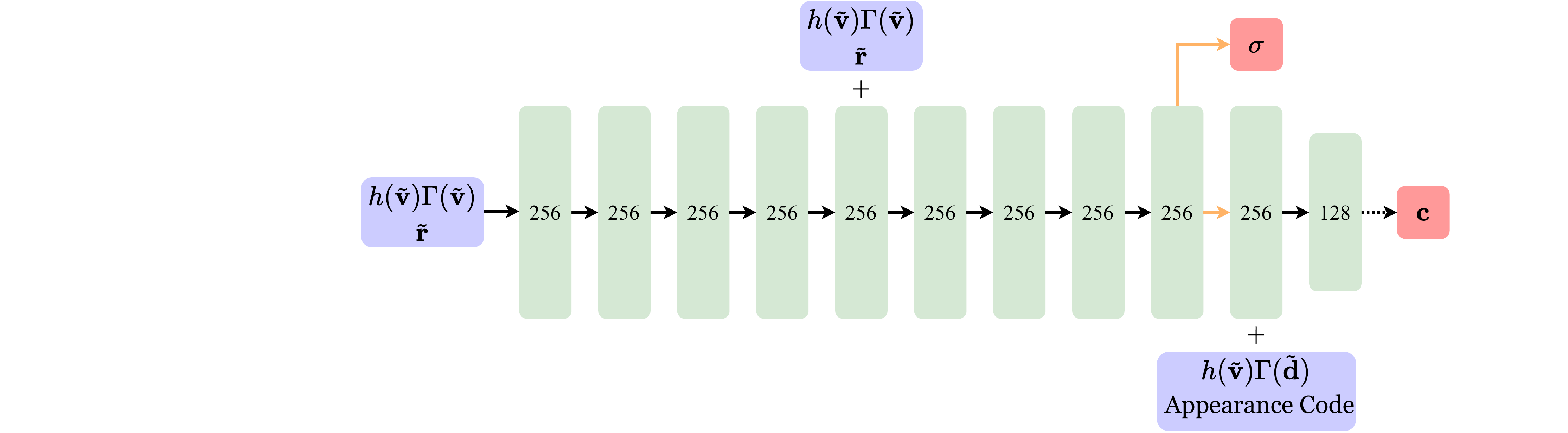}
\caption{A-NeRF architecture follows the original NeRF~\cite{mildenhall2020nerf}, with an additional appearance code similar to concurrent works~\cite{martin2020arxiv_nerfiw,peng2020arxiv_neuralbody} for handling illumination. We use purple blocks for inputs, green blocks for intermediate features with numbers representing their sizes, red blocks for outputs. Black and dashed arrows imply the use of ReLU and sigmoid activations respectively. Orange arrows indicate that no activation is applied. Appearance code and our Rel. Ray. encoding are fed into the network in the second last fully-connect layer so that these representation do not affect human body geometry (as in the original NeRF).}
\label{fig:network-archi}
\end{figure*}
\subsection{Pose Estimation Accuracy Evaluation}

We performed the following experiments on the Human3.6M benchmark using the PA-MPJPE metric.
~\paragraph{Joint-wise PA-MPJPE Improvements.} The MPJPE metric is an average over all the joints. To provide a more detailed view, ~\tabref{tab:ablation-perjoint} shows PA-MPJPE improvements individually for every body part, in relation to the initialization. A-NeRF drastically improves those joints that have a high initial error (e.g., elbows, wrists, knees, ankles). The most significant is the wrist. These individual improvements highlight the importance of refining initial pose estimates, particularly for applications requiring the precise position of end effectors.

\paragraph{Impact of Radial Distance Encoding.} In~\tabref{tab:ablation-refine}, we show that Rel. Pos. encoding $\local{\query}_k$ does not improve the overall PA-MPJPE, despite being able to refine end effectors. By contrast, our proposed Rel. Dist. encoding leads to a 57\% improvement on the wrist over the Rel. Pos. encoding, while improving the overall PA-MPJPE by 0.56. This showcases that %
\textbf{\emph{i) Embedding relative 3D position, $\local{\query}_k$, yields only half as good pose refinements as our proposed Rel. Dist., $\querydist$, which captures radial information.}}
\paragraph{Impact of Cutoff on Refinement.} In~\tabref{tab:ablation-refine}, we show that incorporating \textbf{Cutoff} can further increase the pose accuracy, with a 15\% improvement on wrists over our model without \textbf{Cutoff}.
\paragraph{Impact of Pose Regularization.} In~\tabref{tab:ablation-refine}, we show that the pose regularizer yields moderate improvement on the pose refinement results, and is not strictly essential for our learning framework. 

\subsection{Image Generation Quality Evaluation}
We evaluate the visual quality of different variants using SURREAL dataset (see~\secref{sec:dataset-supp} for dataset details). Similarly, image quality is quantified via the PSNR and SSIM within the character bounding boxes, comparing the rendered to the ground truth image.
We use the test set of 1,500 images showing the training character in novel camera view and novel pose. Here, the ground-truth pose is given as input for training and is not further refined.

On these synthetic sequences, all models are trained using the L2 distance, with ground truth human poses and camera. The batch size is 
$N_\text{batch}=2048$, and each model is trained for 150k iterations unless stated otherwise. 

\paragraph{Impact of the Skeleton-Relative Encoding Variants}
In addition to our proposed Rel. Dist., Rel. Dir. and Rel. Ray. encodings, we experiment with other alternatives:
\begin{itemize}[leftmargin=0.7cm]
\item \textbf{Relative Ray Angle (Ray Ang.)} A low-dimensional alternative to our Rel. Ray. encoding is to calculate the angle between the ray direction and the vector from query point to joint $m$
\begin{equation}
\viewdir'=\veclistki{\viewdir'}{24},\quad\viewdir'_{k,m} = \text{arc cos}(\local{\viewdir}_{k,m} \bullet \local{\query}_{k,m})\in\bR.
\end{equation}
As shown in \tabref{tab:ablation-view}, our proposed Rel. Ray. offers better image quality.
\item \textbf{Bone-relative Position (Rel. Pos.)} \tabref{tab:ablation-joint} shows the result of different query position encoding. Note that we do not apply \textbf{Cutoff} to these models. The straightforward baselines (World Position + Joint Position + $\theta$) that condition on pose directly perform the worst. We find that these baselines achieve 0.7 SSIM by simply rendering background colors. On the other hand, Rel. Pos. significantly outperforms the straightforward baselines. However, it does not provide any gain over our proposed Rel. Dist. + Rel. Dir. encoding, albeit having drastically more dimensions. This shows that \textbf{\emph{ii) Our embedding choices keep the dimensionality moderate while improving on or matching the PSNR of higher-dimensional variants.}}
\item \textbf{Conditioning on pose.} In ~\tabref{tab:ablation-bone}, we further compare different ways of encoding the bone directions. Our proposed Rel. Dir. encoding shows result superior to the baseline that conditioned directly on $\theta$.
\end{itemize}

\paragraph{Numbers of Poses and Views on Visual Quality.} We experiment A-NeRF with varying numbers of poses and views. All models are trained for 300k iterations in this experiment. The result in~\tabref{tab:viewsyn-diffimg} shows not only that A-NeRF can be trained with only a single view but also that adding more views does not help if the poses are not diverse enough. As evidenced by rows 2 and 3, it helps to supply additional images of new poses, nearly as much as adding new views of the same pose. We conclude that \textbf{\emph{iii) For a fixed number of images with accurate poses, the visual quality of one long video plus diverse poses is comparable to multiple shorter multi-view recordings, which makes the simpler monocular capture set-up preferable.}} 

\section{Visual Comparison to NeuralBody.}  
\label{sec:surreal-visual-comp}
To see how our proposed A-NeRF compares to surface-based approaches in capturing articulated textured avatars, without including the uncertainty of inaccurate initial pose estimates, we train both NeuralBody and A-NeRF on SURREAL with accurate poses and ground truth camera, similar to \secref{sec:ablation-std}. Precisely, both models are trained on SURREAL with images from 3 out of 9 cameras in the training data, with 400 randomly sampled poses for 300K iterations (the same setting as in~\tabref{tab:viewsyn-diffimg}). We report the result in~\tabref{tab:surreal_nb}. We observe that NeuralBody produces slightly deformed/expanded body features for poses that are very different from the training data\footnote{We have contacted the authors of NeuralBody, and confirmed that such artifacts are expected for unseen test poses.}. We conclude that, while NeuralBody shows impressive results on reenacting human performance with moderate sequence length, A-NeRF works better for longer sequences and for retargeting to diverse poses.

\section{Discussion on View-dependent Effects}
\label{sec:view-dependent}
To approximate the rendering equation, we need the position in space, incoming light and the outgoing ray direction. In the original NeRF, the light sources are not explicitly reconstructed, but are instead baked into the view directions. Thus, the incoming light information is represented as a black-box function. 
In our case, we assume a static, estimated camera for each of the images. As a result, light sources are inconsistent between frames, and therefore the model can no longer encode the incoming light information into global view directions (World Ray).
To counter this issue, we use the per-image code design~\cite{martin2020arxiv_nerfiw,peng2020arxiv_neuralbody} to represent the illumination effects as a black box. Furthermore, our Rel. Ray. explicitly encodes the outgoing ray direction with respect to each body part, allowing the model to better explain the illumination on the human body.

We conduct experiments on Human3.6M S9 to examine how the per-image code,  Rel. Ray., and World Ray affect the performance on both pose refinement and rendering quality. As shown in~\tabref{tab:view-dependent-effect}, learning with World Ray led to reduced performance, and we observe flickering results for novel view rendering (e.g., bullet-time effect). This indicates that the camera (world) causes overfitting in our single/estimated camera setting and that it is beneficial to encode all quantities relative to the skeleton. Not using any directional information (No Ray) yields the worst results. Finally, the per-image code shows visible improvement in all cases, and it attains the best performance in combination with Rel. Ray. encoding. The outcomes support our claim that, by modeling the outgoing ray direction explicitly in relative space, we enable A-NeRF to better recover the illumination effect. Finally, we acknowledge that, while the lighting effect looks plausible in our novel view synthesis results (\figref{fig:qual-nvs-extra}), the illumination is not physically correct. But since our goal is to model an articulated human avatar instead of a recreating 3D scene with perfect light sources, we argue that our proposed Rel. Ray. serves our purpose well. It is a promising direction for future work to model illumination more explicitly and to enable relighting applications.

\section{Dataset Details}
\label{sec:dataset-supp}
In addition to the real-world datasets introduced in the main paper, we experiment A-NeRF on two synthetic datasets.
\begin{itemize}[leftmargin=0.7cm]%
\item \textbf{Human 3.6M~\cite{ionescu_human36_pami14}} For Protocol \RNum{1}, we follow~\cite{kanazawa2018hmr,kocabas2020vibe,kolotouros2019learning_spin} to evaluate PA-MPJPE on every $5^{th}$ frame of the frontal camera. For Protocol \RNum{2}, PA-MPJPE is calculated on every $64^{th}$ frame from all cameras, following~\cite{nibali20193d,sun2017compositional}.
\item \textbf{SURREAL~\cite{varol17_surreal}} We generate a synthetic dataset using~\cite{varol17_surreal} to examine how each factor affects the visual quality. We use 1500 3D poses from~\cite{CMU}, divide them into a 4-1 train-test split. Each training pose is rendered with 9 different cameras, resulting in a total of 10,800 $512\times512$ training images. We render the 300 testing poses with 5 different cameras to construct a test set of 1,500 images.
\item \textbf{Mixamo~\cite{mixamo}} 
We generate two synthetic subjects, ``James'' and ``Archer'' from Mixamo, and render each subject from varying camera angles with three provided motions, ``Thriller Part 3'', ``Robot-Hip-Hot Dance'', and ``Shoved Reaction with Spin''. Each subject has 1,130 training images. We extract pose and camera parameters with~\cite{kolotouros2019learning_spin}.
\end{itemize}
\section{Implementation Details}
\label{sec:imp-det}
Our A-NeRF model is learned without supervision on a single or multiple videos of the same person. Camera intrinsics, bone lengths for setting $\va_m$, and pose $\theta_k$ are initialized with~\cite{kolotouros2019learning_spin} for every frame $k$. These poses are then optimized for 500k iterations on objective Eq.\ref{eq:objective}, together with the neural parameters $\phi$, which model shape and appearance. We then stop optimizing pose and continue to train $\phi$ with only the data term for additional 200k iterations, which further improves visual fidelity.
We form a training batch by randomly sampling $N_\text{batch}=3072$ rays from all available images. Therefore, an image $k$ can have no or only very few samples, leading to noisy updates of $\smplpose_k$. To counteract, we accumulate the gradient update $\Delta \smplpose$ for $G\geq(30\dot N_\text{batch})/N$ iterations, so that each $\smplpose_k$ is expected to be sampled $30\geq$ times before the gradient update. To speed up the rendering part, we use the same coarse-to-fine importance sampling strategy of \cite{mildenhall2020nerf} and restrict samples to be within a coarse human silhouette estimated with~\cite{chen2017rethinking} unless the background is monochrome.
The training is done on 2 Nvidia Tesla V100 32GB GPUs, which takes approximately 60 hours. It takes around 1-4 seconds to render one $512\times512\times3$ image using a single Nvidia Tesla V100 GPU. We tuned the hyperparameters on Subject 1 of the Human3.6M training set. 

\paragraph*{Skeleton Representation.}
We use a skeleton representation that encodes the connectivity and static bone lengths via a rest pose of the 3D joint locations $\{\jointele_m\}_{m=1}^{24}$, with the root at $\mathbf{0}$ that are connected via $B$ bones. Dynamics are modeled with per-frame skeleton poses $\smplpose_k=\veclistkiz{\smplbone}{24}$ that include the relative rotation of 24 joints, $\subkm{\smplbone}$, to their parents as well as the global position $\omega_{k,0}$ as a special case. For rotation, we experiment with the recently proposed overparametrized representation of \cite{zhou2019continuity} ($\subkm{\smplbone} \in\smplbonespaceNEW$) and the traditional axis-angle representation ($\subkm{\smplbone} \in\smplbonespace$).
The subscript $_{k,m}$ indicates that a variable is related to the $m$-th joint of image $\mI_k$. 
We will exploit that every bone defines a local coordinate system. We write the mapping of a 3D position $\subkm{\vp} \in\jointelespace$ in the $m$-th local bone coordinates to world coordinates $\vq \in\jointelespace$ via forward kinematics using homogeneous coordinates,
\begin{align}
\hmvec{\vq} = \jointtrans \hmvec{\subkm{\vp}}, 
 \text{ with transformation }
\jointtrans = \prod_{l\in A(m)}\begin{bmatrix}
\mR(\subkl{\omega}) & \jointele_{l,l-1} \\ 
\mathbf{0} & 1
\end{bmatrix} \in \R^{4 \times 4},%
\label{eq:bone transformation supp}
\end{align}
$A(m)$ the ordered set of the joint ancestors of $m$, and $\mathbf{\jointele}_{l,l-1}\in\jointelespace$ is the translation between joint $l$ and its child joint $l-1$. Conversely, $\jointtrans^{-1}$ maps world to local bone coordinates. The rotation matrix $\mR(\subkl{\omega})$ is inferred as in~\cite{zhou2019continuity}.
Note that our skeleton is equivalent to SMPL~\cite{loper2015smpl} and others, but without their parametric surface model, and can therefore be initialized with any skeleton pose estimator.
\paragraph*{Multi-view extension.}
We can incorporate a multi-view constraint to improve pose refinement when the motion is captured from multiple cameras. For initializing $\smplpose_k$, we average the individual joint rotation estimates from all $V$ views $v \in [1, \dots, V]$. Since rotations are relative to the parent and root, this works without calibrating the cameras. Only the global position and orientation remain specific to view $v$. To this end, we extend our single-view notation with subscripts, position $\smplbone^{(v)}_{k,1}$ and orientation $\smplbone^{(v)}_{k,2}$ are estimated relative to camera $v$.
With slight abuse of notation we rewrite Eq.~\ref{eq:objective},
\begin{equation}
\cL^\text{MV}(\smplpose, \phi) = \sum_v \sum_k d(\render(\smplpose_k, \smplbone^{(v)}_{k,0}, \smplbone^{(v)}_{k,1}), \mI^{(v)}_k) +\lambda_\smplpose d\left(\smplpose_k -\hat{\smplpose}_k\right) +
\lambda_t \norm{\frac{\partial^2 \smplpose_k}{\partial t^2}}^2_2,
\label{eq:objective multiview}    
\end{equation}
with $\smplpose_k$ shared across all views $v$, except for $\smplbone^{(v)}_{k,0}$ and $\smplbone^{(v)}_{k,1}$ which are estimated independently per camera view $v$ since the relative camera position and orientation is unknown in our setting. 

\paragraph{Efficient Sampling.} 
To increase sampling efficiency, we define a cylinder surrounding the initial skeleton pose (as sketched in~\figref{fig:overview}). The radius of the cylinder is defined as the distance between the root joint and the joint farthest away from it, plus a 250mm buffer length. We sample points along the ray segment that lies inside the cylinder. Furthermore, as explained in the main paper, we restrict samples to be within a foreground mask. Because the mask is only approximate, we dilate the estimated foreground mask by 5 pixels.

\paragraph{Hyperparameter tuning.}
~\tabref{tab:hyperparam} shows our hyperparameters for A-NeRF on different datasets. All parameters are tuned on Subject 1 of H3.6M, with only the learning rate changed to apply to the 3DHP dataset with much shorter sequence sizes and the number of gradient accumulations computed relative to the number of frames as defined in the main document.

\paragraph{Network architecture.}
We start from the same network architectures as in the original NeRF~\cite{mildenhall2020nerf} (see~\figref{fig:network-archi}). 
Following concurrent works on handling illumination changes \cite{martin2020arxiv_nerfiw,peng2020arxiv_neuralbody}, we add a 16-dimensional appearance code to the second last layer of the NeRF network $\nerffunc$. It is individually stored and optimized for every frame. Note that the NeuralBody baselines use 128-dimensional codes. Due to its position at the end of the network and its low dimensionality, it helps learning these global effects while not deteriorating the benefits of the relative encoding.

\paragraph{NeuralBody baseline}
For a fair comparison of A-NeRF to NeuralBody, we increase the batch size for NeuralBody so that the number of training samples per batch will be the same (which improves its performance). We keep other hyperparameters the same as provided in the official implementation and re-train models on our datasets.

\paragraph{Novel view synthesis.}

The novel viewpoints in all our qualitative visualization are generated by regular sampling from two different camera trajectories: (1) cameras spaced from 0 to 360 degrees (bullet-time effect), and (2) a circular trajectory that rotates from -25 to 25 degrees along the y-axis and -15 to 15 degrees along the x-axis (elevation), with an additional zoom-in and zoom-out effect.
\end{appendices}

\end{document}